\documentclass[english]{article}
\usepackage{microtype}
\usepackage{booktabs} 
\usepackage{array}
\usepackage{siunitx}
\usepackage{pifont}
\usepackage[T1]{fontenc}
\usepackage[latin9]{inputenc}
\usepackage{hyperref}
\usepackage{amsmath,mathtools}
\usepackage{amsthm}
\usepackage{amssymb}
\usepackage{graphicx}
\usepackage{subcaption}
\usepackage{thmtools,thm-restate}
\usepackage{tikz}
\usepackage{pgfplots}
\usepackage{babel}
\usepackage{epigraph}
\usepackage{paralist}
\usepackage[affil-it]{authblk}
\usepackage{placeins}
\usepackage{enumerate}   
\usepackage{xcolor}

\usepackage[accepted]{icml2018}

\pgfplotsset{compat=1.13}
\makeatletter
\newtheorem{thm}{Theorem}
\numberwithin{thm}{section} 

\newtheorem{lemma}[thm]{Lemma}
\newtheorem{definition}[thm]{Definition}
\newtheorem{remark}[thm]{Remark}

\newtheorem{innerassumption}{Assumption}
\newenvironment{assumption}[1]
  {\renewcommand\theinnerassumption{#1}\innerassumption}
  {\endinnerassumption}

\newcommand{\raisemath}[1]{\mathpalette{\raisem@th{#1}}}
\newcommand{\raisem@th}[3]{\raisebox{#1}{$#2#3$}}
\newcommand{\T}{{\raisemath{-1pt}{\mathsf{T}}}}
 
\DeclareMathOperator{\sign}{sign}
\DeclareMathOperator{\E}{E}	
\DeclareMathOperator{\supp}{supp}
\DeclareMathOperator{\Real}{Re}
\DeclareMathOperator{\Imag}{Im}

\pgfmathdeclarefunction{gauss}{2}{%
  \pgfmathparse{1/(#2*sqrt(2*pi))*exp(-((x-#1)^2)/(2*#2^2))}%
}

\pgfmathdeclarefunction{linear}{2}{%
  \pgfmathparse{#1 + #2 * x}%
}

\icmltitlerunning{Which Training Methods for GANs do actually Converge?}

\begin{document}
\twocolumn[
\icmltitle{Which Training Methods for GANs do actually Converge?}

\begin{icmlauthorlist}
\icmlauthor{Lars Mescheder}{mpi}
\icmlauthor{Andreas Geiger}{mpi,eth}
\icmlauthor{Sebastian Nowozin}{msr}
\end{icmlauthorlist}

\icmlaffiliation{mpi}{MPI T\"ubingen, Germany}
\icmlaffiliation{msr}{Microsoft Research, Cambridge, UK}
\icmlaffiliation{eth}{ETH Z\"urich, Switzerland}

\icmlcorrespondingauthor{Lars Mescheder}{lars.mescheder@tue.mpg.de}

\icmlkeywords{Generative Models, GANs, Convergence, Optimization, Game theory, Machine Learning, ICML}

\vskip 0.3in
]

\printAffiliationsAndNotice{}

\begin{abstract}
Recent work has shown local convergence of GAN training for absolutely continuous
data and generator distributions.
In this paper, we
show that the requirement of absolute continuity is necessary: we
describe a simple yet prototypical counterexample showing that in the more
realistic case of distributions that are not absolutely continuous,
unregularized GAN training is not always convergent.
Furthermore, we discuss regularization strategies that were
recently proposed to stabilize GAN training. 
Our analysis shows that GAN training with instance noise or
zero-centered gradient penalties
converges.
On the other hand, we show that
Wasserstein-GANs and WGAN-GP with a finite number of discriminator updates
per generator update
do not always converge to the equilibrium point.
We discuss these results, leading us to a new explanation
for the stability problems of GAN training.
Based on our analysis, we extend our convergence results to
more general GANs and prove local convergence 
for simplified gradient penalties even if the generator and 
data distributions lie on lower dimensional manifolds.
We find these penalties to work well in practice and use them
to learn high-resolution generative image models for a
variety of datasets with little
hyperparameter tuning.
\end{abstract}

\section{Introduction}
Generative Adversarial Networks (GANs) \cite{DBLP:conf/nips/GoodfellowPMXWOCB14}
are powerful latent variable models that can be used to learn
 complex real-world distributions.
 Especially for images, GANs have emerged as one of the dominant 
 approaches for generating new realistically looking samples after the model
 has been trained on some dataset.
 
\begin{table}[t]
\newcommand{\notconvergent}{\textcolor{red!70!black} {\ding{55}} }
\newcommand{\convergent}{\textcolor{green!50!black} {\ding{51}} }
\centering
\scalebox{0.8}{  	
\begin{tabular}{p{5.5cm}>{\centering}p{2cm}>{\centering}p{2cm}} \toprule
Method & Local convergence (a.c.~case) & Local convergence (general~case)  \tabularnewline \midrule
unregularized \cite{DBLP:conf/nips/GoodfellowPMXWOCB14}
&  \convergent &   \notconvergent  \tabularnewline
WGAN 
\cite{DBLP:journals/corr/ArjovskyCB17}
&\notconvergent & \notconvergent \tabularnewline
WGAN-GP
\cite{DBLP:conf/nips/GulrajaniAADC17}
& \notconvergent & \notconvergent \tabularnewline
DRAGAN \cite{DBLP:journals/corr/KodaliAHK17}
& \convergent & \notconvergent \tabularnewline
Instance noise  \cite{DBLP:journals/corr/SonderbyCTSH16} 
& \convergent &  \convergent  \tabularnewline
ConOpt
\cite{DBLP:conf/nips/MeschederNG17}
& \convergent & \convergent \tabularnewline
Gradient penalties
\cite{DBLP:conf/nips/RothLNH17} & \convergent & \convergent \tabularnewline
Gradient penalty on real data only &  \convergent &  \convergent \tabularnewline
Gradient penalty on fake data only &  \convergent &  \convergent \tabularnewline
\bottomrule
\end{tabular}
}
\caption{Convergence properties of different GAN training algorithms for
general GAN-architectures.
Here, we distinguish between the case where both the data and generator distributions are
absolutely continuous (a.c.) and the
general case where they may lie on lower dimensional manifolds.}
\vspace{-15pt}
\end{table}

However,
while very powerful, GANs can be hard to train and in practice
it is often observed that gradient descent based GAN
optimization does not lead to convergence.
As a result, a lot of recent research has focused on 
finding better training algorithms
\cite{DBLP:journals/corr/ArjovskyCB17, DBLP:conf/nips/GulrajaniAADC17,
DBLP:journals/corr/KodaliAHK17,DBLP:journals/corr/SonderbyCTSH16,DBLP:conf/nips/RothLNH17}
for GANs  as well as gaining
better theoretically understanding of
their training dynamics 
\cite{DBLP:journals/corr/ArjovskyCB17,DBLP:journals/corr/ArjovskyB17,
DBLP:conf/nips/MeschederNG17,DBLP:conf/nips/NagarajanK17,
DBLP:conf/nips/HeuselRUNH17}.

Despite practical advances, the training dynamics of GANs
are still not completely understood.
Recently, \citet{DBLP:conf/nips/MeschederNG17} and \citet{DBLP:conf/nips/NagarajanK17} 
showed
that local convergence and stability properties of GAN training can
be analyzed by examining the eigenvalues of the Jacobian of the the associated gradient
vector field:
if the Jacobian
has only eigenvalues with negative real-part at the equilibrium point,
GAN training converges locally for small enough learning rates.
On the other hand, if the Jacobian
has eigenvalues on the imaginary axis, it is generally not locally convergent.
Moreover, \citet{DBLP:conf/nips/MeschederNG17} showed that
if there are eigenvalues close but not on the imaginary axis,
the training algorithm can require intractably small learning rates to achieve
convergence.
While \citet{DBLP:conf/nips/MeschederNG17} observed eigenvalues close to the imaginary axis in practice,
this observation does not answer the question if eigenvalues close to the imaginary axis are a general phenomenon and if 
yes, whether they are indeed the root cause for the training instabilities that people
observe in practice.

A partial answer to this question was given by
\citet{DBLP:conf/nips/NagarajanK17}, who showed that for absolutely
continuous
data and generator distributions\footnote{%
\citet{DBLP:conf/nips/NagarajanK17} also proved local convergence for a slightly more general family of 
probability distributions where the support of the generator distribution is equal to the support of the true data distribution near the equilibrium point.
Alternatively, they showed that their results also hold when
the discriminator satisfies certain (strong) smoothness conditions.
However, these conditions are usually hard to satisfy in practice without prior knowledge about the support of the true data distribution.
}
all eigenvalues of the Jacobian
have
negative real-part.
As a result, GANs
are locally convergent for small enough learning rates in this case.
However, the assumption of absolute continuity
is not true for common use cases of GANs, where both distributions may
lie on lower dimensional manifolds \cite{DBLP:journals/corr/SonderbyCTSH16,DBLP:journals/corr/ArjovskyB17}.

In this paper we show that this assumption
is indeed necessary:
by considering a simple yet prototypical example of GAN training we 
analytically show that (unregularized) GAN training is not always locally convergent.
We also discuss how recent techniques
for stabilizing GAN training affect local convergence on our example problem.
Our findings show that neither Wasserstein GANs (WGANs)
\cite{DBLP:journals/corr/ArjovskyCB17} nor Wasserstein GANs with Gradient Penalty 
(WGAN-GP) \cite{DBLP:conf/nips/GulrajaniAADC17} nor DRAGAN
\cite{DBLP:journals/corr/KodaliAHK17}
converge on this simple example for a fixed number of discriminator updates per
generator update.
On the other hand, we show
that instance noise \cite{DBLP:journals/corr/SonderbyCTSH16,DBLP:journals/corr/ArjovskyB17},
zero-centered
gradient penalties \cite{DBLP:conf/nips/RothLNH17}
and consensus optimization \cite{DBLP:conf/nips/MeschederNG17}
lead to local convergence.

Based on our analysis, we give a new explanation for the instabilities commonly observed when training GANs
based on discriminator gradients orthogonal to the tangent space of the data manifold.
We also introduce simplified gradient penalties
for which we prove local convergence. We find that these gradient penalties work well in practice, 
allowing us to learn high-resolution image based generative models for a variety of datasets with little hyperparameter tuning.

In summary, our contributions are as follows:
\begin{compactitem}
 \item We identify a simple yet prototypical counterexample showing that (unregularized) gradient descent based GAN optimization
 is not always locally convergent
 \item We discuss	
 if and how recently introduced regularization techniques stabilize the training
 \item We introduce simplified gradient penalties
 and prove local convergence for the regularized GAN training dynamics 
 \end{compactitem}
All proofs can be found in the supplementary material.

\section{Instabilities in GAN training}\label{sec:main}
\subsection{Background}
GANs are defined by a min-max two-player game between a discriminative network $D_\psi(x)$ and generative network $G_\theta(z)$.
While the discriminator tries to distinguish between real data point and data points produced by the generator,
the generator tries to fool the discriminator. It can be shown \cite{DBLP:conf/nips/GoodfellowPMXWOCB14} that if both the 
generator and discriminator are powerful enough to approximate any real-valued function, the unique Nash-equilibrium
of this two player game is given by a generator that produces the true data distribution and a discriminator which is $0$ everywhere on the data distribution.

Following the notation of \citet{DBLP:conf/nips/NagarajanK17}, the training objective for the two players can be described by an objective function of the form
\begin{multline}\label{eq:gan-loss}
L(\theta,\psi)= \mathrm{E}_{p(z)}\left[f(D_\psi(G_\theta(z)))\right] \\
+ \mathrm{E}_{p_{\mathcal D}(x)}\left[f(-D_\psi(x))\right]
\end{multline}
for some real-valued function $f$. The common choice $f(t) = -\log(1 + \mathrm \exp(-t))$ leads to the loss function considered in 
the original GAN paper  \cite{DBLP:conf/nips/GoodfellowPMXWOCB14}. For technical reasons we assume that $f$ is continuously differentiable and
satisfies $f^\prime(t) \neq 0$ for all $t \in \mathbb R$.

The goal of the generator is to minimize this loss  whereas the discriminator tries to maximize it.
Our goal when training GANs is to find a Nash-equilibrium,
i.e. a parameter assignment $(\theta^*, \psi^*)$ where neither the discriminator nor the generator can improve their utilities unilaterally.

GANs are usually trained using Simultaneous or Alternating Gradient Descent (SimGD and AltGD).
Both algorithms can be described as fixed point algorithms \cite{DBLP:conf/nips/MeschederNG17} that apply some operator $F_h(\theta, \psi)$
to the parameter values $(\theta, \psi)$ of the generator and discriminator, respectively. For example, simultaneous gradient descent corresponds to the operator
$F_h(\theta, \psi) = (\theta, \psi) + h\, v(\theta,\psi)$, where $v(\theta, \psi)$ denotes
the \emph{gradient vector field}
\begin{equation}
  v(\theta, \psi) := 
  \begin{pmatrix}
 - \nabla_\theta L(\theta, \psi) \\
  \nabla_\psi L(\theta, \psi)
 \end{pmatrix}.
\end{equation}
Similarly, alternating gradient descent can be described by an operator $F_h = F_{2,h} \circ F_{1, h}$ where $F_{1, h}$ and $F_{2, h}$ perform
an update for the generator and discriminator, respectively.

Recently, it was shown \cite{DBLP:conf/nips/MeschederNG17} that local convergence of GAN training near an equilibrium point $(\theta^*, \psi^*)$
can be analyzed by looking at the spectrum of the Jacobian $F_h^\prime(\theta^*, \psi^*)$ at the equilibrium:
if $F_h^\prime(\theta^*, \psi^*)$ has eigenvalues with absolute value bigger than $1$ , the training algorithm will generally not converge to $(\theta^*, \psi^*)$.
On the other hand, if all eigenvalues have absolute value smaller than $1$, the training algorithm will converge to $(\theta^*, \psi^*)$
with linear rate $\mathcal O(|\lambda_\mathrm{max}|^k)$ where $\lambda_{\mathrm{max}}$ is the eigenvalue of $F^\prime(\theta^*, \psi^*)$ with the biggest absolute value.
If all eigenvalues of $F^\prime(\theta^*, \psi^*)$ are on the unit circle,
the algorithm can be convergent, divergent or neither, 
but if it is convergent it will generally converge with a sublinear rate.
A similar result \cite{khalil1996noninear,DBLP:conf/nips/NagarajanK17} also holds for the (idealized) continuous system 
\begin{equation}\label{eq:gan-continuous}
 \begin{pmatrix}
  \dot \theta(t) \\
  \dot \psi(t)
 \end{pmatrix}
=
\begin{pmatrix}
 -\nabla_{\psi} L(\theta, \psi) \\
 \nabla_{\theta} L(\theta, \psi)
\end{pmatrix}
\end{equation}
which corresponds to training the GAN with infinitely small learning rate:
if all eigenvalues of the Jacobian $v^\prime(\theta^*, \psi^*)$ at a stationary point $(\theta^*, \psi^*)$ have negative real-part, the continuous system converges locally to $(\theta^*, \psi^*)$ with linear convergence rate. On the other hand, if $v^\prime(\theta^*, \psi^*)$ has eigenvalues with positive real-part, the continuous system 
is not locally convergent. If all eigenvalues have zero real-part, 
it can be convergent, divergent or neither, but if it is convergent, it will generally converge with a sublinear rate.

For simultaneous gradient descent linear convergence can be achieved if and only if all eigenvalues of the Jacobian of the gradient vector field $v(\theta, \psi)$
have negative real part \cite{DBLP:conf/nips/MeschederNG17}.
This situation was also considered by \citet{DBLP:conf/nips/NagarajanK17}
who examined the asymptotic case of step sizes $h$ that go to $0$ and proved local convergence for absolutely
continuous generator and data distributions under certain regularity assumptions.

\subsection{The Dirac-GAN}\label{sec:simple-example}
\epigraph{Simple experiments, simple theorems are the building blocks that help us understand more complicated systems.}
{\textit{Ali Rahimi - Test of Time Award speech, NIPS 2017}}

\begin{figure}[t]
\begin{subfigure}{0.49\linewidth}
\resizebox{\linewidth}{!}{\begin{tikzpicture}
\begin{axis}[
  no markers, domain=-0.4:0.4,
  xmin=-0.5, xmax=0.5,
  ymin=-2, ymax=5,
  samples=400,
  axis lines*=center, xlabel=$x$, ylabel=$y$,
  every axis y label/.style={at=(current axis.above origin),anchor=south},
  every axis x label/.style={at=(current axis.right of origin),anchor=west},
  height=8cm, width=12cm,
  xtick={4,6.5}, ytick=\empty,
  enlargelimits=false, clip=true, axis on top,
  grid = major
  ]
  \addplot[const plot, fill=cyan!20, very thick,draw=cyan!50!black] coordinates {(-0.01, 0) (-0.01, 5) (0.01, 5) (0.01, 0)} ;
  \draw (axis cs:-0.1,3.5) node[black] {$p_{\mathcal D} = \delta_0$};
  \addplot[const plot, fill=cyan!20, very thick,draw=cyan!50!black] coordinates {(0.20, 0) (0.20, 5) (0.22, 5) (0.22, 0)};
  \draw (axis cs:0.3,3.5) node[black] {$p_\theta = \delta_\theta$};

  \addplot [very thick, orange!50!black] {linear(0, 4.)} node[above] {$D_\psi(x)$};
  
  \draw[->, ultra thick, dashed] (axis cs:0.18, 2.5) -- (axis cs:0.1, 2.5);
  \draw[->, ultra thick, dashed] (axis cs:0.3, 2.) -- (axis cs:0.3, 3.);
  
\end{axis}

\end{tikzpicture}}
\caption{$t = t_0$}
\label{fig:counterexample-t0}
\end{subfigure}
\begin{subfigure}{0.49\linewidth}
\resizebox{\linewidth}{!}{\begin{tikzpicture}
\begin{axis}[
  no markers, domain=-0.4:0.4,
  xmin=-0.5, xmax=0.5,
  ymin=-2, ymax=5,
  samples=400,
  axis lines*=center, xlabel=$x$, ylabel=$y$,
  every axis y label/.style={at=(current axis.above origin),anchor=south},
  every axis x label/.style={at=(current axis.right of origin),anchor=west},
  height=8cm, width=12cm,
  xtick={4,6.5}, ytick=\empty,
  enlargelimits=false, clip=true, axis on top,
  grid = major,
  ]
  \addplot[const plot, fill=cyan!20, very thick,draw=cyan!50!black] coordinates {(-0.01, 0) (-0.01, 5) (0.01, 5) (0.01, 0)} ;
  \draw (axis cs:-0.1,3.5) node[black] {$p_{\mathcal D} = \delta_0$};
  \draw (axis cs:0.1,3.5) node[black] {$p_\theta = \delta_\theta$};
  \addplot [very thick, orange!50!black] {linear(0, 10.)} node[above] {$D_\psi(x)$};
  \draw[->, ultra thick, dashed] (axis cs:-0.05, 2.5) -- (axis cs:-0.2, 2.5);

\end{axis}

\end{tikzpicture}}
\caption{$t = t_1$}
\label{fig:counterexample-t1}
\end{subfigure}
\caption{
Visualization of the counterexample showing that gradient descent based GAN optimization is not always convergent:
(a)~In the beginning, the discriminator pushes the generator towards the true data distribution and the discriminator's slope increases.
(b)~When the generator reaches the target distribution, the slope of the discriminator is largest, pushing
the generator away from the target distribution.
This results in oscillatory training dynamics that never converge.
}
\label{fig:counterexample}
\vspace{-0.5cm}
\end{figure}
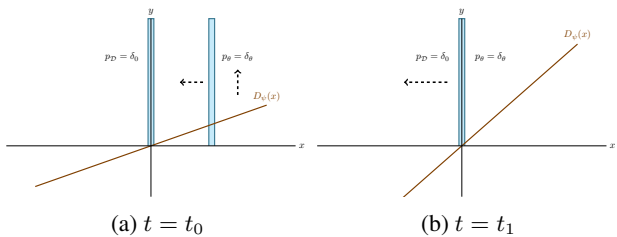

In this section, we describe a simple yet prototypical
counterexample which shows that in the general case unregularized
GAN training is neither locally nor globally convergent.

\begin{definition}
The \emph{Dirac-GAN} consists of a (univariate) generator distribution
$p_{\theta}=\delta_{\theta}$ and a linear discriminator 
$D_{\psi}(x)=\psi\cdot x$. The true data distribution $p_{\mathcal{D}}$ is
given by a Dirac-distribution concentrated at $0$.
\end{definition}
Note that for the Dirac-GAN, both the generator and the discriminator have exactly one parameter.
This situation is visualized in Figure~\ref{fig:counterexample}.
In this setup, the GAN training objective \eqref{eq:gan-loss} is given by
\begin{equation}\label{eq:training-objective}
L(\theta,\psi)=f(\psi\theta) + f(0)
\end{equation}
While using linear discriminators might appear restrictive,
the class of linear discriminators
is in fact as powerful as the class of all real-valued functions for this example:
when we use $f(t)=-\log(1 + \exp(-t))$ and we take the supremum over $\psi$ in \eqref{eq:training-objective},
we obtain (up to scalar and additive constants) the Jensen-Shannon divergence between $p_\theta$ and $p_{\mathcal D}$.
The same holds true for the Wasserstein-divergence, when we use $f(t) = t$ and put a Lipschitz constraint
on the discriminator (see Section~\ref{sec:wgan}).

We show that the training dynamics of GANs do not converge
in this simple setup. 
\begin{restatable}{lemma}{lemmaganeigenvalues}\label{lemma:gan-eigenvalues}
The unique equilibrium point of the training objective in \eqref{eq:training-objective}
is given by $\theta=\psi=0$. Moreover, the Jacobian of the gradient
vector field at the equilibrium point has the two eigenvalues
$\pm f^{\prime}(0)\:i$  which are both on
the imaginary axis.
\end{restatable}

We now take a closer look at the training dynamics produced by various algorithms for training the Dirac-GAN.
First, we consider the (idealized) continuous system in \eqref{eq:gan-continuous}:
while Lemma~\ref{lemma:gan-eigenvalues} shows that the continuous system
is generally not linearly convergent to the equilibrium point, it could in principle converge with a 
sublinear convergence rate.
However, this is not the case as the next lemma shows:

\begin{restatable}{lemma}{lemmacontinuouscase}\label{lemma:continuous-case}
 The integral curves of the gradient vector field $v(\theta, \psi)$
 do not converge to the Nash-equilibrium.
 More specifically, every integral curve $(\theta(t), \psi(t))$ of the gradient vector field
 $v(\theta, \psi)$ satisfies $\theta(t)^2 + \psi(t)^2 = const$
for all $t \in [0, \infty)$.
\end{restatable}

Note that our results do not contradict the results of 
\citet{DBLP:conf/nips/NagarajanK17} and \citet{DBLP:conf/nips/HeuselRUNH17}:
our example violates Assumption~IV in \citet{DBLP:conf/nips/NagarajanK17} that the support
of the generator distribution is equal to the support of the true data distribution near 
the equilibrium.
It also violates the
assumption\footnote{%
This assumption is usually even violated by Wasserstein-GANs,
as the optimal discriminator parameter vector 
as a function of the current generator parameters can have discontinuities near the Nash-equilibrium. 
See Section~\ref{sec:wgan} for details.}
in \citet{DBLP:conf/nips/HeuselRUNH17} that the optimal discriminator parameter vector
is a continuous function of the current generator parameters. 
In fact, unless $\theta=0$, there
is not even an optimal discriminator parameter for the Dirac-GAN.
Indeed, we found that two-time scale updates as suggested by \citet{DBLP:conf/nips/HeuselRUNH17} do not help
convergence towards the Nash-equilibrium (see Figure~\ref{fig:ttur} in the supplementary material).
However, our example seems to be a prototypical situation for (unregularized) GAN training
which usually deals with distributions that are concentrated on lower dimensional manifolds \cite{DBLP:journals/corr/ArjovskyB17}.

\begin{figure}[t]
\centering
\begin{subfigure}{0.49\linewidth}
\centering
\includegraphics[width=\linewidth]{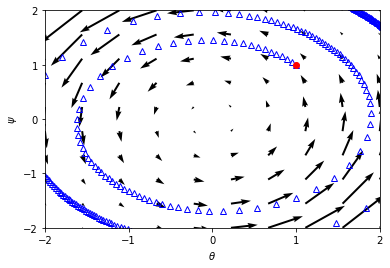}
\caption{SimGD}\label{fig:simgd}
\end{subfigure}
\begin{subfigure}{0.49\linewidth}
\centering
\includegraphics[width=\linewidth]{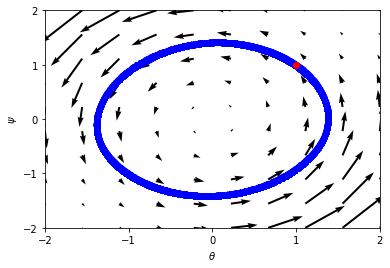}
\caption{AltGD}\label{fig:altgd}
\end{subfigure}
\caption{Training behavior of the Dirac-GAN.
The starting iterate is marked in red.}
\vspace{-0.5cm}
\end{figure}

We now take a closer look at the \emph{discretized system}.
\begin{restatable}{lemma}{lemmasimgd}
For simultaneous gradient descent, the Jacobian of the update operator $F_{h}(\theta,\psi)$
  has eigenvalues $\lambda_{1/2}=1 \pm hf^{\prime}(0)i$
 with absolute values $\sqrt{1+h^{2}f^{\prime}(0)^{2}}$
  at the Nash-equilibrium. Independently of the learning rate, simultaneous gradient descent is therefore not stable near the equilibrium.
Even stronger, for every initial condition and learning rate $h >0$, the norm of the iterates $(\theta_k, \psi_k)$ obtained by simultaneous gradient descent 
is monotonically increasing.
\end{restatable}
The behavior of simultaneous gradient descent for our example problem is visualized
in Figure~\ref{fig:simgd}. 

Similarly, for alternating gradient descent we have
\begin{restatable}{lemma}{lemmaaltgd}\label{lemma:altgd}
For alternating gradient descent with $n_{g}$
  generator and $n_{d}$
  discriminator updates, the Jacobian of the update operator $F_{h}(\theta,\psi)$
  has eigenvalues
\begin{equation}
 \lambda_{1/2} = 
  1  - \frac{\alpha^2}{2}
 \pm \sqrt{\left(
 1  -  \frac{\alpha^2}{2}
 \right)^2 - 1}.
\end{equation}
with $\alpha := \sqrt{n_g n_d} hf^\prime(0)$.
For $ \alpha \leq 2$, all eigenvalues are hence on the unit circle.
Moreover for $\alpha > 2$, there are eigenvalues outside the unit circle. 
\end{restatable}

Even though Lemma~\ref{lemma:altgd} shows that alternating gradient descent does not converge linearly to the Nash-equilibrium,
it could in principle converge with a sublinear convergence rate.
However, this is very unlikely because -- as Lemma~\ref{lemma:continuous-case} shows -- even the continuous system does not converge.
Indeed, we empirically found that alternating gradient descent oscillates in stable cycles around the equilibrium and shows no sign of convergence
(Figure~\ref{fig:altgd}).

\subsection{Where do instabilities come from?}\label{sec:explanations}
Our simple example shows that naive gradient based GAN optimization does not always converge to the equilibrium point.
To get a better understanding of what can go wrong for more complicated GANs,
it is instructive to analyze these instabilities in depth for this simple example problem.

To understand the instabilities, we have to take a closer look at the oscillatory behavior that GANs exhibit both
for the Dirac-GAN and for more complex systems.
An intuitive explanation for the oscillations is given in Figure~\ref{fig:counterexample}: when the generator is far from
the true data distribution, the discriminator pushes the generator towards the true data distribution. At the same time, the discriminator becomes
more certain,
which increases the discriminator's slope (Figure~\ref{fig:counterexample-t0}).
Now, when the generator reaches the target distribution (Figure~\ref{fig:counterexample-t1}), the slope of the discriminator is largest, pushing
the generator away from the target distribution. 
As a result, the generator moves away again from the true data distribution and the discriminator has to change its slope
from positive to negative.
After a while, we end up with a similar situation as in the beginning of training, only on the other side of the true data distribution.
This process repeats indefinitely and does not converge.

Another way to look at this is to consider the local behavior of the training algorithm near the Nash-equilibrium.
Indeed, near the Nash-equilibrium, there is nothing that pushes the discriminator towards having zero slope
on the true data distribution.
Even if the generator is initialized \emph{exactly} on the target distribution,
there is no incentive for the discriminator to move to the equilibrium discriminator.
As a result, training is unstable near the equilibrium point.

This phenomenon of discriminator gradients orthogonal to the data distribution can also arise for more complex examples:
as long as the data distribution is concentrated on a low dimensional manifold and the class of discriminators
is big enough, there is no incentive for the discriminator to produce zero gradients orthogonal to the tangent space
of the data manifold and hence converge to the equilibrium discriminator. Even if the generator produces \emph{exactly}
the true data distribution, there is no incentive for the discriminator to produce zero gradients orthogonal to the tangent space.
When this happens, the discriminator does not provide useful gradients for the generator orthogonal to the data distribution and 
the generator does not converge.

Note that these instabilities can only arise if the true data distribution is concentrated on a lower dimensional manifold.
Indeed, \citet{DBLP:conf/nips/NagarajanK17} showed that - under some suitable assumptions - gradient descent based GAN optimization is locally convergent
for absolutely continuous distributions. 
Unfortunately, this assumption may not be satisfied for data distributions like natural images to which GANs are commonly applied \cite{DBLP:journals/corr/ArjovskyB17}. 
Moreover, even if the data distribution is absolutely continuous but concentrated along some lower dimensional manifold, the eigenvalues
of the Jacobian of the gradient vector field will be very close to the imaginary axis, resulting in a highly ill-conditioned problem.
This was observed by \citet{DBLP:conf/nips/MeschederNG17} who examined the spectrum of the Jacobian for 
a data distribution given by a circular mixture of Gaussians with small variance.

\begin{figure}[t]
\begin{subfigure}[t]{0.45\linewidth}
\includegraphics[width=\linewidth]{img/altgd1/gan}
\vspace{-15pt}
\caption{Standard GAN}\label{fig:std-gan}
\end{subfigure}
\begin{subfigure}[t]{0.45\linewidth}
\includegraphics[width=\linewidth]{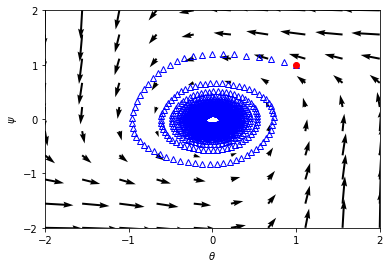}
\vspace{-15pt}
\caption{Non-saturating GAN}\label{fig:ns-gan}
\end{subfigure}

\begin{subfigure}[t]{0.45\linewidth}
\includegraphics[width=\linewidth]{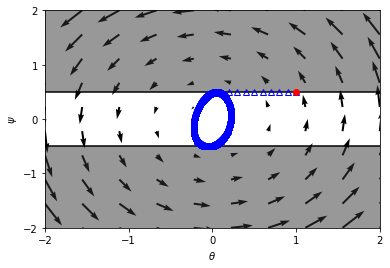}
\vspace{-15pt}
\caption{WGAN ($n_{d}=5$)}\label{fig:wgan}
\end{subfigure}
\begin{subfigure}[t]{0.45\linewidth}
\includegraphics[width=\linewidth]{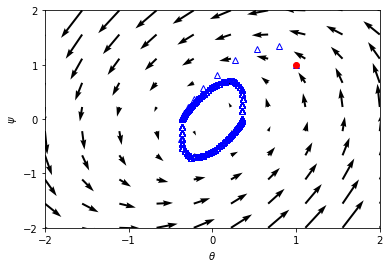}
\vspace{-15pt}
\caption{WGAN-GP ($n_{d}=5$)}\label{fig:wgan-gp}
\end{subfigure}

\begin{subfigure}[t]{0.45\linewidth}
\includegraphics[width=\linewidth]{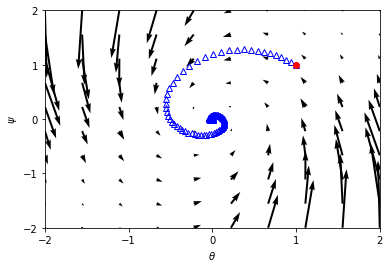}
\vspace{-15pt}
\caption{Consensus optimization}\label{fig:consensus}
\end{subfigure}
\begin{subfigure}[t]{0.45\linewidth}
\includegraphics[width=\linewidth]{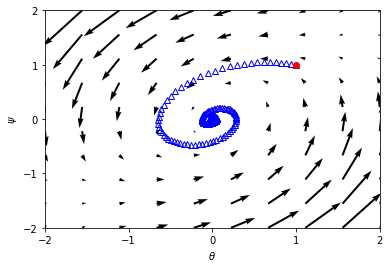}
\vspace{-15pt}
\caption{Instance noise}\label{fig:instnoise}
\end{subfigure}

\begin{subfigure}[t]{0.45\linewidth}
\includegraphics[width=\linewidth]{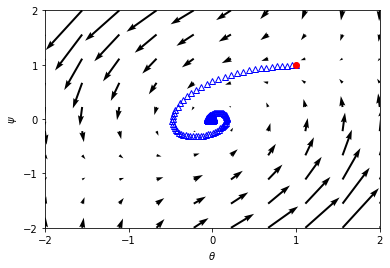}
\vspace{-15pt}
\caption{Gradient penalty}\label{fig:gradpen}
\end{subfigure}
\begin{subfigure}[t]{0.45\linewidth}
\includegraphics[width=\linewidth]{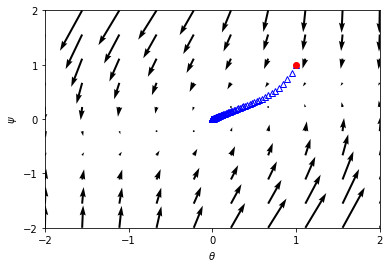}
\vspace{-15pt}
\caption{Gradient penalty (CR)}\label{fig:gradpen-critical}
\end{subfigure}
\caption{Convergence properties of different GAN training algorithms using alternating
gradient descent with recommended number of discriminator updates
per generator update ($n_d=1$ if not noted otherwise).
The shaded area in Figure~\ref{fig:wgan} visualizes the set of forbidden values for 
the discriminator parameter $\psi$.
The starting iterate is marked in red.
}
\vspace{-0.5cm}
\end{figure}

\section{Regularization strategies}\label{sec:regularization}
As we have seen in Section~\ref{sec:main}, unregularized GAN training 
does not always converge to the Nash-equilibrium.
In this section, we discuss how several regularization
techniques that have recently been proposed, influence convergence
of the Dirac-GAN.

Interestingly, we also find that the non-saturating loss proposed in the original
GAN paper \cite{DBLP:conf/nips/GoodfellowPMXWOCB14} leads to 
convergence of the continuous system, albeit with an extremely slow convergence rate.
A more detailed discussion and
an analysis of consensus optimization \cite{DBLP:conf/nips/MeschederNG17} 
can be found in the supplementary material.

\subsection{Wasserstein GAN}\label{sec:wgan}
The two-player GAN game can be interpreted as minimizing a probabilistic
divergence between the
true data distribution and the distribution
produced by the generator \cite{DBLP:conf/nips/NowozinCT16,DBLP:conf/nips/GoodfellowPMXWOCB14}.
This divergence is obtained by considering the best-response strategy for 
the discriminator, resulting in an objective function that only contains the 
generator parameters.
Many recent regularization techniques for GANs are based on the observation
\cite{DBLP:journals/corr/ArjovskyB17} that
this divergence may be discontinuous with respect to the parameters of the generator or may
even take on infinite values if the support of the data distribution and the
generator distribution do not match.

To make the divergence continuous with respect to the parameters of the generator,
Wasserstein GANs (WGANs) \citet{DBLP:journals/corr/ArjovskyCB17}
replace the Jensen-Shannon divergence used in the original
derivation of GANs \cite{DBLP:conf/nips/GoodfellowPMXWOCB14} with the Wasserstein-divergence.
As a result, \citet{DBLP:journals/corr/ArjovskyCB17} propose to use $f(t) = t$ and
restrict the class of discriminators to Lipschitz
continuous functions with Lipschitz constant equal to some $g_0 > 0$. While
a WGAN converges if the discriminator
is always trained until convergence, in practice WGANs are usually
trained by running only a fixed finite number of discriminator updates
per generator update. However, near the Nash-equilibrium the optimal discriminator
parameters can have a discontinuity as a function of the current generator parameters:
for the Dirac-GAN, the optimal discriminator has to move from $\psi=-1$ to $\psi=1$ when $\theta$ changes signs.
As the gradients get smaller near the equilibrium point, the gradient
updates do not lead to convergence for the discriminator. Overall,
the training dynamics are again determined by the Jacobian of the
gradient vector field near the Nash-equilibrium:
\begin{restatable}{lemma}{lemmawgan}
 A WGAN trained with simultaneous or alternating gradient descent with a fixed number of discriminator updates
 per generator update
 and a fixed learning rate $h>0$ does generally not converge to the
 Nash equilibrium for the Dirac-GAN.
\end{restatable}

The training behavior of the WGAN is visualized in Figure~\ref{fig:wgan}.
We stress that this analysis only holds if the discriminator 
is trained with a fixed number of discriminator updates
(as it is usually done in practice).
  More careful training that ensures that the discriminator is 
kept exactly optimal or two-timescale training \cite{DBLP:conf/nips/HeuselRUNH17} might be able to
ensure convergence for WGANs.

The convergence properties of WGANs were also
considered by \citet{DBLP:conf/nips/NagarajanK17} who showed that even for absolutely 
continuous densities and infinitesimal learning rates, WGANs are not always 
locally convergent.

We also found that WGAN-GP~\cite{DBLP:conf/nips/GulrajaniAADC17}
does not converge for the Dirac-GAN (Figure~\ref{fig:wgan-gp}).
Please see the supplementary material for details.\footnote{%
Despite these negative results, WGAN-GP has been successfully applied in practice
\cite{DBLP:conf/nips/GulrajaniAADC17,karras2017progressive} and we leave
a theoretical analysis of these empirical results to future research.} 

\subsection{Instance noise}\label{sec:instance-noise}
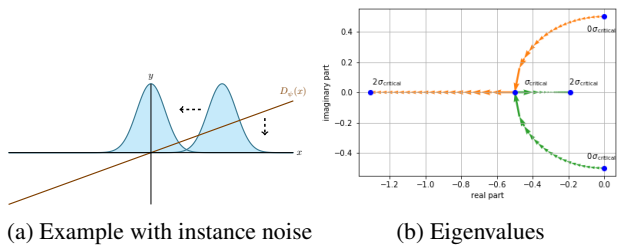
\begin{figure}[t]
\centering
\begin{subfigure}[t]{0.49\linewidth}
\centering
\resizebox{\linewidth}{!}{\begin{tikzpicture}
\begin{axis}[
  no markers, domain=-1:1, samples=400,
  axis lines*=center, xlabel=$x$, ylabel=$y$,
  every axis y label/.style={at=(current axis.above origin),anchor=south},
  every axis x label/.style={at=(current axis.right of origin),anchor=west},
  height=6cm, width=12cm,
  xtick={4,6.5}, ytick=\empty,
  enlargelimits=false, clip=false, axis on top,
  grid = major
  ]
  \addplot [fill=cyan!20, draw=none] {gauss(0,0.1)} \closedcycle;
  \addplot [fill=cyan!20, draw=none] {gauss(0.5,0.1)} \closedcycle;
  \addplot [very thick,cyan!50!black] {gauss(0.,0.1)};
  \addplot [very thick,cyan!50!black] {gauss(0.5,0.1)};
  \addplot [very thick, orange!50!black] {linear(0, 3)} node[above] {$D_\psi(x)$};

  \draw[->, ultra thick, dashed] (axis cs:0.35, 2.5) -- (axis cs:0.2, 2.5);
  \draw[->, ultra thick, dashed] (axis cs:0.8, 2.) -- (axis cs:0.8, 1.);
\end{axis}

\end{tikzpicture}}
\caption{Example with instance noise}
\label{fig:instance-noise-setup}
\end{subfigure}
\begin{subfigure}[t]{0.49\linewidth}
\centering
 \includegraphics[width=\linewidth]{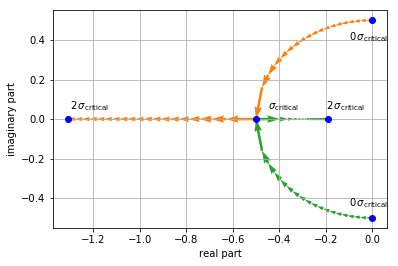}
 \caption{Eigenvalues}
 \label{fig:instance-noise-eigval}
\end{subfigure}
\caption{Dirac-GAN with instance noise.
While unregularized GAN training is inherently unstable, instance noise can stabilize it:
(a) Near the Nash-equilibrium, the discriminator is pushed towards the zero discriminator.
(b) As we increase the noise level $\sigma$ from $0$ to $\sigma_\mathrm{critical}$, 
the real part of the eigenvalues at the equilibrium point becomes negative
and the absolute value of the imaginary part becomes smaller.
For noise levels bigger than $\sigma_\mathrm{critical}$ all eigenvalues are real-valued and
GAN training hence behaves like a normal optimization problem.}
\label{fig:instance-noise}
\vspace{-0.5cm}
\end{figure}

A common technique to stabilize GANs is to add \emph{instance noise} \cite{DBLP:journals/corr/SonderbyCTSH16,DBLP:journals/corr/ArjovskyB17}, i.e.
independent Gaussian noise, to the data points.
While the original motivation was to make the probabilistic divergence
between data and generator distribution
well-defined for distributions that do not have common support, this does not clarify the effects of instance noise on the \emph{training algorithm} itself
and its ability to find a Nash-equilibrium. Interestingly, however, it was recently shown \cite{DBLP:conf/nips/NagarajanK17} that in the case
of absolutely continuous distributions, gradient descent based GAN optimization is - under suitable assumptions - locally convergent.

Indeed, for the Dirac-GAN we have:
\begin{restatable}{lemma}{lemmainstancenoise}\label{lemma:instance-noise}
When using Gaussian instance noise with standard deviation $\sigma$, the eigenvalues of the Jacobian of the gradient vector field
are given by
\begin{equation}
 \lambda_{1/2}=f^{\prime\prime}(0)\sigma^{2} \pm \sqrt{f^{\prime\prime}(0)^{2}\sigma^{4}-f^{\prime}(0)^{2}}.
\end{equation}
In particular, all eigenvalues of the Jacobian have negative real-part at the Nash-equilibrium if $f^{\prime\prime}(0) < 0$ and $\sigma > 0$.
Hence, simultaneous and alternating gradient descent are both locally convergent 
for small enough learning rates.
\end{restatable}

Interestingly, Lemma~\ref{lemma:instance-noise} shows that there is a critical noise level given by
$\sigma_{\text{critical}}^2 = |f^{\prime}(0)|/ |f^{\prime\prime}(0)|$.
If the noise level is smaller than the critical noise level, the eigenvalues of the Jacobian
have non-zero imaginary part which results in a rotational component in the gradient vector field
near the equilibrium point.
If the noise level is larger than the critical noise level,
all eigenvalues of the Jacobian become real-valued and the rotational component in the gradient vector field
disappears. The optimization problem is best behaved when we select $\sigma = \sigma_{\text{critical}}$:
in this case we can even achieve quadratic convergence for $h = |f^\prime(0)|^{-1}$.
The effect of instance noise on the eigenvalues is visualized in Figure~\ref{fig:instance-noise-eigval}, which shows
the traces of the two eigenvalues as we increase $\sigma$ from $0$ to $2\sigma_{\text{critical}}$.

Figure~\ref{fig:instnoise} shows the training behavior of the GAN with instance noise, showing that instance noise
indeed creates a strong radial component in the gradient vector field which makes the training algorithm converge.

\subsection{Zero-centered gradient penalties}\label{sec:gradient-penalty}
Motivated by the success of instance noise to make the $f$-divergence 
between two distributions well-defined,
\citet{DBLP:conf/nips/RothLNH17} derived a local approximation to instance noise
that results in a zero-centered\footnote{%
In contrast to the gradient regularizers used in WGAN-GP and DRAGAN which
are not zero-centered.}
gradient penalty
for the discriminator.

For the Dirac-GAN, a penalty on the squared norm of the gradients of the discriminator
(no matter where) results in the regularizer
\begin{equation}
R(\psi)=\frac{\gamma}{2}\psi^{2}.
\end{equation}
This regularizer does not include the weighting terms considered by \citet{DBLP:conf/nips/RothLNH17}.
However, the same analysis can also be applied to the regularizer with the 
additional weighting, yielding almost exactly the same results (see Section~\ref{sec:convergence-extensions}
of the supplementary material).

\begin{restatable}{lemma}{lemmagradientpenalty}\label{lemma:gradient-penalty}
The eigenvalues of the Jacobian of the gradient vector field for the gradient-regularized 
Dirac-GAN at the equilibrium point are given by
\begin{equation}
\lambda_{1/2}=-\frac{\gamma}{2}\pm\sqrt{\frac{\gamma^{2}}{4}-f^{\prime}(0)^{2}}.
\end{equation}
In particular,  for $\gamma>0$ all eigenvalues
have negative real part.
Hence, simultaneous and alternating gradient descent are both locally convergent 
for small enough learning rates.
\end{restatable}

Like for instance noise, there is a critical regularization parameter
$ \gamma_{\text{critical}} = 2 |f^{\prime}(0)|$
that results in a locally rotation free vector field.
A visualization of the training behavior of the Dirac-GAN with gradient penalty is shown in Figure~\ref{fig:gradpen}.
Figure~\ref{fig:gradpen-critical} illustrates the training behavior of the GAN with gradient penalty
and critical regularization (CR). In particular, we see that near the Nash-equilibrium the vector field does not have a rotational
component anymore and hence behaves like a normal optimization problem.

\section{General convergence results}\label{sec:theory}
In Section~\ref{sec:regularization} we analyzed the convergence properties of various regularization strategies
for the Dirac-GAN. 
In this section, we consider general GANs.
First, we introduce two simplified versions of the zero-centered gradient penalty proposed
by \citet{DBLP:conf/nips/RothLNH17}. We then 
show that these gradient penalties allow us to extend the convergence proof 
by \citet{DBLP:conf/nips/NagarajanK17} to the case where the generator and data distribution do not locally have
the same support.\footnote{%
Assumption IV in \citet{DBLP:conf/nips/NagarajanK17}
}
As a result, our convergence proof for the regularized training dynamics also holds for
the more realistic case where both the generator and data distributions may lie on lower dimensional manifolds.

\subsection{Simplified gradient penalties}\label{sec:method}
Our analysis suggests that the main effect of the zero-centered gradient penalties proposed by
\citet{DBLP:conf/nips/RothLNH17}
on local stability is to penalize the discriminator for deviating from the Nash-equilibrium. 
The simplest way to achieve this is to penalize the gradient on real data alone: when the generator distribution produces the true data distribution and the discriminator is equal to 0
on the data manifold, the gradient penalty ensures that the discriminator cannot create a non-zero gradient orthogonal to the data manifold without suffering a loss in the GAN game.

This leads to the following regularization term:
\begin{equation}\label{eq:discriminator}
R_1(\psi) :=
\frac{\gamma}{2} \E_{ p_{\mathcal D}(x)}
\left[
\|\nabla D_\psi(x)\|^2
\right].
\end{equation}
Note that this regularizer is a simplified version of to the regularizer derived by \citet{DBLP:conf/nips/RothLNH17}.
However, our regularizer does not contain the additional weighting terms and penalizes the discriminator gradients only on the true data distribution.

We also consider a similar regularization term given by
\begin{equation}
R_2(\theta, \psi) :=
\frac{\gamma}{2} \E_{p_\theta(x)}
\left[
\|\nabla D_\psi(x)\|^2
\right]
\end{equation}
where we penalize the discriminator gradients on the current generator distribution
instead of the true data distribution.

Note that for the Dirac-GAN from Section~\ref{sec:main}, both regularizers reduce
to the gradient penalty from Section~\ref{sec:gradient-penalty}
whose behavior is visualized in Figure~\ref{fig:gradpen} and Figure~\ref{fig:gradpen-critical}.

\subsection{Convergence}
In this section we present convergence results for the regularized GAN-training dynamics
for both regularization terms $R_1(\psi)$ and $R_2(\psi)$ 
under some suitable assumptions.\footnote{%
Our results also hold for any convex combination of $R_1$ and $R_2$
and the regularizer with the additional weighting terms derived by
\citet{DBLP:conf/nips/RothLNH17}.
See the supplementary material for details.
}

Let $(\theta^*, \psi^*)$ denote an equilibrium point of the regularized
training dynamics.
In our convergence analysis, we consider the realizable
case, i.e. we assume that there are generator parameters
that make the generator produce the true data distribution:
\begin{assumption}{I}\label{as:realizable}
We have $p_{\theta^*} = p_{\mathcal D}$ and
$D_{\psi^*} (x) = 0$ in some local neighborhood of $\supp p_{\mathcal D}$.
\end{assumption}

Like  \citet{DBLP:conf/nips/NagarajanK17}, we assume that $f$  satisfies 
the following property:
\begin{assumption}{II}\label{as:f}
We have $f^\prime(0) \neq 0$ and $f^{\prime\prime}(0) < 0$.
\end{assumption}
An extension of our convergence proof to $f(t) = t$
(as in WGANs) can be found in the supplementary material.

The convergence proof is complicated by the fact
that for neural networks, there generally is not a single equilibrium point
$(\theta^*, \psi^*)$, but a submanifold of equivalent equilibria
corresponding to different parameterizations of the same function.
We therefore define the \emph{reparameterization manifolds}
$\mathcal M_G$ and $\mathcal M_D$.
To this end, let
\begin{equation}
h(\psi) := \E_{p_{\mathcal D}(x)}
\left[
|D_\psi(x)|^2 + 
\|\nabla_x D_\psi(x) \|^2
\right].
\end{equation}
The \emph{reparameterization manifolds} are then defined as
\begin{equation}
 \mathcal M_G := \{ \theta \mid p_\theta = p_{\mathcal D} \} 
 \quad
 \mathcal M_D := \{ \psi \mid h(\psi) = 0 \}.
\end{equation}
To prove local convergence, we have to assume some
regularity properties for $\mathcal M_G$ and $\mathcal M_D$
near the equilibrium point. 
To state these assumptions, we need
\begin{equation}
g(\theta) := 
 \E_{p_{\theta}(x)}
 \left[
 \nabla_\psi D_{\psi}(x)|_{\psi=\psi^*}
 \right].
\end{equation}

\begin{assumption}{III}\label{as:regular}
 There are $\epsilon$-balls $B_\epsilon(\theta^*)$ and $B_\epsilon(\psi^*)$
 around $\theta^*$ and $\psi^*$
 so that
 $\mathcal M_G \cap B_\epsilon(\theta^*)$
 and
 $\mathcal M_D \cap B_\epsilon(\psi^*)$
define $\mathcal C^1$- manifolds.
Moreover, the following holds:
\begin{compactenum}[(i)]
 \item if $v \in \mathbb R^n$ is not in the tangent space
 of $\mathcal M_D$ at $\psi^*$, then $\partial^2_v h(\psi^*) \neq 0$.
 \item if $w \in \mathbb R^m$ is not in the tangent space
 of $\mathcal M_G$ at $\theta^*$, then $\partial_w g(\theta^*) \neq 0$. 
\end{compactenum}
\end{assumption}
While formally similar, the two conditions in Assumption~\ref{as:regular} have
very different meanings:
the first condition is a simple regularity property that
means that the geometry of $\mathcal M_D$ can be locally described by the second derivative of $h$.
The second condition implies that the discriminator is strong enough
so that it can detect any deviation from the equilibrium generator distribution.
Indeed, this is the only point where we assume that the class of representable
discriminators is sufficiently expressive (and excludes, for example, the trivial case
$D_\psi = 0$ for all $\psi$).

We are now ready to state our main convergence result.
To this end, consider the regularized gradient vector field
\begin{equation}\label{eq:regularized-vf}
\tilde v_i(\theta, \psi)
:=
\begin{pmatrix}
- \nabla_\theta L(\theta, \psi) \\
\nabla_\psi L(\theta, \psi) - \nabla_\psi R_i(\theta, \psi)
\end{pmatrix}.
\end{equation}

\begin{restatable}{thm}{thmconvergence}\label{thm:convergence}
Assume Assumption~\ref{as:realizable}, \ref{as:f} and \ref{as:regular} hold
for $(\theta^*, \psi^*)$.
For small enough learning rates, simultaneous and
alternating gradient descent for $\tilde v_1$ and  $\tilde v_2$
are both convergent to $\mathcal M_G \times \mathcal M_D$
in a neighborhood of $(\theta^*, \psi^*)$.
Moreover, the rate of convergence is at least linear.
\end{restatable}

Theorem~\ref{thm:convergence} shows that GAN training with our gradient penalties is convergent
when initialized sufficiently close to the equilibrium point. While this does not show that the
method is globally convergent, it at least shows that near the equilibrium the method is
well-behaved.

\subsection{Stable equilibria for unregularized GAN training}
As we have seen in Section~\ref{sec:main}, unregularized GAN training does not always
converge to the Nash-equilibrium.
However, this does not rule out the existence of stable equilibria for every GAN architecture.
In Section~\ref{sec:stable-equilbria-unreg} of the supplementary material, we identify two
forms of stable equilibria that may exist for unregularized GAN training
(\emph{energy solutions} and \emph{full-rank solutions}).
However,
it is not yet clear  under what conditions such solutions exist for high dimensional data distributions.

\section{Experiments}

\paragraph{2D-Problems}
Measuring convergence for GANs is hard for high dimensional problems, because we lack
a metric that can reliably detect non-convergent behavior.
We therefore first examine the behavior of the different regularizers on simple 2D examples where
we can assess convergence using an estimate of the Wasserstein-1-distance.

To this end, we run $5$ different training algorithms on $4$ different 2D-examples
for $6$ different GAN architectures. For each method, we try both stochastic gradient descent and RMS-Prop
with $4$ different learning rates. For the $R_1$-, $R_2$- and WGAN-GP-regularizers
we try $3$ different regularization parameters.
We train all methods for 50k iterations and report the results for the best hyperparameter setup. 
Please see the supplementary material for details.

\begin{figure}
\centering
\begin{subfigure}{0.45\linewidth}
\includegraphics[width=\linewidth]{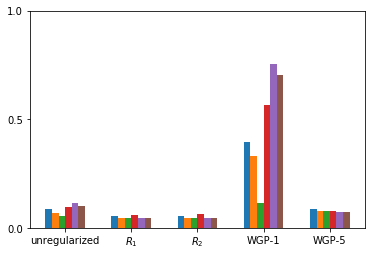}
\caption{2D Gaussian}
\end{subfigure}
\begin{subfigure}{0.45\linewidth}
\includegraphics[width=\linewidth]{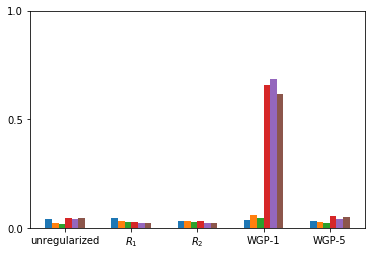}
\caption{Line segment}
\end{subfigure}

\begin{subfigure}{0.45\linewidth}
\includegraphics[width=\linewidth]{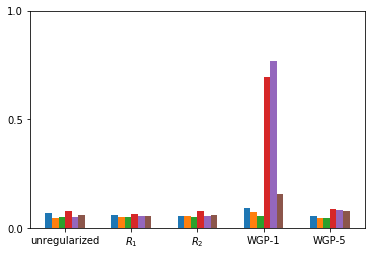}
\caption{Circle}
\end{subfigure}
\begin{subfigure}{0.45\linewidth}
\includegraphics[width=\linewidth]{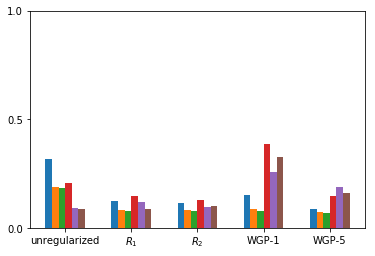}
\caption{Four line segments}
\end{subfigure}
\vspace{-0.1cm}
\caption{Wasserstein-1-distance to true data distribution for $4$
different 2D-data-distributions, $6$ different architectures (small bars) and $5$ different training methods.
Here, we abbreviate WGAN-GP with $1$ and $5$ discriminator update(s) per generator update as WGP-1 and WGP-5.
}
\label{fig:2D-examples-Wasserstein1}
\vspace{-0.5cm}
\end{figure}

The results are shown in Figure~\ref{fig:2D-examples-Wasserstein1}.
We see that the $R_1$- and $R_2$-regularizers perform similarly and they achieve
slightly better results than unregularized training or training with WGAN-GP.
In the supplementary material we show that the $R_1$- and $R_2$-regularizers find solutions where the discriminator is $0$
in a neighborhood of the true data distribution, whereas
unregularized training and WGAN-GP converge to \emph{energy solutions}.

\paragraph{Images}
To test how well the gradient penalties from Section~\ref{sec:method}
perform on more complicated tasks, we train convolutional GANs
on a variety of datasets, including a generative model for all $1000$
Imagenet classes and a generative model for the celebA-HQ dataset \cite{karras2017progressive} at  resolution $1024 \times 1024$.
While we find that unregularized GAN training quickly leads to mode-collapse for these problems,
our simple $R_1$-regularizer enables stable training. 
Random samples from the models and more details on the experimental setup can be found in the supplementary material.

\section{Conclusion}
In this paper,
we analyzed the stability of GAN training on a simple
yet prototypical example. Due to the simplicity of the example, we
were able to analyze the convergence properties of the training dynamics 
analytically
and we showed that (unregularized)
gradient based
GAN optimization is not always
locally convergent.
Our findings also show that WGANs and WGAN-GP do not 
always lead to local convergence whereas instance noise and zero-centered gradient penalties do.
Based on our analysis, we extended our results to more general GANs and 
we proved local convergence for simplified zero-centered gradient penalties
under suitable assumptions.	
In the future, we would like to extend our theory to  the non-realizable case and 
examine the effect of finite sampling sizes on the 
GAN training dynamics.

\FloatBarrier
\section*{Acknowledgements}
We would like to thank Vaishnavh Nagarajan and Kevin Roth for insightful discussions.
We also thank Vaishnavh Nagarajan for giving helpful feedback on an early draft of this manuscript.
We thank NVIDIA for donating the GPUs for the experiments presented in the supplementary material.
This work was supported by Microsoft Research through its PhD Scholarship Programme.

\bibliography{bib/bibliography}
\bibliographystyle{icml2018}

\setcounter{section}{0}
\renewcommand{\thesection}{\Alph{section}}
\twocolumn[
\icmltitle{Which Training Methods for GANs do actually Converge?\\Supplementary Material}




\vskip 0.3in
]


\section{Preliminaries}
In this section we first summarize some results from the theory of discrete dynamical systems.
We also prove a discrete version of a basic convergence theorem for continuous dynamical systems  from \citet{DBLP:conf/nips/NagarajanK17}
which allows us to make statements about training algorithms for GANs for finite learning rates.
Afterwards, we summarize some results from \citet{DBLP:conf/nips/MeschederNG17} about the convergence properties
of simultaneous and alternating gradient descent.
Moreover, we state some eigenvalue bounds that were derived by \citet{DBLP:conf/nips/NagarajanK17}
which we need to prove Theorem~\ref{thm:convergence} on the convergence of the regularized GAN training 
dynamics.

\subsection{Discrete dynamical systems}
In this section, we recall some basic definitions from
the theory of discrete nonlinear dynamical systems.
For a similar description of the theory
of continuous nonlinear  dynamical systems see for example \citet{khalil1996noninear}
and \citet{DBLP:conf/nips/NagarajanK17}.

In this paper, we consider continuously differentiable operators
$F: \Omega \to \Omega$
acting on an open set $\Omega \subset \mathbb R^n$.
A fixed point of $F$ is a point $\bar x \in \Omega$ such that 
$F(\bar x) = \bar x$.
We are interested in stability and convergence of the fixed point iteration $F^{(k)}(x)$ near 
the fixed point.
To this end, we first have to define what we mean by stability and local convergence:
\begin{definition} \label{def:stable-fixed-point}
Let $\bar x \in \Omega$ be a fixed point of a continuously differentiable operator
$F: \Omega \to \Omega$. We call $\bar x$
\begin{compactitem}
 \item \emph{stable}  if for every $\epsilon > 0$ there is $\delta > 0$ such that
 $\| x- \bar x\| < \delta$ implies $\|F^{(k)}(x) - \bar x\| < \epsilon$ for all $k \in \mathbb N$.
 \item \emph{asymptotically stable} if it is stable and there is $\delta > 0$ such that
 $\|x - \bar x\| < \delta$ implies that $F^{(k)}(x)$ converges to $\bar x$
 \item \emph{exponentially stable} if there is $\lambda \in [0, 1)$, $\delta > 0$  and $C>0$ such that
 $\| x- \bar x\| < \delta$ implies 
 \begin{equation}\label{eq:def-exponentially-stable}
  \|F^{(k)}(x) - \bar x\| < C\|x - \bar x\| \lambda^k
 \end{equation}
  for all $k \in \mathbb N$.
\end{compactitem}
\end{definition}

If $\bar x$ is asymptotically stable fixed point of $F$, we call the algorithm obtained by iteratively 
applying $F$ \emph{locally convergent} to $\bar x$. If $\bar x$ is exponentially stable, 
we call the corresponding algorithm linearly convergent. Moreover, if $\bar x$ is exponentially stable, 
we call the infimum of all $\lambda$ so that \eqref{eq:def-exponentially-stable} holds for some $C > 0$
the \emph{convergence rate} of the fixed point iteration.

As it turns out, local convergence of fixed point iterations can be 
analyzed by examining the spectrum of the Jacobian of the fixed point operator.
We have the following central Theorem:
\begin{thm}\label{thm:fixed-point-theorem}
Let $F: \Omega \rightarrow \Omega$ be a $\mathcal C^1$-mapping on an open subset $\Omega$ of $\mathbb R^n$
and $\bar x \in \Omega$ be a fixed point of $F$. 
Assume that the absolute values of the eigenvalues of the Jacobian $F^\prime(\bar x)$ are all smaller than 1.
Then the fixed point iteration $F^{(k)}(x)$ is locally convergent to $\bar x$.
Moreover, the rate of convergence is at least linear with convergence rate
$|\lambda_{max}|$ 
where $\lambda_{max}$ denotes the eigenvalue of $F^\prime(\bar x)$ with the largest absolute value.
\end{thm}
\begin{proof}
 See \citet{bertsekas1999nonlinear}, Proposition~4.4.1.
\end{proof}

For the proof of Theorem~\ref{thm:convergence} in Section~\ref{sec:proof-convergence},
we need a generalization of Theorem~\ref{thm:fixed-point-theorem} that takes into account submanifolds
of fixed points.
The next theorem is
a discrete version of Theorem~A.4  from \citet{DBLP:conf/nips/NagarajanK17} and
we prove it in a similar way:
\begin{thm}\label{thm:fixed-point-theorem-manifold}
Let $F(\alpha, \gamma)$ define a $\mathcal C^1$-mapping that maps some domain $\Omega$ to itself.
Assume that there is a local neighborhood $U$ of $0$ such that
$F(0, \gamma) = (0, \gamma)$ for $\gamma \in U$.
Moreover, assume that all eigenvalues of $J:=\nabla_\alpha F(\alpha, 0)\mid_{\alpha=0}$ have absolute value smaller than $1$.
Then the fixed point iteration defined by $F$ is locally convergent to 
$\mathcal M := \{(0, \gamma) \mid \gamma \in U\}$ with linear convergence rate  in a neighborhood of $(0, 0)$.
Moreover, the convergence rate is $|\lambda_{\mathrm{max}}|$ with
$\lambda_\mathrm{max}$ the eigenvalue of $J$
with largest absolute value.
\end{thm}
\begin{proof}
In the following, we write $F(\alpha, \gamma) = (F_1(\alpha, \gamma), F_2(\alpha, \gamma))$, so that the 
fixed point iteration can be written as
\begin{equation}
  \alpha_{k+1} = F_1(\alpha_k, \gamma_k)
  \quad
  \gamma_{k+1} = F_2(\alpha_k, \gamma_k).
\end{equation}
We first examine the behavior of $F_1$ near $(0, 0)$. To this end, we develop $F_1$ into a Taylor-series
\begin{equation}
F_1(\alpha, \gamma)
= J \alpha + g_1(\alpha, \gamma)
\end{equation}

We first show that for any $c > 0$ we have
$\|g_1(\alpha, \gamma)\| \leq c\|\alpha\|$ sufficiently close to $(0, 0)$:
because $F_1(0, \gamma) = 0$ for all $\gamma$ close to $0$, 
$g_1(\alpha, \gamma)$ must be of the form
$g_1(\alpha, \gamma) = h_1(\alpha, \gamma) \alpha$ with $h_1(0, 0) = 0$.
This shows
that  for any $c > 0$ there is indeed an open neighborhood $V$ of $(0, 0)$ so that
$|g_1(\alpha, \gamma)| \leq c \|\alpha\|$ for all $(\alpha, \gamma) \in V$.

According to \citet{bertsekas1999nonlinear}, Proposition A 15,
 we can select for every $\epsilon >0 $ a norm $\|\cdot\|_Q$ on $\mathbb R^n$ such that
\begin{equation}
\|J\alpha \|_Q < (|\lambda_{\text{max}}| + \epsilon) \|\alpha\|_Q
\end{equation}
for $\alpha \in \mathbb R^n$ where $|\lambda_{\text{max}}|$ denotes
the eigenvalue of $J$ with the largest absolute value.

Hence, for $(\alpha, \gamma) \in V$,
\begin{multline}
\|F_1(\alpha, \gamma)\|_Q
\leq \|J\alpha\|_Q + \|g_1(\alpha, \gamma)\|_Q \\
< (|\lambda_{\text{max}}| + \epsilon + c) \|\alpha\|_Q
\end{multline}

Because we can make $c + \epsilon$ as small as we want, this shows that $\|\alpha_k\| \leq C \lambda^{k} \|\alpha_0\|$ for some
$C > 0$ and $\lambda \in [0, 1)$, if $\alpha_0$ 
and all $\gamma_l$ for $l=0, \dots, k-1$ are sufficiently close to~$0$ .
We therefore have to show that the iterates $\gamma_k$ stay in a given local neighborhood of $0$,
i.e. $\|\gamma_k \| < \delta$ for some $\delta > 0$, when $\alpha_0$ and $\gamma_0$ are initialized sufficiently close to~$0$.

To show this, we develop $F_2$ into a Taylor-series around $0$:
\begin{equation}
F_2( \alpha, \gamma)
= \gamma + g_2(\alpha, \gamma).
\end{equation}
Again, we see that $g_2$ must be of the form $g_2(\alpha, \gamma) = h_2(\alpha, \gamma)\alpha$, showing that
$\|g_2(\alpha, \gamma)\| \leq c' \|\alpha\|_Q$ for some fixed constant $c' > 0$ (note that in general $h_2(0, 0) \neq 0$).
We therefore have
\begin{multline}
\|\gamma_{k} - \gamma_0\|
\leq \sum_{l=0}^{k-1} \|g_2(\alpha_l, \gamma_l) \|
\leq \sum_{l=0}^{k-1} c' \|\alpha_{l}\|_Q \\
\leq \sum_{l=0}^{k-1} Cc' \lambda^l \|\alpha_0\|_Q
\leq \frac{Cc'}{1 - \lambda} \|\alpha_0\|_Q
\end{multline}
Hence, if we initialize $\alpha_0$ within 
$\|\alpha_0\|_Q < \tfrac{1-\lambda}{2 CC'}\delta$ and $\gamma_0$ within $\|\gamma_0\| < \tfrac{\delta}{2}$,
we have $\|\gamma_k\| < \delta$ for all $k\in \mathbb N$, concluding the proof.
\end{proof}

\subsection{Simultaneous and Alternating Gradient Descent}
In this section, we recall some results from \citet{DBLP:conf/nips/MeschederNG17}
about the convergence properties of 
simultaneous and alternating gradient descent
as algorithms for training generative adversarial networks.

Recall that simultaneous gradient descent can be described by an update operator of the form
\begin{equation}
 F_h(\theta, \psi) = \begin{pmatrix}
 \theta - h \nabla_\theta L(\theta, \psi) \\
 \psi + h \nabla_\psi L(\theta, \psi)
\end{pmatrix}
\end{equation}
where $L(\theta, \psi)$ is the GAN training objective 
defined in \eqref{eq:gan-loss}.

Similarly, alternating gradient descent can be described by an update operator of the form
 $F_h = F_{2,h} \circ F_{1, h}$ where $F_{1, h}$ and $F_{2, h}$ are given by
 \begin{align}
  F_{1, h}(\theta, \psi) & = \begin{pmatrix}
 \theta - h \nabla_\theta L(\theta, \psi) \\
 \psi 
 \end{pmatrix} \\
   F_{2, h}(\theta, \psi) & = \begin{pmatrix}
 \theta  \\
 \psi + h \nabla_\psi L(\theta, \psi)
 \end{pmatrix}.
\end{align}
Moreover, we defined the gradient vector field
\begin{equation}
  v(\theta, \psi) = 
  \begin{pmatrix}
 - \nabla_\theta L(\theta, \psi) \\
  \nabla_\psi L(\theta, \psi)
 \end{pmatrix}.
\end{equation}

To understand convergence of simultaneous and alternating
gradient descent, we have to understand when the Jacobian of the corresponding update operator
has only eigenvalues with absolute value smaller than $1$.
\begin{lemma}\label{lemma:eigval-simgd}
The eigenvalues of the Jacobian of the update operator for
simultaneous gradient descent are given by
$
\lambda = 1 + h \mu
$
with $\mu$  the eigenvalues of $v^\prime(\theta^*, \psi^*)$. 
Assume that  $v^\prime(\theta^*, \psi^*)$ has only eigenvalues with negative real part.
The eigenvalues of the Jacobian of the update operator $F_h$
for simultaneous gradient descent are then all in the unit circle if and only if
\begin{equation}\label{eq:maximum-stepsize}
h < 
\frac{1}{|\Real(\lambda)|} \, 
\frac{2}{1 + \left(\tfrac{\Imag(\lambda)}{\Real(\lambda)}\right)^2}
\end{equation}
for all eigenvalues $\lambda$ of $v^\prime(\theta^*, \psi^*)$.
\end{lemma}
\begin{proof}
For simultaneous gradient descent we have
\begin{equation}
F_h(\theta, \psi) = (\theta, \psi) + h v(\theta,\psi)
\end{equation}
and hence $F_h^\prime(\theta^*, \psi^*) = I + h v^\prime(\theta^*, \psi^*)$.
Therefore the eigenvalues are given by $\lambda = 1 + h\mu$ with 
$\mu$  the eigenvalues of $v^\prime(\theta^*, \psi^*)$. 

To see when $|\lambda| < 1$, we write $\mu = -a + ib$ with $a, b \in \mathbb R$
and $a > 0$.
Then 
\begin{equation}
|\lambda|^2 = (1  - ha)^2 + h^2 b^2
\end{equation}
which is smaller than $1$ if and only if
\begin{equation}
 h < \frac{2a}{a^2 + b^2}.
\end{equation}
Dividing both the numerator and denominator by $a^2$ shows the assertion.
\end{proof}

\begin{lemma}\label{lemma:eigval-altgd}
Assume that  $v^\prime(\theta^*, \psi^*)$ has only eigenvalues with negative real part.
For $h>0$ small enough, the eigenvalues of the Jacobian of the update operator $F_h$
for alternating gradient descent are then all in the unit circle.
\end{lemma}
\begin{proof}
The Jacobian of the update operator $F_h = F_{h, 2} \circ F_{h, 1}$ at an equilibrium
point $(\theta^*, \psi^*)$ is
\begin{equation}
F^\prime_h(\theta^*, \psi^*) 
= F^\prime_{h, 2}(\theta^*, \psi^*) \cdot F^\prime_{h, 1}(\theta^*, \psi^*).
\end{equation}
However, we have
\begin{equation}
 F^\prime_{h, i}(\theta^*, \psi^*) 
= I + h v_i^\prime(\theta^*, \psi^*)
\end{equation}
for $i\in\{1,2\}$ where 
\begin{align}
 v_1(\theta, \psi)
 & = 
\begin{pmatrix}
-\nabla_\theta L(\theta, \psi) \\
0
\end{pmatrix}
\\
v_2(\theta, \psi) 
& =
\begin{pmatrix}
0\\
\nabla_\psi L(\theta, \psi)
\end{pmatrix}
\end{align}
denote the components
of the gradient vector field. Hence
\begin{multline}
F^\prime_h(\theta^*, \psi^*) 
= I + h(v_1^\prime(\theta^*, \psi^*) + v_2^\prime(\theta^*, \psi^*)) \\ 
+ h^2  v_2^\prime(\theta^*, \psi^*)v_1^\prime(\theta^*, \psi^*)
\\
= I + h (v^\prime(\theta^*, \psi^*) + h\, R(\theta^*, \psi^*)).
\end{multline}
with $ R(\theta^*, \psi^*) := v_2^\prime(\theta^*, \psi^*)v_1^\prime(\theta^*, \psi^*)$.
For $h > 0$ small enough, all eigenvalues of $v^\prime(\theta^*, \psi^*) + h\, R(\theta^*, \psi^*)$
will be arbitrarily close to the eigenvalues of $v^\prime(\theta^*, \psi^*)$.
Because all eigenvalues of $v^\prime(\theta^*, \psi^*)$ have negative real-part, 
all eigenvalues of $F^\prime_h(\theta^*, \psi^*)$
will hence lie inside the unit circle for $h > 0$ small enough.
\end{proof}
In the proof of Theorem~\ref{thm:convergence} we will use local coordinates, i.e.
a diffeomorphism $\phi$ that maps a local neighborhood of $(\theta^*, \psi^*)$ 
to an open subset of $\mathbb R^{n + m}$.
The vector field $v$ and the update operator $F_h$ then have the following representation in
the local coordinates:
\begin{align}
F^\phi_h(\alpha)
& := \phi \circ F_h \circ \phi^{-1}(\alpha) \\
v^\phi(\alpha)
& = \phi^\prime(\theta, \psi) \cdot (v\circ \phi^{-1}(\alpha))
\end{align}
While in local coordinates, the simple relationships between
$F^\phi_h(\alpha)$ and $v^\phi(\alpha)$  
needed to prove Lemma~\ref{lemma:eigval-simgd} and Lemma~\ref{lemma:eigval-altgd}  
do not hold anymore,
the spectrum can be described in the same way:
\begin{remark}\label{remark:general-spectrum-local-coordinates}
Assume $(\theta^*, \psi^*)$ is a fixed point of $F_h$ and a stationary
point of $v$.
Let $\alpha^* = \phi(\theta^*, \psi^*)$. Then
\begin{align}
 (F^\phi_h)^\prime(\alpha^*)  &= \phi^\prime(\theta^*, \psi^*) F^\prime_h(\theta^*, \psi^*) \phi^\prime(\theta^*, \psi^*)^{-1} \\
 (v^\phi)^\prime(\alpha^*)  &= \phi^\prime(\theta^*, \psi^*) v^\prime(\theta^*, \psi^*) \phi^\prime(\theta^*, \psi^*)^{-1}
\end{align}
Hence, $(F^\phi_h)^\prime(\alpha^*)$ and $F_h^\prime(\theta^*, \psi^*)$ have the same spectrum.
The same also holds for $(v^\phi)^\prime(\alpha^*)$ and $v^\prime(\theta^*, \psi^*)$.
\end{remark}
\begin{proof}
 This follows from the chain and product rules by using the fact that
 $F_h(\theta^*, \psi^*) = (\theta^*, \psi^*)$ and $v(\theta^*, \psi^*) = 0$.
\end{proof}

As we will see in the proof of Theorem~\ref{thm:convergence}, 
Remark~\ref{remark:general-spectrum-local-coordinates} allows us to apply Theorem~\ref{thm:fixed-point-theorem-manifold} to situations where the stationary
points lie on a lower dimensional manifold instead of a space of the form
$\{0\}^k \times \mathbb R^{n + m - k}$.

\subsection{Eigenvalue bounds}
When analyzing the convergence properties of GANs, we have to analyze the spectrum
of real-valued matrices of the from
\begin{equation}
 \begin{pmatrix}
  0 & -B^\T \\
  B &  -Q 
 \end{pmatrix}
\end{equation}
with $Q$ symmetric positive definite.
To this end, we need the following important theorem from \citet{DBLP:conf/nips/NagarajanK17}
which gives explicit bounds on the real part of the eigenvalues:
\begin{thm}\label{thm:eigenvalue-bounds}
Assume $J \in \mathbb R^{(n + m) \times (n + m)}$ is of the following form:
\begin{equation}
J =
 \begin{pmatrix}
  0 & -B^\T \\
  B &  -Q
 \end{pmatrix}
\end{equation}
where $Q \in \mathbb{R}^{m \times m}$ is a symmetric positive definite matrix and $B \in \mathbb R^{m \times n}$
has full column rank.
Then all eigenvalues $\lambda$ of $J$ satisfy $\Real(\lambda) < 0$.
More precisely
\begin{compactitem}
\item if $\Imag(\lambda) = 0$
\begin{equation}
\Real(\lambda) 
\leq 
-\frac{\lambda_{\min}(Q) \lambda_{\min} (B^\T B)}
{\lambda_{\max}(Q)\lambda_{\min}(Q) +  \lambda_{\min} (B^\T B)}
\end{equation}
\item if $\Imag(\lambda) \neq 0$
\begin{equation}
\Real(\lambda) 
\leq 
-\frac{\lambda_{\min}(Q)}{2}
\end{equation}
\end{compactitem}

\end{thm}
\begin{proof}
 See \citet{DBLP:conf/nips/NagarajanK17}, Lemma G.2.
\end{proof}

In Section~\ref{sec:energy-solutions}, we need a generalization of Theorem~\ref{thm:eigenvalue-bounds}.
Using almost exactly the same proof as for Theorem~\ref{thm:eigenvalue-bounds}, we obtain
\begin{thm}\label{thm:eigenvalue-bounds-generalized}
Assume $J \in \mathbb R^{(n + m) \times (n + m)}$ is of the following form:
\begin{equation}
J =
 \begin{pmatrix}
  -P & -B^\T \\
  B &  -Q
 \end{pmatrix}
\end{equation}
where $P \in \mathbb R^{n \times n}$ 
is a symmetric positive semi-definite matrix, $Q \in \mathbb{R}^{m \times m}$ is a symmetric positive definite matrix 
and $B \in \mathbb R^{m \times n}$
has full column rank.
Then all eigenvalues $\lambda$ of $J$ satisfy $\Real(\lambda) < 0$.
\end{thm}
\begin{proof}
Let $v^\T=(a^\T, b^\T)$ denote some eigenvector of $J$ with corresponding eigenvalues
$\lambda = \lambda_r + i \lambda_i$, where $\lambda_r, \lambda_i \in \mathbb R$.
Then
\begin{equation}
\lambda_r = \frac{1}{2} \bar v^\T (J + J^\T) v
= -\bar a^\T P a - \bar b^\T Q b.
\end{equation}
Because both $P$ and $Q$ are positive semi-definite, we have $\lambda_r \leq 0$.
Because $Q$ is positive definite, it suffices to show that $b \neq 0$
to prove $\lambda_r < 0$.

Assume that $ b= 0$. Because $v$ is an eigenvector of $J$, we have $B a - Qb = \lambda b$ 
and therefore $Ba = 0$.
Because $B$ has full-column rank, this shows $a = 0$ and hence $v = 0$. However, this 
contradicts the fact that $v$ is an eigenvector of $J$.
All in all, this show that $b \neq 0$ and thus $\lambda_r \leq -\bar b^\T Q b < 0$
as required.
\end{proof}

For applying Theorems~\ref{thm:fixed-point-theorem}, we
have to show that the Jacobian of the update operator $F_h$ only has eigenvalues with absolute value smaller
than $1$. For simultaneous and alternating 
gradient descent this can be achieved (Lemma~\ref{lemma:eigval-simgd} and \ref{lemma:eigval-altgd}), if 
the Jacobian of the gradient vector field $v$ only has eigenvalues with negative real-part.
While this condition suffices to prove convergence for small learning rates, \citet{DBLP:conf/nips/MeschederNG17} showed 
that simultaneous and alternating gradient descent might still require intractably small learning rates 
if the imaginary part of the eigenvalues is large.
However, in our case we have the following simple bound on the imaginary part of the eigenvalues:
\begin{lemma}
Let
\begin{equation}
J =
 \begin{pmatrix}
  -P & -B^\T \\
  B &  -Q
 \end{pmatrix}
\end{equation}
where $P \in \mathbb R^{n \times n}$ and $Q \in \mathbb{R}^{m \times m}$ are symmetric.
All eigenvalues $\lambda$ of $J$ then satisfy
\begin{equation}
|\Imag(\lambda)| \leq \sqrt{\lambda_{\max}(B^T B)}.
\end{equation}
Note that this bound is independent from $P$ and $Q$.
\end{lemma}
\begin{proof}
Assume $v$, $\|v\| = 1$, is an eigenvector of $J$ with eigenvalue $\lambda$.
Then 
\begin{equation}
\Imag(\lambda) =  \bar v^\T J_a v.
\end{equation}
with $J_{a} := \frac{1}{2i} (J - J^\T)$.
Hence, by the Cauchy-Schwarz inequality
\begin{equation}
|\Imag(\lambda)| \leq \|v\| \|J_a v\| = \|J_a v\|.
\end{equation}
But, if $v^\T = (a^\T, b^\T)$,
\begin{equation}
\|J_a v\|^2 = b B B^\T b + a B^\T B a
\leq \lambda_{\max}(B^\T B).
\end{equation}
This shows 
\begin{equation}
|\Imag(\lambda)| \leq \sqrt{\lambda_{\max}(B^T B)}.
\end{equation}
\end{proof}

\section{Proofs for the Dirac-GAN}
This section contains the proofs for our results  from Section~\ref{sec:main} and Section~\ref{sec:regularization} 
on the properties of the Dirac-GAN.

\lemmaganeigenvalues*
\begin{proof}
The loss in (\ref{eq:training-objective}) can be rewritten as 
\begin{equation}\label{eq:proof-gan-ev-objective}
L(\theta,\psi)=f(\theta\psi)+const
\end{equation}
 It is easy to check that the gradient vector field is given by
\begin{equation}
v(\theta,\psi)=\left(\begin{array}{c}
-f^{\prime}(\theta\psi)\psi\\
f^{\prime}(\theta\psi)\theta
\end{array}\right).
\end{equation}
Because $L(\theta, 0) = L(0, \psi) = const$ for all $\theta, \psi \in \mathbb R$, 
$(\theta, \psi) = (0, 0)$ is indeed a Nash-equilibrium for the game defined by \eqref{eq:proof-gan-ev-objective}.
Because we assume $f^\prime(t) \neq 0$ for all $t \in \mathbb R$, we have
$v(\theta, \psi)= 0$ if and only if $(\theta, \psi) = (0, 0)$, showing that
$(0, 0)$ is indeed the unique Nash-equilibrium.

Moreover, the Jacobian $v^{\prime}(\theta,\psi)$ of $v$ is given by
\begin{equation}
\begin{pmatrix}
-f^{\prime\prime}(\theta\psi)\psi^{2} 
& -f^{\prime}(\theta\psi) - f^{\prime\prime}(\theta\psi)\theta\psi\\
f^{\prime}(\theta\psi) + f^{\prime\prime}(\theta\psi)\theta\psi
& f^{\prime\prime}(\theta\psi)\theta^{2}
\end{pmatrix}.
\end{equation}
Evaluating it at the Nash equilibrium $\theta=\psi=0$, we obtain 
\begin{equation}
v^{\prime}(0,0)=\left(\begin{array}{cc}
0 & -f^{\prime}(0)\\
f^{\prime}(0) & 0
\end{array}\right)
\end{equation}
which has the eigenvalues $\pm f^{\prime}(0)i$.
\end{proof}

\lemmacontinuouscase*
\begin{proof}
Let $R(\theta, \psi) := \frac{1}{2}(\theta^2 + \psi^2)$. Then
\begin{multline}
 \frac{\mathrm d}{\mathrm d t} R(\theta(t), \psi(t)) \\
 = \theta(t) v_1(\theta(t), \psi(t)) + \psi(t)  v_2(\theta(t), \psi(t))
 = 0,
\end{multline}
showing that $R(\theta, \psi)$ is indeed constant for all $t \in [0, \infty)$.

\end{proof}
\lemmasimgd*
\begin{proof}
 The first part is a direct consequence of Lemma~\ref{lemma:eigval-simgd} and Lemma~\ref{lemma:gan-eigenvalues}.
 
 To see the the norms of the iterates $(\theta_k, \psi_k)$ is monotonically increasing,
 we calculate
 \begin{multline}
  \theta_{k+1}^2 + \psi_{k+1}^2 \\ 
  = (\theta_k - h f^\prime(\theta_k\psi_k)\psi_k)^2 + (\psi_k + h f^\prime(\theta_k\psi_k)\theta_k)^2 \\
  = \theta_k^2 + \psi_k^2 + h^2 f^\prime(\theta_k\psi_k)^2 (\theta_k^2 + \psi_k^2) \\
  \geq \theta_k^2 + \psi_k^2.
 \end{multline}
\end{proof}

\lemmaaltgd*
\begin{proof}
 The update operators for alternating gradient descent are given by
\begin{align}
F_1(\theta, \psi) 
& = 
\begin{pmatrix}
  \theta - h f^\prime(\theta\psi) \psi \\
  \psi
\end{pmatrix} 
\\
F_2(\theta, \psi) 
& = 
\begin{pmatrix}
  \theta\\
  \psi + h f^\prime(\theta\psi) \theta
\end{pmatrix}.
\end{align}
Hence, the Jacobians of these operators at $0$ are given by
\begin{align}
 F_1^\prime(0, 0)& =
 \begin{pmatrix}
  1 & -h f^\prime(0) \\
  0 & 1
 \end{pmatrix}
\\
 F_2^\prime(0, 0)& =
  \begin{pmatrix}
  1 & 0 \\
  h f^\prime(0) & 1
 \end{pmatrix}.
\end{align}
As a result, the Jacobian of the combined update operator is
\begin{multline}
 (F_2^{n_d} \circ F_1^{n_g})^\prime (0, 0) = F_2^\prime(0, 0)^{n_d}\cdot F_1^\prime (0, 0)^{n_g}
 \\ = 
\begin{pmatrix}
 1 & -n_g h f^\prime(0) \\
 n_d hf^\prime(0) & -n_g n_d h^2 f^\prime(0)^2 + 1
\end{pmatrix}.
\end{multline}
An easy calculation shows that the eigenvalues of this matrix are
\begin{equation}
 \lambda_{1/2} = 
 1 - \frac{\alpha^2}{2}
 \pm \sqrt{\left(1 - \frac{\alpha^2}{2}\right)^2 - 1}
\end{equation}
with $\alpha = \sqrt{n_g n_d} h f^\prime(0)$
which are on the unit circle if and only if
$\alpha \leq 2$.

\end{proof}

\lemmawgan*
\begin{proof}
 First, consider simultaneous gradient descent. 
 Assume that the iterates $(\theta_k, \psi_k)$ converge towards the equilibrium point $(0, 0)$. 
 Note that $(\theta_{k+1}, \psi_{k+1}) \neq 0$ if $(\theta_k, \psi_k) \neq 0$. 
 We can therefore assume without loss of generality that $(\theta_k, \psi_k) \neq 0$ for all $k\in \mathbb N$.
 
 Because $\lim_{k \to \infty} \psi_k = 0$, there exists $k_0$ such that for all $k \geq k_0$
 we have $| \psi_k | < 1$. For $k \geq k_0$ we therefore have
\begin{equation}\label{eq:WGAN-proof-iterate}
\begin{pmatrix}
\theta_{k+1}\\
\psi_{k+1}
\end{pmatrix}
=
\begin{pmatrix}
1 & -h  \\
h & 1
\end{pmatrix}
\begin{pmatrix}
 \theta_k\\
 \psi_k
\end{pmatrix}.
\end{equation}
For $k \geq k_0$, the iterates are therefore given by
 \begin{equation}
\begin{pmatrix}
\theta_{k}\\
\psi_{k}
\end{pmatrix}
=
A^{k - k_0}
\begin{pmatrix}
 \theta_{k_0}\\
 \psi_{k_0}
\end{pmatrix}
\quad\text{with}\quad
A = 
\begin{pmatrix}
1 & -h  \\
h & 1
\end{pmatrix}.
\end{equation}
However, the eigenvalues of $A$ are given by
$\lambda_{1/2} = 1 \pm h i$ which both have absolute value
$\sqrt{1 + h^2} > 1$. This contradicts the assumption that $(\theta_k, \psi_k)$
converges to $(0, 0)$.

A similar argument also hold for alternating gradient descent. 
In this case, $A$ is given by
\begin{equation}\label{eq:WGAN-proof-A-altgd}
\begin{pmatrix}
1 & 0  \\
h & 1
\end{pmatrix}^{n_d}
\begin{pmatrix}
1 & -h  \\
0 & 1
\end{pmatrix}^{n_g}
=
\begin{pmatrix}
1 & -h n_g  \\
h n_d & 1 - h^2 n_g n_d
\end{pmatrix}.
\end{equation}
The eigenvalues of $A$ as in \eqref{eq:WGAN-proof-A-altgd} are given
by
\begin{equation}
 1 - \frac{h^2 n_g n_h}{2} \pm \sqrt{\left(1 - \frac{h^2 n_g n_h}{2}\right)^2 - 1}.
\end{equation}
At least one of these eigenvalues has absolute value greater or equal to $1$.
Note that for almost all initial conditions $(\theta_0, \psi_0)$, the the inner product 
between the eigenvector corresponding to the eigenvalue with modulus bigger than $1$
will be nonzero for all $k \in \mathbb N$.
Since the recursion in \eqref{eq:WGAN-proof-iterate} is linear, 
this contradicts the fact that $(\theta_k, \psi_k) \to (0, 0)$,
showing that alternating gradient descent generally does not converge to the 
Nash-equilibrium either.
\end{proof}

\lemmainstancenoise*
\begin{proof}
When using instance noise, the GAN training objective \eqref{eq:gan-loss} is given by
\begin{equation}
 \E_{\tilde\theta \sim \mathcal N(\theta, \sigma^2)}
 \left[ f(\tilde\theta\psi) \right]
 + \E_{x \sim \mathcal N(0, \sigma^2)}
 \left[ f(-x\psi) \right]
.
\end{equation}
The corresponding gradient vector field is hence given by
\begin{equation}
 \tilde v(\theta, \psi)
 = 
 \E_{\tilde\theta, x}
 \begin{pmatrix}
- \psi f^\prime(\tilde\theta\psi) \\
\tilde\theta f^\prime(\tilde\theta\psi) - x f^\prime(-x\psi)
\end{pmatrix}.
\end{equation}
The Jacobian $\tilde{v}^{\prime}(\theta,\psi)$
is therefore
\begin{equation}
 \E_{\tilde\theta, x}
\begin{pmatrix}
-f^{\prime\prime}(\tilde\theta \psi)\psi^{2}
& -f^{\prime}(\tilde\theta \psi) - f^{\prime\prime}(\tilde\theta\psi)\tilde\theta\psi\\
f^{\prime}(\tilde\theta \psi) + f^{\prime\prime}(\tilde\theta\psi)\tilde\theta \psi
& f^{\prime\prime}(\tilde\theta\psi)\tilde\theta^{2} + x^2 f(-x \psi)
\end{pmatrix}
\end{equation}
Evaluating it at $\theta=\psi=0$ yields
\begin{equation}
\tilde{v}^{\prime}(0,0)=\left(\begin{array}{cc}
0 & -f^{\prime}(0)\\
f^{\prime}(0) & 2 f^{\prime\prime}(0)\sigma^{2}
\end{array}\right)
\end{equation}
whose eigenvalues are given by
\begin{equation}
 \lambda_{1/2}=f^{\prime\prime}(0)\sigma^{2} \pm \sqrt{f^{\prime\prime}(0)^{2}\sigma^{4}-f^{\prime}(0)^{2}}.
\end{equation}

\end{proof}

\lemmagradientpenalty*
\begin{proof}
The regularized gradient vector field becomes
\begin{equation}
\tilde{v}(\theta,\psi)=\left(\begin{array}{c}
-f^{\prime}(\theta\psi)\psi\\
f^{\prime}(\theta\psi)\theta-\gamma\psi
\end{array}\right).
\end{equation}
The Jacobian $\tilde{v}^{\prime}(\theta,\psi)$ is therefore given by
\begin{equation}
\begin{pmatrix}
-f^{\prime\prime}(\theta\psi)\psi^{2} & -f^{\prime}(\theta\psi)-f^{\prime\prime}(\theta\psi)\theta\psi\\
f^{\prime}(\theta\psi)+f^{\prime\prime}(\theta\psi)\theta\psi & f^{\prime\prime}(\theta\psi)\theta^{2}-\gamma
\end{pmatrix}.
\end{equation}
Evaluating it at $\theta=\psi=0$ yields
\begin{equation}
\tilde{v}^{\prime}(0,0)=\left(\begin{array}{cc}
0 & -f^{\prime}(0)\\
f^{\prime}(0) & -\gamma
\end{array}\right)
\end{equation}
whose eigenvalues are given by
\begin{equation}
\lambda_{1/2}=-\frac{\gamma}{2} \pm \sqrt{\frac{\gamma^{2}}{4}-f^{\prime}(0)^{2}}.
\end{equation}
\end{proof}

\section{Other regularization strategies}
In this section we discuss further regularization techniques for GANs 
on our example problem that 
were omitted in the main text due to space constraints.

\subsection{Nonsaturating GAN}
Especially in the beginning of training, the discriminator can reject samples
produced by the generator with high confidence \cite{DBLP:conf/nips/GoodfellowPMXWOCB14}.
When this happens, the loss for the generator may saturate so that the generator receives almost no
gradient information anymore. 

To circumvent this problem \citet{DBLP:conf/nips/GoodfellowPMXWOCB14}
introduced a nonsaturating objective for the generator.
In nonsaturating GANs, the generator objective is replaced with\footnote{%
\citet{DBLP:conf/nips/GoodfellowPMXWOCB14} used $f(t) = -\log(1 + \exp(-t))$.
}
\begin{equation}
 \max_\theta \mathrm E_{p_\theta(x)} f(-D_\psi(x)).
\end{equation}
In our example, this is 
$\max_\theta f(-\psi \theta)$.

While the nonsaturating generator objective was originally motivated by global
stability considerations, we investigate its effect on local convergence.
A linear analysis similar to normal GANs yields
\begin{restatable}{lemma}{lemmanonsaturating}\label{lemma:nonsaturating}
 The unique Nash-equilibrium for the nonsaturating GAN on the example problem
 is given by $\theta = \psi=0$.
 The eigenvalues of the Jacobian of the gradient vector field at the equilibrium
 are $\pm f^\prime(0) i$ which are both on the imaginary axis.
\end{restatable}
\begin{proof}
 The gradient vector field for the nonsaturating GAN is given by
\begin{equation}
 v(\theta, \psi)
= \begin{pmatrix}
   -f^\prime(-\theta \psi) \psi \\
   f^\prime(\theta\psi)\theta
  \end{pmatrix}.
\end{equation}
As in the proof of Lemma~\ref{lemma:gan-eigenvalues}, we see that
$(\psi, \theta) = (0, 0)$ defines the unique Nash-equilibrium for the nonsaturating GAN.

Moreover, the Jacobian $ v^\prime(\theta, \psi)$ is
\begin{equation}
\begin{pmatrix}
   f^{\prime\prime}(-\theta \psi) \psi^2 & 
   - f^\prime(-\theta\psi) + f^{\prime\prime}(-\theta\psi) \theta\psi  \\
   f^\prime(\theta\psi) +  f^{\prime\prime}(\theta\psi)\theta \psi &
   f^{\prime\prime}(\theta\psi) \theta^2
  \end{pmatrix}.
\end{equation}
At $\theta = \psi = 0$ we therefore have
\begin{equation}
  v^\prime(0, 0)
= \begin{pmatrix}
0 & 
- f^\prime(0)   \\
f^\prime(0) &
0
\end{pmatrix}.
\end{equation}
with eigenvalues $\lambda_{1/2} = \pm f^\prime(0) i$.
\end{proof}

Lemma~\ref{lemma:nonsaturating} implies that simultaneous gradient descent is not locally convergent for a nonsaturating GAN and any
learning rate $h >0$, because the eigenvalues of the Jacobian of the corresponding update operator $F_h$ 
all have absolute value larger
than $1$ (Lemma~\ref{lemma:eigval-simgd}).
While Lemma~\ref{lemma:nonsaturating} also rules out linear convergence towards the Nash-equilibrium
in the continuous case (i.e. for $h\to 0$), the continuous training dynamics could in principle still converge with a sublinear 
convergence rate.
Indeed, we find this to be the case for the Dirac-GAN. We have
\begin{restatable}{lemma}{lemmanonsaturatingconv}
For every integral curve of the gradient vector field of the nonsaturating Dirac-GAN we have
\begin{equation}
 \frac{\mathrm d}{\mathrm d t}(\theta(t)^2 + \psi(t)^2) 
 = 2 \left[
 f^\prime(\theta\psi) - f^\prime(-\theta\psi)
 \right] \theta \psi.
\end{equation}
For concave $f$ this is nonpositive.
Moreover, for $f^{\prime\prime}(0) < 0$, the continuous training dynamics of the nonsaturating Dirac-GAN converge  
with logarithmic convergence rate.
\end{restatable}
\begin{proof}
The gradient vector field for the nonsaturating Dirac-GAN is given by
\begin{equation}
 v(\theta, \psi)
= \begin{pmatrix}
   -f^\prime(-\theta \psi) \psi \\
   f^\prime(\theta\psi)\theta
  \end{pmatrix}.
\end{equation}
Hence, we have
\begin{multline}
  \frac{\mathrm d}{\mathrm d t}(\theta(t)^2 + \psi(t)^2) 
  = v_1(\theta, \psi) \theta + v_2(\theta, \psi)\psi \\
  = 2\theta\psi \left[ f^\prime(\theta\psi) - f^\prime(-\theta\psi) \right].
\end{multline}
For concave $f$, we have
\begin{equation}
 \frac{f^\prime(\theta\psi) - f^\prime(-\theta\psi)}{2 \theta\psi} \leq 0
\end{equation}
and hence
\begin{equation}
  \frac{\mathrm d}{\mathrm d t}(\theta(t)^2 + \psi(t)^2) \leq 0.
\end{equation}

Now assume that $f^\prime(0) \neq 0 $ and $f^{\prime\prime}(0) < 0 $.

To intuitively understand why the continuous system converges with logarithmic convergence rate,
note that near the equilibrium point
we asymptotically have in polar coordinates $(\theta, \psi) = (\sqrt{w}\cos(\phi), \sqrt{w}\sin(\phi))$:
\begin{align}
\dot \phi &  = f^\prime(0) + \mathcal O(|w|^{1/2}) \\
\dot w & = 4 f^{\prime\prime}(0) \theta^2\psi^2 + \mathcal O(|\theta\psi|^4) \\
& = f^{\prime\prime}(0) w^2 \sin^2(2\phi)
+ \mathcal O(|w|^4).
\end{align}
When we ignore higher order terms, we can solve this system analytically\footnote{%
For solving the ODE we use the \emph{separation of variables}-technique
and the identity
\begin{equation}
 \int 2 \sin^2(ax) \mathrm d x = x - \frac{\sin(2 ax)}{2a}.
\end{equation}

}
for $\phi$ and $w$:
\begin{align}
  \phi(t) &  = f^\prime(0) ( t - t_0) \\
  w(t) & = \frac{2}{-f^{\prime\prime}(0)t + \frac{f^{\prime\prime}(0)}{4f^\prime(0)} \sin(4 f^\prime(0)(t - t_0)) + c}
\end{align}
The training dynamics are hence convergent with logarithmic convergence rate $\mathcal O\left(\tfrac{1}{\sqrt{t}}\right)$.

For a more formal proof, first note that $w$ is nonincreasing by the first part of the proof. Moreover, for every $\epsilon > 0$ there is
$\delta > 0$ such that for $w < \delta$:
\begin{gather}
  f^\prime(0) - \epsilon 
  \leq \dot \phi
  \leq f^\prime(0) + \epsilon \\
 \dot w  
 \leq (f^{\prime\prime}(0) \sin^2(2\phi) + \epsilon) w^2 .
\end{gather}
This implies that for every time interval $[0, T]$, $\phi(t)$ is in
\begin{equation}
 \bigcup_{k\in\mathbb Z} \left[\frac{\pi}{8} + k\frac{\pi}{2}, \frac{3\pi}{8} + k\frac{\pi}{2}\right]
\end{equation}
for $t$ in a union of intervals $Q_T \subseteq [0, T]$ with total length at least $\beta\lfloor\alpha T\rfloor$ with some constants $\alpha, \beta >0$
which are independent of $T$.

For these $t \in Q_T$ we have $\sin^2(2\phi(t)) \geq \frac{1}{2}$. 
Because $f^{\prime\prime}(0) < 0$, this shows
\begin{equation}
 \dot w(t) \leq \left (\frac{1}{2}f^{\prime\prime}(0) + \epsilon \right) w(t)^2
\end{equation}
for $t\in Q_T$ and $\epsilon$ small enough.
Solving the right hand formally yields
\begin{equation}
 w(t) \leq \frac{1}{-(\frac{1}{2}f^{\prime\prime}(0) + \epsilon)t + c}.
\end{equation}
As $w(t)$ is nonincreasing for $t \notin Q_T$ and the total length of $Q_T$ is at least
 $\beta\lfloor\alpha T\rfloor$ 
 this shows that
 \begin{equation}
 w(T) \leq \frac{1}{-(\frac{1}{2}f^{\prime\prime}(0) + \epsilon)\beta\lfloor\alpha T\rfloor + c}.
\end{equation}
The training dynamics hence converge with logarithmic convergence rate $\mathcal O\left(\tfrac{1}{\sqrt{t}}\right)$.
\end{proof}
Note that the standard choice $f(t) = -\log(1 + \exp(-t))$ is concave and satisfies
$f^{\prime\prime}(0) = -\tfrac{1}{4} < 0$.
Lemma~\ref{lemma:nonsaturating} is hence applicable and
shows that the GAN training dynamics for the standard choice of $f$
converge with logarithmic convergence rate in the continuous case.
The training behavior of the nonsaturating GAN on our example problem is visualized in Figure~\ref{fig:ns-gan}.

\subsection{Wasserstein GAN-GP}
In practice, it can be hard to enforce the Lipschitz-constraint for
WGANs. A practical solution to this problem was given by \citet{DBLP:conf/nips/GulrajaniAADC17},
who derived a simple gradient penalty with a similar effect as the Lipschitz-constraint.
The resulting training objective is commonly referred to as WGAN-GP.

Similarly to WGANs, we find that WGAN-GP does not converge for the Dirac-GAN.
A similar analysis also applies to the DRAGAN-regularizer proposed by \citet{DBLP:journals/corr/KodaliAHK17}.

The regularizer proposed by \citet{DBLP:conf/nips/GulrajaniAADC17} is given by
\begin{equation}
R(\psi)=\frac{\gamma}{2}\mathrm{E}_{\hat x}
\left(\|\nabla_{x}D_{\psi}(\hat x)\|-g_{0}\right)^{2}
\end{equation}
where $\hat x$ is sampled uniformly on the line segment between
two random points $x_1 \sim p_\theta(x_1)$, $x_2 \sim p_{\mathcal D}(x_2)$.

For the Dirac-GAN, it simplifies to
\begin{equation}
R(\psi)=\frac{\gamma}{2}(|\psi|-g_{0})^{2}
\end{equation}

The corresponding gradient vector field is given by
\begin{equation}
 \tilde v(\theta, \psi)
 = 
 \begin{pmatrix}
 - \psi \\
 \theta - \sign(\psi) \gamma ( |\psi| - g_0)
 \end{pmatrix}.
\end{equation}
Note that the gradient vector field has a discontinuity at the equilibrium point, as
the gradient vector field
takes on values with norm bigger than some fixed constant in every neighborhood of the 
equilibrium point. As a result, we have

\begin{restatable}{lemma}{lemmawgangp}
 WGAN-GP trained with simultaneous or alternating gradient descent with a fixed number of generator and discriminator updates
 and a fixed learning rate $h>0$ does not converge locally  to the
 Nash equilibrium for the Dirac-GAN.
\end{restatable}
\begin{proof}
First, consider simultaneous gradient descent. 
Assume that the iterates $(\theta_k, \psi_k)$ converge towards the equilibrium point $(0, 0)$. 
For almost all initial conditions\footnote{%
Depending on $\gamma$, $h$ and $g_0$ modulo a set of measure $0$.
}
we have $(\theta_k, \psi_k) \neq (0, 0)$ for all $k \in \mathbb N$.
This implies
\begin{equation}
 |\psi_{k+1} - \psi_{k}| = h | \theta_k - \gamma \psi_k - \sign(\psi_k) g_0|\\
\end{equation}
and hence
$
\lim_{k\to \infty} |\psi_{k+1} - \psi_{k}| = h |g_0|\neq 0
$,
showing that $(\theta_k, \psi_k)$ is not a Cauchy sequence.
This contradicts the
assumption that $(\theta_k, \psi_k)$ converges to the equilibrium point $(0, 0)$.

A similar argument also holds for alternating gradient descent.
\end{proof}
The training behavior of WGAN-GP on our example problem is visualized in Figure~\ref{fig:wgan-gp}.

As for WGANs, we stress that this analysis only holds if the discriminator 
is trained with a fixed number of discriminator updates per generator update.
Again, more careful training that ensures that the discriminator is 
kept exactly optimal or two-timescale training \cite{DBLP:conf/nips/HeuselRUNH17} might be able to
ensure convergence for WGAN-GP.

\subsection{Consensus optimization}
Consensus optimization \cite{DBLP:conf/nips/MeschederNG17} is an algorithm that attempts to 
solve the problem of eigenvalues with zero real-part by introducing a regularization term
that explicitly moves the eigenvalues to the left. The regularization term in consensus optimization is given by
\begin{multline}
 R(\theta, \psi) = \frac{\gamma}{2} \| v(\theta, \psi)\|^2 \\
 = \frac{\gamma}{2} (\| \nabla_\theta L(\theta, \psi) \|^2  + \| \nabla_\psi L(\theta, \psi) \|^2 ).
\end{multline}
As was proved by \citet{DBLP:conf/nips/MeschederNG17}, consensus optimization 
converges locally for small learning rates $h>0$ 
provided that the Jacobian $v^\prime(\theta^*, \psi^*)$ is invertible.\footnote{%
\citet{DBLP:conf/nips/MeschederNG17} considered only the case of isolated equilibrium
points. However, by applying Theorem~\ref{thm:fixed-point-theorem-manifold},
it is straightforward to generalize their proof to the case where
we are confronted with a submanifold of equivalent equilibrium points.
}

Indeed, for the Dirac-GAN we have
\begin{restatable}{lemma}{lemmaconsensus}
The eigenvalues of the Jacobian of the gradient vector field for consensus optimization
at the equilibrium point
are given by
\begin{equation}
\lambda_{1/2}=-\gamma f^\prime(0)^2 \pm i f^\prime(0)
\end{equation}
In particular, all eigenvalues	
have a negative real part $-\gamma f^\prime(0)^2$.
Hence, simultaneous and alternating gradient descent are both locally convergent using consensus optimization
for small enough learning rates.
\end{restatable}
\begin{proof}
As was shown by \citet{DBLP:conf/nips/MeschederNG17}, the Jacobian of the modified vector field $\tilde v$ at the equilibrium
 point is
 \begin{equation}
  \tilde v^{\prime}(0, 0)  = v^{\prime}(0, 0) - \gamma v^{\prime}(0, 0)^\intercal v^{\prime}(0, 0).
 \end{equation}
 In our case, this is
 \begin{equation}
   \begin{pmatrix}
    -\gamma f^\prime(0)^2 & -f^\prime(0) \\
    f^\prime(0) & - \gamma f^\prime(0)^2.
   \end{pmatrix}
 \end{equation}
A simple calculation shows that the eigenvalues of $   \tilde v^{\prime}(0, 0)$ are given by
\begin{equation}
 \lambda_{1/2} = -\gamma f^\prime(0)^2 \pm i f^\prime(0).
\end{equation}
This concludes the proof.
\end{proof}
A visualization of consensus optimization for the Dirac-GAN is given in Figure~\ref{fig:consensus}.

Unfortunately, consensus optimization has the drawback that it can introduce
new spurious points of attraction to the GAN training dynamics.
While this is usually not a problem for simple examples, it can be a problem for
more complex ones like deep neural networks.

A similar regularization term as in consensus optimization was also independently proposed by 
\citet{DBLP:conf/nips/NagarajanK17}. However, \citet{DBLP:conf/nips/NagarajanK17} proposed
to only regularize the component $\nabla_\psi L(\theta, \psi)$ of the gradient vector field 
corresponding to the discriminator parameters.
Moreover, the regularization term is only added to the generator objective to give the generator
more foresight.
It can be shown \cite{DBLP:conf/nips/NagarajanK17} that this simplified regularization term can in
certain situations also make
the training dynamics locally convergent, but  might be better behaved at stationary points of the GAN training dynamics
that do not correspond to a local Nash-equilibrium.
Indeed, a more detailed analysis shows that this simplified regularization term behaves similarly to instance noise and
gradient penalties
(which we discussed in Section~\ref{sec:instance-noise} and Section~\ref{sec:gradient-penalty}) for the Dirac-GAN.

\section{General convergence results}\label{sec:proof-convergence}
In this section, we prove Theorem~\ref{thm:convergence}.
To this end, we extend the convergence proof by \citet{DBLP:conf/nips/NagarajanK17} to
our setting. We show that by introducing the gradient penalty terms $R_i(\theta, \psi)$,
we can get rid of the assumption that the generator and data distributions locally have
the same support. As we have seen, this makes the theory applicable to more realistic cases,
where both the generator and data distributions typically lie on lower dimensional manifolds.

\subsection{Convergence proof}\label{sec:convergence-proof}
To prove Theorem~\ref{thm:convergence}, we first need to understand the local structure of 
the gradient vector field $v(\theta, \psi)$. Recall that the gradient vector field $v(\theta, \psi)$ is defined as
\begin{equation}\label{eq:general-v}
v(\theta, \psi)
:=
\begin{pmatrix}
- \nabla_\theta L(\theta, \psi) \\
\nabla_\psi L(\theta, \psi)
\end{pmatrix}
\end{equation}
with
\begin{multline}\label{eq:general-L}
L(\theta,\psi)
= \mathrm{E}_{p(z)}\left[f(D_\psi(G_\theta(z)))\right] \\
+ \mathrm{E}_{p_{\mathcal D}(x)}\left[f(-D_\psi(x))\right].
\end{multline}

\begin{lemma}\label{lemma:general-gradients-v}
The gradient of $L(\theta, \psi)$ with respect to $\theta$ is given by
\begin{multline}\label{eq:general-gradients-v-theta}
\nabla_\theta L(\theta, \psi) = 
  \E_{p(z)}\bigl[
  f^\prime(D_\psi(G_\theta(z)) \left[ \nabla_\theta G_\theta(z) \right]^\T \\
  \cdot \nabla_x D_\psi(G_\theta(z)) 
  \bigr].
\end{multline}
Similarly, the gradient of $L(\theta, \psi)$ with respect to $\psi$ is given by
\begin{multline}\label{eq:general-gradients-v-psi}
\nabla_\psi L(\theta, \psi) = 
  \E_{p_\theta(x)}\left[
  f^\prime(D_\psi(x)) \nabla_\psi D_\psi(x)
  \right] \\
  -\E_{p_{\mathcal D}(x)}\left[
    f^\prime(-D_\psi(x)) \nabla_\psi D_\psi(x)
  \right]. 
\end{multline}	
\end{lemma}
\begin{proof}
This is just the chain rule.
\end{proof}

\begin{lemma}\label{lemma:general-Jacobian}
Assume that $(\theta^*, \psi^*)$ satisfies Assumption~\ref{as:realizable}.
The Jacobian of the gradient vector field $ v(\theta, \psi)$
at $(\theta^*, \psi^*)$ is then
\begin{equation}
v^\prime(\theta^*, \psi^*)
= \begin{pmatrix}
0 & -K_{DG}^\T \\
K_{DG} & K_{DD}
\end{pmatrix}.
\end{equation}
The terms  $K_{DD}$ and $K_{DG}$ are given by
\begin{align}
\label{eq:general-Jacobian-KDD}
K_{DD}
& = 2 f^{\prime\prime}(0) \E_{p_{\mathcal D}(x)} \left[
    \nabla_\psi D_{\psi^*}(x) \nabla_\psi D_{\psi^*}(x)^\T
    \right]
\\
\label{eq:general-Jacobian-KDG}
K_{DG}
& = f^\prime(0)
    \nabla_{\theta} \E_{p_\theta(x)}\left[
    \nabla_\psi D_{\psi^*}(x)
    \right]\mid_{\theta=\theta^*}
\end{align}
\end{lemma}
\begin{proof}
First note that by the definition of $v(\theta, \psi)$ in \eqref{eq:general-v},
the Jacobian $v^\prime(\theta^*, \psi^*)$ of $v(\theta, \psi)$ is given by
\begin{equation}
\begin{pmatrix}
-\nabla^2_{\theta} L(\theta^*, \psi^*)
& -\nabla^2_{\theta, \psi} L(\theta^*, \psi^*)
\\
\nabla^2_{\theta, \psi} L(\theta^*, \psi^*) 
& \nabla^2_{\psi} L(\theta^*, \psi^*)
\end{pmatrix}.
\end{equation}

By Assumption~\ref{as:realizable}, $D_{\psi^*}(x) = 0$ in some neighborhood
of $\supp p_{\mathcal D}$. Hence, we also have $\nabla_x D_{\psi^*}(x) = 0$
and $\nabla_x^2 D_{\psi^*}(x) = 0$
for $x \in \supp p_{\mathcal D}$.
By taking the derivative of \eqref{eq:general-gradients-v-theta} with respect to
$\theta$ and using $\nabla_x D_{\psi^*}(x) = 0$ and $\nabla_x^2 D_{\psi^*}(x) = 0$
for $x \in \supp p_{\mathcal D}$ we see that
$\nabla^2_\theta L(\theta^*, \psi^*) = 0$.

To show \eqref{eq:general-Jacobian-KDD} and \eqref{eq:general-Jacobian-KDG}, simply
take the derivative of \eqref{eq:general-gradients-v-psi}
with respect to $\theta$ and $\psi$
and evaluate at it at $(\theta, \psi) = (\theta^*, \psi^*)$.
\end{proof}

We now take a closer look at the regularized vector field.
Recall that we consider the two regularization terms
\begin{align}
\label{eq:reg1}
R_1(\theta, \psi) 
& := \frac{\gamma}{2} \E_{p_{\mathcal D}(x)}
\left[
\|\nabla_x D_\psi(x)\|^2
\right]
\\
\label{eq:reg2}
R_2(\theta, \psi)
& := \frac{\gamma}{2} \E_{p_\theta(x)}
\left[
\|\nabla_x D_\psi(x)\|^2
\right].
\end{align}
As discussed in Section~\ref{sec:method}, the regularization is only applied to the discriminator.
The regularized vector field is hence given by
\begin{equation}
\tilde v(\theta, \psi)
:=
\begin{pmatrix}
- \nabla_\theta L(\theta, \psi) \\
\nabla_\psi L(\theta, \psi) - \nabla_\psi R_i(\theta, \psi)
\end{pmatrix}.
\end{equation}

\begin{lemma}\label{lemma:general-gradients-reg} 
The gradient $\nabla_{\psi} R_i(\theta, \psi)$
of the regularization terms $R_i$, $i\in\{1,2\}$, with respect to $\psi$
are
\begin{align}
\label{eq:general-gradients-reg1}
\nabla_{\psi} R_1(\theta, \psi)
& = \gamma \E_{p_{\mathcal D(x)}} \left[
\nabla_{\psi, x} D_{\psi}(x) \nabla_x D_{\psi}(x) 
\right]
\\
\label{eq:general-gradients-reg2}
\nabla_{\psi} R_2(\theta, \psi)
& = \gamma \E_{p_\theta(x)} \left[
\nabla_{\psi, x} D_{\psi}(x) \nabla_x D_{\psi}(x) 
\right].
\end{align}
\end{lemma}
\begin{proof}
These equations can be derived by taking the derivative of
\eqref{eq:reg1} and \eqref{eq:reg2} with respect to $\psi$.
\end{proof}

\begin{lemma}\label{lemma:general-Jacobian-reg}
The second derivatives $\nabla^2_{\psi} R_i(\theta^*, \psi^*)$
of the regularization terms $R_i$, $i\in\{1,2\}$, with respect to $\psi$
at $(\theta^*, \psi^*)$ are both given by
\begin{equation}
L_{DD} := \gamma \E_{p_{\mathcal D}(x)} \left[
\nabla_{\psi,x} D_{\psi^*}(x) \nabla_{\psi,x} D_{\psi^*}(x)^\T
 \right].
\end{equation}
Moreover, both regularization terms satisfy
$\nabla_{\theta,\psi} R_i(\theta^*, \psi^*) = 0$.
\end{lemma}
\begin{proof}
$\nabla^2_{\psi} R_i(\theta^*, \psi^*)$, $i \in \{1,2\}$,
can be computed by taking the
derivative of 
\eqref{eq:general-gradients-reg1}
and
\eqref{eq:general-gradients-reg2}
with respect to $\psi$ and using the fact
that $\nabla_x D_{\psi^*}(x) = 0$ in a neighborhood of $\supp p_{\mathcal D}$.

Moreover, we clearly have
$\nabla_{\theta,\psi} R_1(\theta^*, \psi^*) = 0$,
because $R_1$ does not depend on $\theta$.
To see that $\nabla_{\theta,\psi} R_2(\theta^*, \psi^*) = 0$,
take the derivative of \eqref{eq:general-gradients-reg2} with respect to $\theta$
and use the fact that $\nabla_x D_{\psi^*}(x) = 0$ and
$\nabla_x^2 D_{\psi^*}(x) = 0$ for $x \in \supp p_{\mathcal D}$.
\end{proof}

As a result, the Jacobian $\tilde v^\prime(\theta^*, \psi^*)$ of the regularized
gradient vector field at the equilibrium point is given by
\begin{equation}
\tilde v^\prime(\theta^*, \psi^*)
= \begin{pmatrix}
0 & -K_{DG}^\T \\
K_{DG} & K_{DD} - L_{DD}
\end{pmatrix}.
\end{equation}
For brevity, we define $M_{DD} := K_{DD} - L_{DD}$.

To prove Theorem~\ref{thm:convergence}, we have to show that
$\tilde v^\prime(\theta^*, \psi^*)$ is well behaved when restricting it
to the space orthogonal to the tangent space of $\mathcal M_G \times \mathcal M_D$ 
at $(\theta^*, \psi^*)$:
\begin{lemma}\label{lemma:general-MDD-regular}
Assume that Assumptions~\ref{as:f}  and \ref{as:regular} hold.
If $v \neq 0$ is not in the tangent space of $\mathcal M_D$ at $\psi^*$, then
$
\bar v^\T M_{DD} v < 0.
$
\end{lemma}
\begin{proof}
By Lemma~\ref{lemma:general-Jacobian}, we have
\begin{equation}
v^\T K_{DD} v
= 
2 f^{\prime\prime}(0) \E_{p_{\mathcal D}(x)} 
\left[
\left(
\nabla_\psi D_{\psi^*}(x)^\T v
\right)^2
\right]
\end{equation}
and by Lemma~\ref{lemma:general-Jacobian-reg}
\begin{equation}
v^\T L_{DD} v
= 
\gamma \E_{p_{\mathcal D}(x)} 
\left[
\left\|
\nabla_{x,\psi} D_{\psi^*}(x) v	
\right\|^2
\right].
\end{equation}
By Assumption~\ref{as:f}, we have $f^{\prime\prime}(0) < 0$.
Hence, $v^\T M_{DD} v \leq 0 $ and $v^\T M_{DD} v = 0$ implies 
\begin{equation}
\nabla_\psi D_{\psi^*}(x)^\T v = 0
\quad\text{and}\quad
\nabla_{x,\psi} D_{\psi^*}(x) v = 0
\end{equation}
for all $x \in \supp p_{\mathcal D}$. 

Let
\begin{equation}
h(\psi) := \E_{p_{\mathcal D}(x)}
\left[
|D_\psi(x)|^2 + 
\|\nabla_x D_\psi(x) \|^2
\right].
\end{equation}
Using the fact that $D_\psi(x) = 0$ and $\nabla_x D_\psi(x) = 0$ for $x \in \supp p_{\mathcal D}$,
we see that
the Hessian of $h(\psi)$ at $\psi^*$
is
\begin{multline}
\nabla^2_\psi h(\psi^*)
=
 2 \E_{p_{\mathcal D}(x)}
\bigl[
\nabla_\psi D_\psi(x) \nabla_\psi D_\psi(x)^\T \\
 + 
\nabla_{\psi, x} D_\psi(x) \nabla_{\psi, x} D_\psi(x)^\T 
\bigr] 
\end{multline}

The  second directional derivate $\partial^2_v h(\psi)$
is therefore
\begin{multline}
 \partial^2_v h(\psi) =
 2 \E_{p_{\mathcal D}(x)}
\bigl[
|\nabla_\psi D_\psi(x)^\T v|^2 \\
 + 
 \|\nabla_{x, \psi} D_\psi(x) v\|^2 
\bigr] = 0.
\end{multline}

By Assumption~\ref{as:regular}, this can only hold if $v$
is in the tangent space of $\mathcal M_D$ at $\psi^*$.
\end{proof}

\begin{lemma}\label{lemma:general-KDG-regular}
Assume that Assumption~\ref{as:regular} holds.
If $w \neq 0$ is not in the tangent space of $\mathcal M_G$ at $\theta^*$, then
$
K_{DG} w \neq 0.
$
\end{lemma}
\begin{proof}
By Lemma~\ref{lemma:general-Jacobian}, we have
\begin{multline}
K_{DG} w
= 
f^\prime(0)
\left[
\nabla_{\theta} \E_{p_\theta(x)}\left[
\nabla_\psi D_{\psi^*}(x)
\right]\mid_{\theta=\theta^*}
\right]
w \\
= f^\prime(0) \partial_w g(\theta).
\end{multline}
for
\begin{equation}
g(\theta) := 
 \E_{p_{\theta}(x)}
 \left[
 \nabla_\psi D_{\psi^*}(x)
 \right].
\end{equation}
By Assumption~\ref{as:regular}, this implies $K_{DG} w \neq 0$
if $w$ is not in the tangent space of $\mathcal M_G$ at $\theta^*$.
\end{proof}

We are now ready to prove Theorem~\ref{thm:convergence}:
\thmconvergence*
\begin{proof}
First note that by
Lemma~\ref{lemma:general-gradients-v} and Lemma~\ref{lemma:general-gradients-reg} 
$v(\theta, \psi) = 0$ for all
points $(\theta, \psi) \in \mathcal M_G \times \mathcal M_D$,
because $D_{\psi}(x) = 0$ and $\nabla_x D_{\psi}(x) = 0$ for 
all $x \in \supp p_{\mathcal D}$ and $\psi \in \mathcal M_D$.
Hence,  $\mathcal M_G \times \mathcal M_D$ consists only of equilibrium points of the
regularized gradient vector fields.

Let  $\mathcal T_{\theta^*}\mathcal M_G$  and $\mathcal T_{\psi^*}\mathcal M_D$ denote the tangent 
spaces of $\mathcal M_G$ and $\mathcal M_D$ at $\theta^*$ and $\psi^*$.

We now want to show that both simultaneous and alternating gradient descent are
locally convergent to $\mathcal M_G \times \mathcal M_D$ for the regularized gradient vector field $\tilde v(\theta, \psi)$.
To this end, we want to apply Theorem~\ref{thm:fixed-point-theorem-manifold}.
By choosing local coordinates $\theta(\alpha, \gamma_G)$ and $\psi(\beta, \gamma_D)$ for $\mathcal M_G$ and $\mathcal M_D$
and using Remark~\ref{remark:general-spectrum-local-coordinates}, 
we can assume without loss of generality
that
$\theta^* = 0$, $\psi^* = 0$ as well as
\begin{align}
\mathcal M_G & = \mathcal T_{\theta^*}\mathcal M_G =  \{0\}^{k} \times \mathbb R^{n -k} \\
\mathcal M_D & = \mathcal T_{\psi^*}\mathcal M_D = \{0\}^{l} \times \mathbb R^{m-l}.
\end{align}
This allows us to write\footnote{%
By abuse of notation, we simply write
$\theta = (\alpha, \gamma_G)$ and $\psi = (\beta, \gamma_D)$.
}
$\tilde v(\theta, \psi)=\tilde v(\alpha, \gamma_G, \beta, \gamma_D)$
In order to apply Theorem~\ref{thm:fixed-point-theorem-manifold},
we have to show that
$\nabla_{(\alpha, \beta)} \tilde v(\theta^*, \psi^*)$ only has eigenvalues with negative real-part.

By Lemma~\ref{lemma:general-Jacobian}, $\nabla_{(\alpha, \beta)} \tilde v(\theta^*, \psi^*)$ is of the form
\begin{equation}
\begin{pmatrix}
 0 & -\tilde K_{DG}^\T \\
 \tilde K_{DG} & \tilde K_{DD} - \tilde L_{DD}
 \end{pmatrix}
\end{equation}
where $\tilde K_{DD}$, $\tilde K_{DG}$ and $\tilde L_{DD}$ denote the 
submatrices of $K_{DD}$, $K_{DG}$ and $L_{DD}$ corresponding to the $(\alpha, \beta)$ coordinates.

We now show that $\tilde M_{DD} := \tilde K_{DD} - \tilde L_{DD}$ is negative definite and $\tilde K_{DG}$ has full column rank.

To this end, first note that
\begin{equation}
\tilde v^\T \tilde M_{DD} \tilde v
 =
v^\T M_{DD}  v
\end{equation}
with $v^\T := (\tilde v^\T, 0)$. Note that
$v \notin \mathcal T_{\psi^*} \mathcal M_D$ for $\tilde v\neq 0$.
Hence, by Lemma~\ref{lemma:general-MDD-regular} we have that
$\tilde v^\T \tilde M_{DD} \tilde v < 0$ if $\tilde v \neq 0$.
As a result, we see that $\tilde M_{DD}$ is symmetric negative definite. 

Similarly, for $w^\T := (\tilde w^\T, 0)$, the components of $ K_{DG} w$ corresponding to 
the $\beta$-coordinates are given by 
$\tilde K_{DG} \tilde w$.
Again, we have
$w \notin \mathcal T_{\theta^*} \mathcal M_G$ for $\tilde w\neq 0$.
Hence, by Lemma~\ref{lemma:general-KDG-regular} we have that
$K_{DG}  w \neq 0 $ if $\tilde w \neq 0$.
Because the components of $K_{DG} w$ corresponding to the $\gamma_D$ coordinates 
are $0$, this shows that $\tilde K_{DG} \tilde w \neq 0$.
$\tilde K_{DG}$ therefore has full column rank.

Theorem~\ref{thm:eigenvalue-bounds} now implies that all eigenvalues of $\nabla_{(\alpha, \beta)} \tilde v(\theta^*, \psi^*)$
have negative real part.
By Lemma~\ref{lemma:eigval-simgd}, Lemma~\ref{lemma:eigval-altgd} and Theorem~\ref{thm:fixed-point-theorem-manifold},
simultaneous and alternating gradient descent are therefore both
convergent to $\mathcal M_G \times \mathcal M_D$ near $(\theta^*, \psi^*)$ for small enough learning rates.
Moreover, the rate of convergence is at least linear.
\end{proof}

\subsection{Extensions}\label{sec:convergence-extensions}
In the proof of Theorem~\ref{thm:convergence} we have assumed that $f^{\prime\prime}(0) < 0$.
This excludes the function $f(t) = t$ which is used in Wasserstein-GANs.
We now show that our convergence proof extends to the case where $f(t) = t$ when
we modify Assumption~\ref{as:regular} as little bit:
\begin{remark}
 When we replace $h(\psi)$ with
\begin{equation}
\tilde h(\psi) := \E_{p_{\mathcal D}(x)}
\left[
\|\nabla_x D_\psi(x) \|^2
\right]
\end{equation}
and $\mathcal M_D$ with
$
\tilde{\mathcal M}_D := \{ \psi \mid \tilde h(\psi) = 0 \}
$
the results of Theorem~\ref{thm:convergence} still hold for $f(t) = t$.
\end{remark}
\begin{proof}
Almost everything in the proof of Theorem~\ref{thm:convergence} still holds for
these modified assumptions. The only thing that we have to show is that
$\mathcal M_G \times \mathcal M_D$ still consists only of equilibrium points
and that Lemma~\ref{lemma:general-MDD-regular} still holds in this setting.

To see the former, note that by Lemma~\ref{lemma:general-gradients-v} we still have
$\nabla_\theta L(\theta, \psi) = 0$ for $(\theta, \psi) \in \mathcal M_G \times \mathcal M_D$, because we have
$\nabla_x D_\psi(x) = 0$ for $\psi \in \mathcal M_D$ and $x \in \supp p_{\mathcal D}$.
On the other hand, for $f(t) = t$ we also have $\nabla_\psi L(\theta, \psi) = 0$
if $\theta \in \mathcal M_G$, because for 
$\theta \in \mathcal M_G$ the definition of $ \mathcal M_G$
 implies that $p_\theta = p_{\mathcal D}$ and hence, by
 Lemma~\ref{lemma:general-gradients-v},
\begin{multline}
 \nabla_\psi L(\theta, \psi) 
 = \E_{x \sim p_\mathcal D}\left[ \nabla_\psi D_\psi(x)\right] \\
 - \E_{x \sim p_{\mathcal D}}\left[ \nabla_\psi D_\psi(x)\right] 
 = 0.
\end{multline}

To see why Lemma~\ref{lemma:general-MDD-regular}
still holds, first note that for $f(t) = t$, we have $f^{\prime\prime}(0) = 0$, so that
by Lemma~\ref{lemma:general-Jacobian}
$K_{DD} = 0$.
Hence,
\begin{equation}
v^\T M_{DD} v = -v^\T L_{DD} v.
\end{equation}
We therefore have to show that $v^\T L_{DD} v \neq 0$
if $v$ is not in the tangent space of $\mathcal M_D$.

However, we have seen in the proof of Lemma~\ref{lemma:general-MDD-regular} that
\begin{equation}
 v^\T L_{DD} v
= 
\gamma \E_{p_{\mathcal D}(x)} 
\left[
\left\|
\nabla_{x,\psi} D_{\psi^*}(x) v	
\right\|^2
\right].
\end{equation}

Hence  $v^\T L_{DD} v = 0$ implies
$\nabla_{x,\psi} D_{\psi^*}(x) v = 0$ 
for $x \in \supp p_{\mathcal D}$ and thus
\begin{multline}
\partial^2_v h(\psi) =
2 \E_{p_{\mathcal D}(x)}
\bigl[
 \| \nabla_{x, \psi} D_\psi(x) v \|^2 
\bigr] = 0.
\end{multline}
By Assumption~\ref{as:regular}, this can only be the case if 
$v$ is in the tangent space of $\mathcal M_D$. 
This concludes the proof.
\end{proof}

In Section~\ref{sec:convergence-proof}, we showed that both regularizers $R_1$ and $R_2$
from Section~\ref{sec:method} make the GAN training dynamics locally convergent.
A similar, but slightly more complex regularizer was also proposed by \citet{DBLP:conf/nips/RothLNH17}
who tried to find a computationally efficient approximation to instance noise.
The regularizer proposed by \citet{DBLP:conf/nips/RothLNH17} is given by
a linear combination of $R_1$ and $R_2$ where the weighting is adaptively
chosen depending on the logits of $D_\psi(x)$ of the current discriminator at a data point $x$:
\begin{multline}
R_{\mathrm{Roth}}(\theta, \psi) = 
\E_{p_\theta(x)} \left[
(1 - \sigma(D_\psi(x)))^2 \|\nabla_x D_\psi(x) \|^2 
\right] \\
+ \E_{p_{\mathcal D}(x)}
\left[
\sigma(D_\psi(x)))^2\|\nabla_x D_\psi(x) \|^2
\right]
\end{multline}
Indeed, we can show that our convergence proof extends to this regularizer
(and a slightly more general class of regularizers):
\begin{remark}
When we replace the regularization terms $R_1$ and $R_2$ with
\begin{multline}
R_3(\theta, \psi) = 
\E_{p_\theta(x)} \left[
w_1(D_\psi(x))\|\nabla_x D_\psi(x) \|^2 
\right] \\
+ \E_{p_{\mathcal D}(x)}
\left[
w_2(D_\psi(x))\|\nabla_x D_\psi(x) \|^2
\right]
\end{multline}
so that $w_1(0) > 0$ and $w_2(0) > 0$,
the results of Theorem~\ref{thm:convergence} still hold.
\end{remark}
\begin{proof}
Again, we have to show that
$\mathcal M_G \times \mathcal M_D$ still consists only of equilibrium points
and that Lemma~\ref{lemma:general-MDD-regular} still holds in this setting.

However, by using $\nabla_x D_\psi(x) = 0$ for 
$x \in \supp p_{\mathcal D}$ and $\psi \in \mathcal M_D$, 
it is easy to see
that $\nabla_\psi R_3(\theta, \psi) = 0$ for all
$(\theta, \psi) \in \mathcal M_G \times \mathcal M_D$, which implies that
$\mathcal M_G \times \mathcal M_D$ still consists only of equilibrium points.

To see why Lemma~\ref{lemma:general-MDD-regular} still holds in this setting,
note that (after a little bit of algebra) we still have
$\nabla_{\theta,\psi} R_3(\theta^*, \psi^*) = 0$ and
\begin{equation}
\nabla^2_\psi R_3(\theta^*, \psi^*)
= \frac{1}{\gamma}(w_1(0) + w_2(0)) L_{DD}.
\end{equation}
The proof of Lemma~\ref{lemma:general-MDD-regular} therefore still
applies in this setting.
\end{proof}

\section{Stable equilibria for unregularized GAN training}\label{sec:stable-equilbria-unreg}
In Section~\ref{sec:main}, we have seen that unregularized GAN training
is not always locally convergent to the equilibrium point.
Moreover, in Section~\ref{sec:theory}, we have shown that zero-centered gradient penalties
make general GANs locally convergent under some suitable assumptions.

While our results demonstrate that we cannot expect  unregularized GAN training to lead to
local convergence for general GAN architectures, there can be situations where
unregularized GAN training has stable equilibria. 
Such equilibria usually require additional assumptions on the class of representable discriminators.

In this section, we identify two types of stable equilibria.
For the first class of stable equilibria, which we call \emph{energy solutions}, the
equilibrium discriminator forms an energy function for the true data distributions and might be a partial explanation
for the success of autoencoder-based discriminators \cite{DBLP:journals/corr/ZhaoML16,DBLP:journals/corr/BerthelotSM17}.
For the second class, which we call \emph{full-rank solutions},
the discriminator learns a representation of the data distribution 
with certain properties and
might be a partial explanation
for the success of batch-normalization for training GANs \cite{radford2015unsupervised}.

\subsection{Energy Solutions}\label{sec:energy-solutions}
For technical reasons, we assume that $\supp p_{\mathcal D}$ defines a $\mathcal C^1$-manifold in this section.

\emph{Energy solutions} are solutions where the discriminator forms a potential function for the true 
data distribution. Such solutions $(\theta^*, \psi^*)$ satisfy the following property:
\begin{assumption}{I$'$}\label{as:energy-realizable}
We have $p_{\theta^*} = p_{\mathcal D}$, $D_{\psi^*}(x) = 0$,  $\nabla_x D_{\psi^*}(x) = 0$ and
$v^\T\nabla^2_x D_{\psi^*}(x) v > 0$ for all $x \in \supp p_{\mathcal D}$ and $v$ not in the tangent space of $\supp p_{\mathcal D}$
at $x$.
\end{assumption}

We also need a  modified version of Assumption~\ref{as:regular} which ensures
certain regularity properties of the reparameterization manifolds $\mathcal M_G$ and 
$\mathcal M_D$ near the equilibrium $(\theta^*, \psi^*)$.
To formulate Assumption~\ref{as:energy-regular},
we need
\begin{equation}
\tilde g(\psi) :=
\nabla_\theta
\E_{p_\theta(x)} \left[ D_\psi(x) \right]
\bigl|_{\theta=\theta^*}.
\end{equation}
\begin{assumption}{III$'$}\label{as:energy-regular}
 There are $\epsilon$-balls $B_\epsilon(\theta^*)$ and $B_\epsilon(\psi^*)$
 around $\theta^*$ and $\psi^*$ 
 so that
 $\mathcal M_G \cap B_\epsilon(\theta^*)$
 and
 $\mathcal M_D \cap B_\epsilon(\psi^*)$
define $\mathcal C^1$- manifolds.
Moreover, the following holds:
\begin{compactenum}[(i)]
 \item if $v$ is not in the tangent space
 of $\mathcal M_D$ at $\psi^*$, then $\partial_v \tilde g(\psi^*) \neq 0$.
 \item if $w$ is not in the tangent space
 of $\mathcal M_G$ at $\theta^*$, then there is a latent code $z \in \mathbb R^k$ so that 
 $\nabla_\theta G_{\theta^*}(z)w$
 is not in the tangent space of  $\supp p_{\mathcal D}$ at $G_{\theta^*}(z) \in \supp p_{\mathcal D}$.
\end{compactenum}
\end{assumption}

The first part of Assumption~\ref{as:energy-regular} implies that the generator gradients 
become nonzero
whenever the discriminator moves away from an equilibrium discriminator. The second part of 
Assumption~\ref{as:energy-regular} means that every time the generator leaves the equilibrium, it pushes 
some data point aways from $\supp p_{\mathcal D}$, i.e. the generator is not simply
redistributing mass on $\supp p_{\mathcal D}$.

In Theorem~\ref{thm:energy-convergence} we show that energy solutions lead to local convergence
of the unregularized GAN training dynamics.
For the proof, we first need a generalization of
Lemma~\ref{lemma:general-Jacobian}:
\begin{lemma}\label{lemma:general-Jacobian-generalization}
Assume that $(\theta^*, \psi^*)$ satisfies Assumption~\ref{as:energy-realizable}.
The Jacobian of the gradient vector field $ v(\theta, \psi)$
at $(\theta^*, \psi^*)$ is then given by
\begin{equation}
v^\prime(\theta^*, \psi^*)
= \begin{pmatrix}
K_{GG} & -K_{DG}^\T \\
K_{DG} & K_{DD}
\end{pmatrix}.
\end{equation}
The terms  $K_{DD}$ and $K_{DG}$ are given by
\begin{align}
K_{GG}
& 
= 
-\begin{multlined}[t][3cm]
f^\prime(0) \E_{p(z)} \bigl[
\left[\nabla_\theta G_{\theta^*}(z)\right]^\T
\\
\nabla^2_x D_{\psi^*}(G_{\theta^*}(z))
\nabla_\theta G_{\theta^*}(z)
\bigr]
\end{multlined}
\\
K_{DD}
& = 2 f^{\prime\prime}(0) \E_{p_{\mathcal D}(x)} \left[
    \nabla_\psi D_{\psi^*}(x) \nabla_\psi D_{\psi^*}(x)^\T
    \right]
\\
K_{DG}
& = f^\prime(0) \left[
    \nabla_{\theta} \E_{p_\theta(x)}\left[
    \nabla_\psi D_{\psi^*}(x)
    \right]\mid_{\theta=\theta^*}
    \right]^\T
\end{align}
\end{lemma}
\begin{proof}
Almost all parts of the proof of Lemma~\ref{lemma:general-Jacobian} are still valid.
The only thing that remains to show is that $\nabla^2_\theta L(\theta^*, \psi^*) = -K_{GG}$.
To see this, just take the derivative of \eqref{eq:general-gradients-v-theta} with respect to $\theta$
and use the fact that $\nabla_x D_\psi(x) = 0$ for $x \in \supp p_{\mathcal D}$.
\end{proof}

We are now ready to formulate our convergence result for energy solutions:
\begin{thm}\label{thm:energy-convergence}
Assume Assumption~\ref{as:energy-realizable}, \ref{as:f} and \ref{as:energy-regular} hold
for $(\theta^*, \psi^*)$. Moreover, assume that $f^\prime(0) > 0$.
For small enough learning rates, simultaneous and
alternating gradient descent for the (unregularized) gradient vector field $v$ 
are both convergent to $\mathcal M_G \times \mathcal M_D$
in a neighborhood of $(\theta^*, \psi^*)$.
Moreover, the rate of convergence is at least linear.
\end{thm}
\begin{proof}[Proof (Sketch)]
The proof is similar to the proof of Theorem~\ref{thm:convergence}.

First, note that $\mathcal M_G \times \mathcal M_D$ still only consists of equilibrium points.
Next, we introduce local coordinates and show that for $v$ not in the tangent space of
$\mathcal M_G$ at $\theta^*$, we have $v^\T K_{GG} v < 0$. This can be shown using
Lemma~\ref{lemma:general-Jacobian-generalization},
Assumption~\ref{as:energy-realizable} and the second part of Assumption~\ref{as:energy-regular}.

Moreover, we need to show that
for $w$ not in the tangent space of $\mathcal M_D$ at $\psi^*$, we have
$K_{DG}^\T w \neq 0$.
This can be shown by applying the first part of Assumption~\ref{as:energy-regular}.

The rest of the proof is the same as the proof of Theorem~\ref{thm:convergence},
except that we have to apply Theorem~\ref{thm:eigenvalue-bounds-generalized} instead of 
Theorem~\ref{thm:eigenvalue-bounds}.
\end{proof}

Note that energy solutions are only possible, if the discriminator is able to 
satisfy Assumption~\ref{as:energy-realizable}.
This is not the case for the Dirac-GAN from Section~\ref{sec:main}.
However, if we use a quadratic discriminator instead, there are also energy solutions
to the unregularized GAN training dynamics for the Dirac-GAN.
To see this, we can parameterize $D_\psi(x)$ as 
\begin{equation}\label{eq:dirac-quadratic}
 D_\psi(x) := \psi_1 x^2 + \psi_2 x.
\end{equation}
It is easy to check that the Dirac-GAN with a discriminator as in \eqref{eq:dirac-quadratic} 
indeed has energy solutions:
every $(\theta, \psi)$ with $\theta=0$ and $\psi_2=0$ defines an equilibrium point of the 
Dirac-GAN and the GAN-training dynamics are locally convergent near this point if $\psi_1 > 0$.
Note however, that even though all equilbria with $\psi_1 > 0$ are points of attraction for
the \emph{continuous} GAN training dynamics, they may not be attractors for the \emph{discretized system} when
$\psi_1$ is large and the learning rate $h$ is fixed.
In general, the conditioning of energy solutions depends on the condition numbers of the Hessians
$\nabla_x^2 D_{\psi^*} (x)$ at all $x \in \supp p_{\mathcal D}$. 
Indeed, the presence of ill-conditioned energy solutions might be one possible
explanation why WGAN-GP often works well in practice although it is 
not even locally convergent for the Dirac-GAN.

\subsection{Full-Rank Solutions}
In practice, $D_\psi(x)$ is usually implemented by a deep neural network.
Such discriminators can be described by functions of the form
\begin{equation}
D_\psi(x) = \psi_1^\T \eta_{\psi_2} (x)
\end{equation}
with a vector-valued $\mathcal C^1$-functions $\eta_{\psi_2}$ and $\psi = (\psi_1, \psi_2)$. 
$\eta_{\psi_2}$ can be regarded as a feature-representation of the data point $x$.

We now state several assumptions that lead to local convergence
in this situation.

The first assumption can be seen as a variant of Assumption~\ref{as:realizable}
adapted to this specific situation:
\begin{assumption}{I$''$}\label{as:cdr-realizable}
We have $p_{\theta^*} = p_{\mathcal D}$ and $\psi_1^* = 0$.
\end{assumption}

We again consider  \emph{reparameterization manifolds}, which we define as follows in this section:
\begin{equation}
 \mathcal M_G := \{ \theta \mid p_\theta = p_{\mathcal D} \} 
 \quad
 \mathcal M'_D := \{ \psi \mid \psi_1 = 0 \}.
\end{equation}

Moreover, let
\begin{equation}
 g(\theta) =  \E_{p_\theta(x)}
 \left[ \eta_{\psi^*_2}(x) \right].
\end{equation}

Assumption~\ref{as:regular} now becomes:
\begin{assumption}{III$''$}\label{as:cdr-regular}
 There is an $\epsilon$-ball $B_\epsilon(\theta^*)$ around $\theta^*$
so that $\mathcal M_G$
defines a $\mathcal C^1$- manifold\footnote{%
Note that $\mathcal M'_D$ is a $\mathcal C^1$-manifold by definition
in this setup.}.
Moreover, the following holds:
\begin{compactenum}[(i)]
 \item
 The matrix
 $\E_{p_{\mathcal D}(x)}\left[ \eta_{\psi^*_2}(x)  \eta_{\psi^*_2}(x)^\T\right]$
 has full rank.
 \item if $w$ is not in the tangent space
 of $\mathcal M_G$ at $\theta^*$, then
 $\partial_w g(\theta^*) \neq 0$.
\end{compactenum}
\end{assumption}
We call a function $\eta_{\psi_2^*}$ that satisfies the first part of Assumption~\ref{as:cdr-regular}
a \emph{full-rank representation} of $p_{\mathcal D}$.
Moreover, if $\eta_{\psi^*_2}$ satisfies the second part of Assumption~\ref{as:cdr-regular},
we call $\eta_{\psi^*_2}$  a \emph{complete representation}, because
the second part of Assumption~\ref{as:cdr-regular} implies
that every deviation from the Nash-equilibrium $p_{\theta^*} = p_{\mathcal D}$ is detectable 
using $\eta_{\psi^*_2}$.

In practice, complete full-rank representations might only exist if the class of discriminators is very powerful or the class
of generators is limited.
Especially the second part of Assumption~\ref{as:cdr-regular} might be hard to satisfy in practice.
Moreover, finding such representations might be much harder than finding equilibria
for the regularized GAN-training dynamics from Section~\ref{sec:theory}.

Nonetheless, we have the following convergence result for GANs that allow for complete full-rank
representations:
\begin{thm}\label{thm:cdr-convergence}
Assume Assumption~\ref{as:energy-realizable}, Assumption~\ref{as:f} and \ref{as:energy-regular} hold
for $(\theta^*, \psi^*)$.
For small enough learning rates, simultaneous and
alternating gradient descent for the (unregularized) gradient vector field $v$ 
are both convergent to $\mathcal M_G \times \mathcal M'_D$
in a neighborhood of $(\theta^*, \psi^*)$.
Moreover, the rate of convergence is at least linear.
\end{thm}
\begin{proof}[Proof (Sketch)]
The proof is again similar to the proof of Theorem~\ref{thm:convergence}.
We again introduce local coordinates and show that for $w$ not in the tangent space of
$\mathcal M'_D$ at $\psi^*$, we have $w^\T K_{DD} w < 0$. To see this, note that
$w$ must have a nonzero $\psi_1$ component if it is not in the tangent space of
$\mathcal M'_D$ at $\psi^*$. However, using \eqref{eq:general-Jacobian-KDD}, we see that the submatrix of $K_{DD}$ corresponding to the
$\psi_1$ coordinates is given by
\begin{equation}
 \tilde K_{DD}
 = 2 f^{\prime\prime}(0) 
 \E_{p_{\mathcal D}(x)}\left[ \eta_{\psi^*_2}(x)  \eta_{\psi^*_2}(x)^\T\right].
\end{equation}
This matrix is negative definite by Assumption~\ref{as:f}
and the first part of Assumption~\ref{as:cdr-regular}.

Moreover, by applying \eqref{eq:general-Jacobian-KDG},
we see that the component of
$K_{DG} w$, $w \in \mathbb R^n$, corresponding to the $\psi_1$ coordinates is given by
\begin{equation}
 \partial_w g(\theta^*)
 =
f^\prime(0)
\nabla_\theta \E_{p_\theta(x)}
\left[ \eta_{\psi^*_2}(x) \right] 
\bigl|_{\theta = \theta^*} w.
\end{equation}
Using the second part of Assumption~\ref{as:cdr-regular}, we therefore see that
for $w$ not in the tangent space of $\mathcal M_G$ at $\theta^*$, we have
$K_{DG} w \neq 0$.

The rest of the proof is the same as the proof of Theorem~\ref{thm:convergence}.
\end{proof}

For the Dirac-GAN from Section~\ref{sec:main}, we can obtain a complete full-rank representation, when we parameterize
the discriminator $D_\psi$ as $D_\psi(x) = \psi \exp(x)$, i.e. if we set
$\psi_1 := \psi$ and $\eta_{\psi_2}(x) := \exp(x)$. 
It is easy to check that $\eta_{\psi_2}$ indeed defines a complete full-rank representation and that
the Dirac-GAN is locally convergent to $(\theta^*, \psi^*) = (0, 0)$ for this parameterization of $D_\psi(x)$.

\FloatBarrier
\section{Experiments}
In this section, we describe additional experiments and give more details on our experimental setup.
If not noted otherwise, we always use the nonsaturating GAN-objective introduced by \citet{DBLP:conf/nips/GoodfellowPMXWOCB14}
for training the generator. For WGAN-GP we use the generator and discriminator objectives introduced by \citet{DBLP:conf/nips/GulrajaniAADC17}.\footnote{%
The code to reproduce the experiments presented in this section can be found
under
\url{https://github.com/LMescheder/GAN_stability}.
}

\begin{figure*}
\centering
\begin{subfigure}[t]{0.2\linewidth}
\includegraphics[width=\linewidth]{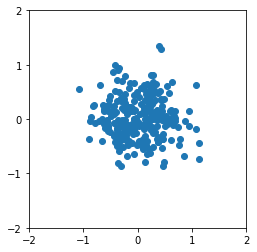}
\caption{2D-Gaussian}
\end{subfigure}
\begin{subfigure}[t]{0.2\linewidth}
\includegraphics[width=\linewidth]{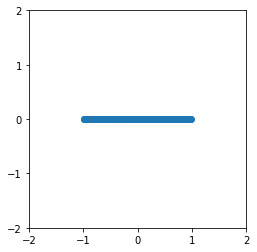}
\caption{Line}
\end{subfigure}
\begin{subfigure}[t]{0.2\linewidth}
\includegraphics[width=\linewidth]{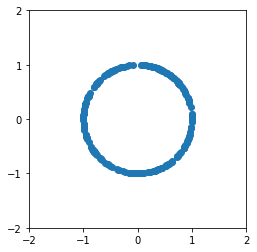}
\caption{Circle}
\end{subfigure}
\begin{subfigure}[t]{0.2\linewidth}
\includegraphics[width=\linewidth]{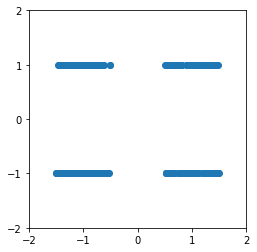}
\caption{Four lines}
\end{subfigure}
\caption{The four 2D-data distributions on which we test the different algorithms.}
\label{fig:2d-data}
\end{figure*}

\paragraph{2D-Problems} 
For the 2D-problems, we run unregularized GAN training,
$R_1$-regularized and $R_2$-regularized GAN training 
as well WGAN-GP with $1$ and $5$ discriminator update per generator update.
We run each method on $4$ different 2D-examples
for $6$ different GAN architectures.
The $4$ data-distributions are visualized in Figure~\ref{fig:2d-data}.
All $6$ GAN architectures consist of $4$-layer fully connected neural networks for both the
generator and discriminator, where we select the number of hidden units from $\{8, 16, 32\}$ and 
use select either leaky RELUs (i.e. $\varphi(t) = \max(t, 0.2 t)$) or Tanh-activation functions.

For each method, we try both Stochastic Gradient Descent (SGD) and RMS-Prop
with $4$ different learning rates: 
for SGD, we select the learning rate from $\{5 \cdot 10^{-3}, 10^{-2}, 2 \cdot 10^{-2}, 5 \cdot 10^{-2}\}$.
For RMSProp, we select it from $\{5 \cdot 10^{-5}, 10^{-4}, 2 \cdot 10^{-4}, 5 \cdot 10^{-4}\}$.
For the $R_1$-, $R_2$- and WGAN-GP-regularizers we try the regularization parameters
$\gamma = 1$, $\gamma=3$ and $\gamma = 10$.
For each method and architecture,
we pick the hyperparameter setting which achieves
the lowest Wasserstein-1-distance to the true data distribution.
We train all methods for 50k iterations and we 
report the Wasserstein-1-distance averaged over the last 10k iterations.
We estimate the Wasserstein-1-distance using the
Python Optimal Transport package\footnote{\url{http://pot.readthedocs.io}}
by drawing $2048$ samples from both the generator distribution
and the true data distribution.

\begin{figure*}
\centering
\begin{subfigure}[t]{0.18\linewidth}
\includegraphics[width=\linewidth]{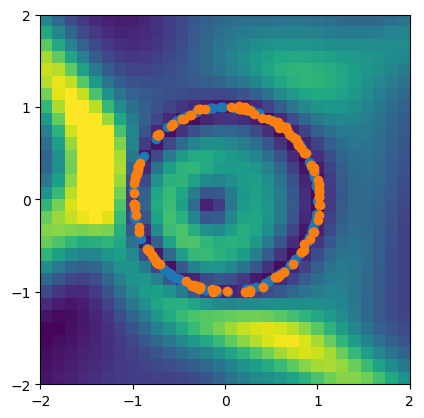}
\caption{unregularized}
\end{subfigure}
\begin{subfigure}[t]{0.18\linewidth}
\includegraphics[width=\linewidth]{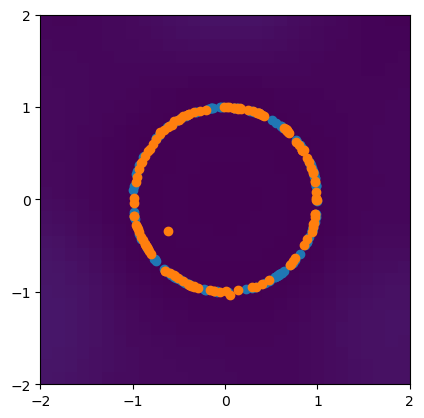}
\caption{$R_1$}
\end{subfigure}
\begin{subfigure}[t]{0.18\linewidth}
\includegraphics[width=\linewidth]{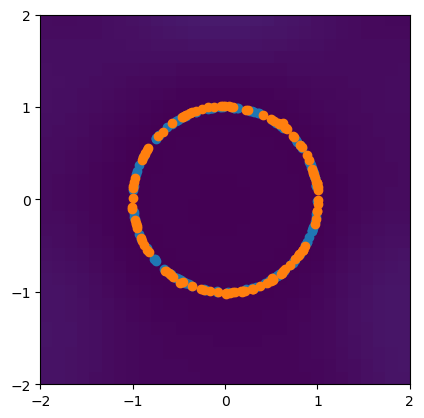}
\caption{$R_2$}
\end{subfigure}
\begin{subfigure}[t]{0.18\linewidth}
\includegraphics[width=\linewidth]{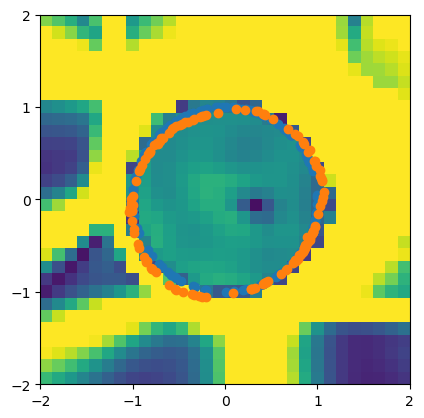}
\caption{WGAN-GP-1}
\end{subfigure}
\begin{subfigure}[t]{0.18\linewidth}
\includegraphics[width=\linewidth]{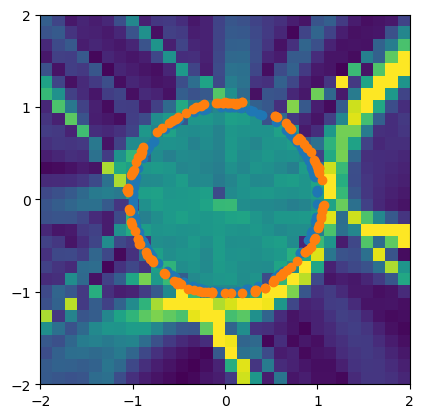}
\caption{WGAN-GP-5}
\end{subfigure}
\caption{Best solutions found by the different algorithms for learning a circle.
The blue points are samples from the true data distribution, the orange points are samples from 
the generator distribution. The colored areas visualize the gradient magnitude of the equilibrium discriminator.
We find that while the $R_1$- and $R_2$-regularizers converge to equilibrium discriminators that are $0$ in a neighborhood
of the true data distribution, unregularized training and WGAN-GP converge to energy solutions (Section~\ref{sec:energy-solutions}).}
\label{fig:2d-solutions}
\end{figure*}

The best solution found by each method for the ``Circle''-distribution 
is shown in Figure~\ref{fig:2d-solutions}.
We see that the $R_1$- and $R_2$-regularizers converge to solutions for which the discriminator is $0$ in a
neighborhood of the true data distribution.
On the other hand,
unregularized training and WGAN-GP converge to \emph{energy solutions} where the discriminator
forms a potential function for the true data distribution. Please see Section~\ref{sec:energy-solutions} for details.

\paragraph{CIFAR-10}
To test our theory on real-world tasks, we train a 
DC-GAN architecture \cite{radford2015unsupervised}
with $3$ convolutional layers and no
batch-normalization on the CIFAR-10 dataset \cite{krizhevsky2009learning}. 
We apply different regularization strategies to stabilize the training.
To compare the different regularization strategies, we measure the 
inception score \cite{DBLP:conf/nips/SalimansGZCRCC16} over Wall-clock-time.
We implemented the network in the Tensorflow framework \cite{DBLP:conf/osdi/AbadiBCCDDDGIIK16}.
For all regularization techniques, we use the RMSProp optimizer \cite{tieleman2012lecture}
with $\alpha = 0.9$ and a learning rate of $10^{-4}$.

For the $R_1$ and $R_2$ regularizers from Section~\ref{sec:method} we use a regularization parameter
of $\gamma = 10$. For the WGAN-GP regularizer we also use a regularization parameter of $\gamma =10$
as suggested by \citet{DBLP:conf/nips/GulrajaniAADC17}. We train all methods using $1$ discriminator update
per generator update except for WGAN-GP, for which we try both $1$ and $5$ discriminator updates

\begin{figure}[t]
\centering
\includegraphics[width=1.\linewidth]{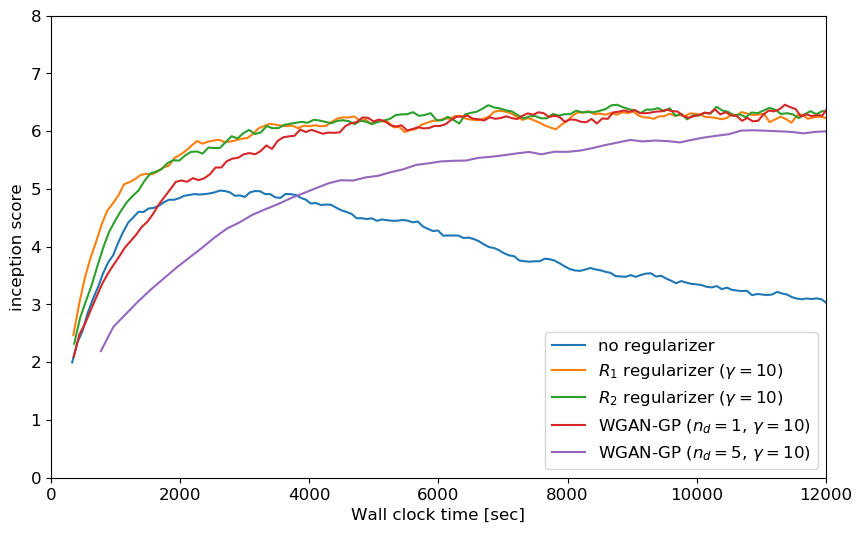}
\caption{
Inception score 
over time for various regularization strategies when training on CIFAR-10.
While the inception score can be problematic for evaluating
probabilistic models \cite{barratt2018note}, 
it still gives a rough idea about the
convergence and stability properties of different training methods.
}
\label{fig:cifar-10}
\end{figure}

The inception score over time for the different
regularization strategies is shown in Figure~\ref{fig:cifar-10}.
As predicted by our theory, we see that the $R_1$- and $R_2$-regularizers
from Section~\ref{sec:method} lead to stable training whereas unregularized
GAN training is not stable.
We also see that WGAN-GP with $1$ or $5$ discriminator updates per generator update
lead to similar final inception scores on this architecture.
The good behavior of WGAN-GP is surprising considering the fact that it does not
even converge locally for the Dirac-GAN.
One possible explanation is that WGAN-GP oscillates in narrow circles around the equilibrium 
which might be enough to produce images of sufficiently high quality.
Another possible explanation is that WGAN-GP converges to an energy or full-rank solution
(Section~\ref{sec:stable-equilbria-unreg}) for this example.

\paragraph{Imagenet}
In this experiment, we use the $R_1$-regularizer to learn a generative model of all $1000$ Imagenet \cite{ILSVRC15} classes at resolution $128 \times 128$ in a single GAN.
Because of the high variability of this dataset, this is known to be a challenging task and only few prior works have managed to obtain recognizable samples for this dataset.
Prior works that report results for this dataset either show a high amount of mode collapse
\cite{DBLP:conf/nips/SalimansGZCRCC16, odena2016conditional},
report results only at a lower resolution \cite{hjelm2017boundary}
or use advanced normalization layers to stabilize the training \cite{miyato2018spectral}.

For the Imagenet experiment, we use ResNet-architectures\footnote{%
We used more complicated architectures for the generator and discriminator in 
an earlier version of this manuscript. 
The simplified architectures presented in this version are more efficient
and perform slightly better.}
for the generator and discriminator, both
having $26$ layers in total.
Both the generator and discriminator are conditioned on the labels of the input data.
The architectures for the generator and discriminator are
shown in Table~\ref{tab:architecture-imagenet}.
We use preactivation ResNet-blocks and Leaky RELU-nonlinearities
everywhere. We also multiply the output of the ResNet blocks with $0.1$. 
For the generator, we sample a latent variable $z$ from a $256$-dimensional unit Gaussian distribution and concatenate it with a $256$ dimensional embedding of the labels,
which we normalize to the unit sphere.
The resulting $512$-dimensional vector is then fed into the first fully connected layer
of the generator.
The discriminator takes as input an image and outputs a $1000$ dimensional vector.
Depending on the label of the input, we select the corresponding index in this vector
and use it as the logits for the GAN-objective.

For training, we use the RMSProp optimizer
with $\alpha = 0.99$, $\epsilon=10^{-8}$ and an initial learning rate of $10^{-4}$.
We use a batch size of $128$ and
we train the networks on $4$ GeForce GTX 1080 Ti GPUs for $500.000$ iterations.
Similarly to prior work \cite{karras2017progressive,yazici2018unusual,gidel2018variational},
we use an exponential moving average\footnote{%
In an earlier version of this manuscript, we instead annealed the learning rate
which has a similar effect. However, taking a moving average introduces fewer 
hyperparameters and led to higher inception scores in our experiments.}
with decay $0.999$ over the weights to produce the final model.

We find that while training this GAN without any regularization quickly leads to mode collapse,
using the $R_1$-regularizers from Section~\ref{sec:method} leads to stable training.

\begin{figure*}[ht!]
\centering
\begin{subfigure}[t]{0.4\linewidth}
\centering
\includegraphics[width=\linewidth]{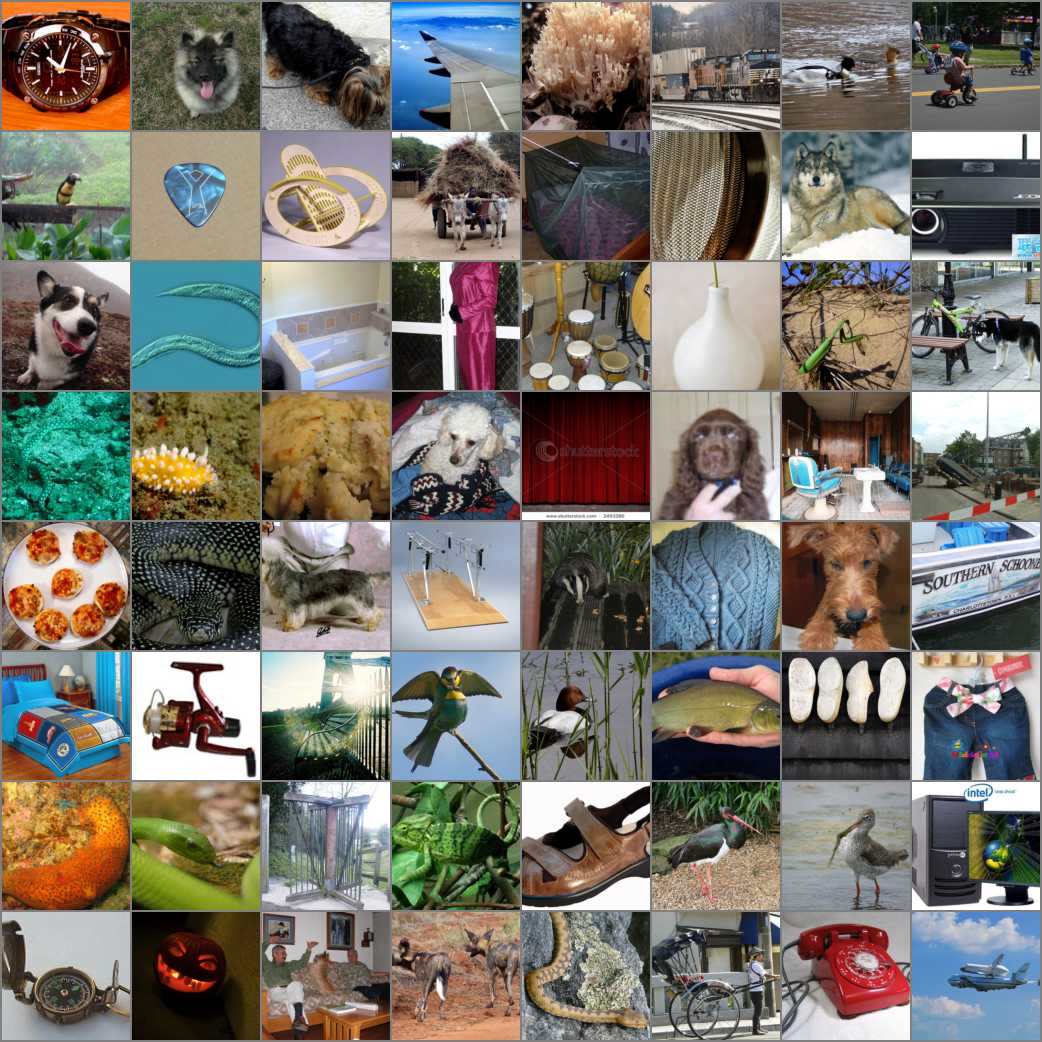}
 \caption{Training distribution}
\end{subfigure}
\quad
\begin{subfigure}[t]{0.4\linewidth}
\centering
\includegraphics[width=\linewidth]{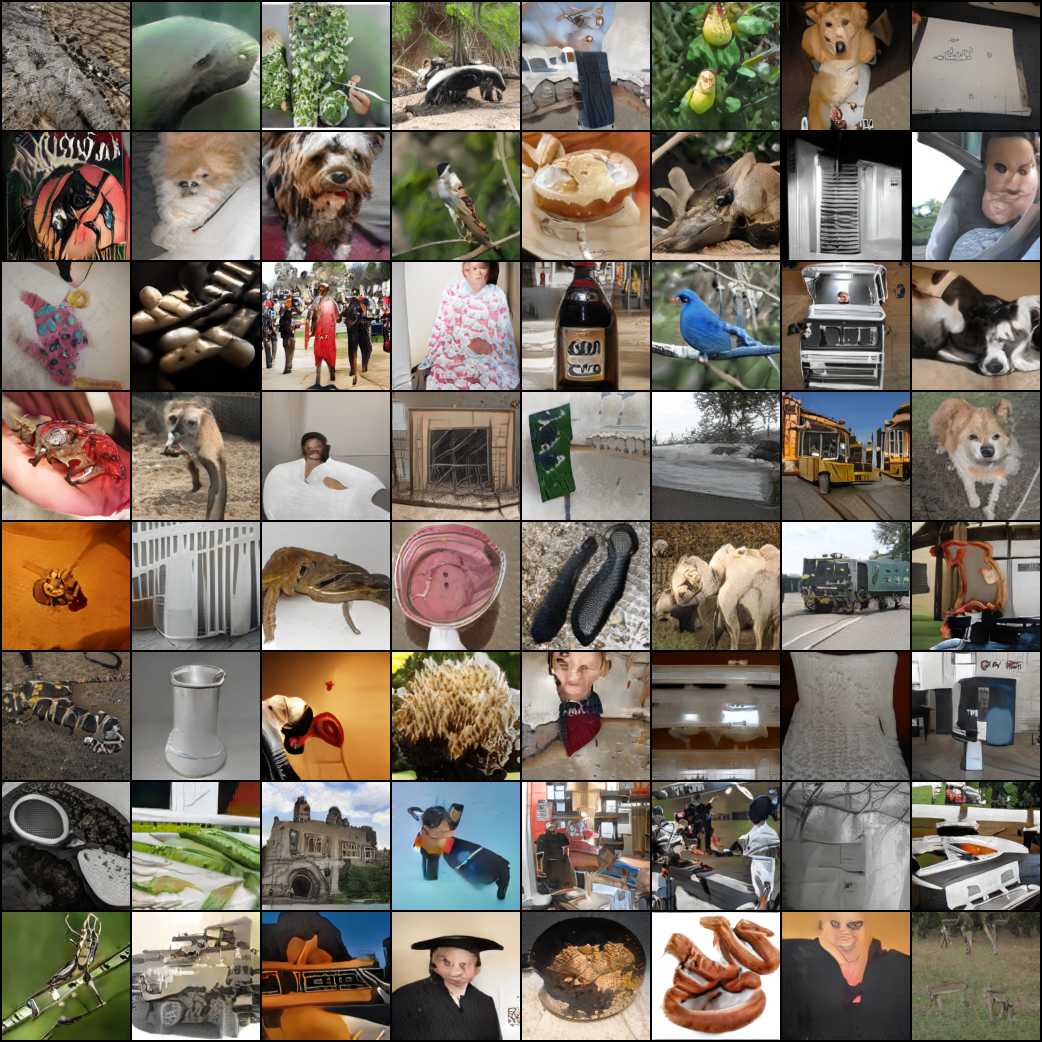}
 \caption{Random samples}
\end{subfigure}
\caption{Unconditional random samples for a GAN trained on the ILSVRC dataset \cite{ILSVRC15} with resolution $128 \times 128$.
The final inception score
is $30.2 \pm 0.5$.
}
\label{fig:result-imagenet}
\end{figure*}

\begin{figure*}[ht!]
\centering
\begin{subfigure}[t]{0.3\linewidth}
\includegraphics[width=\linewidth]{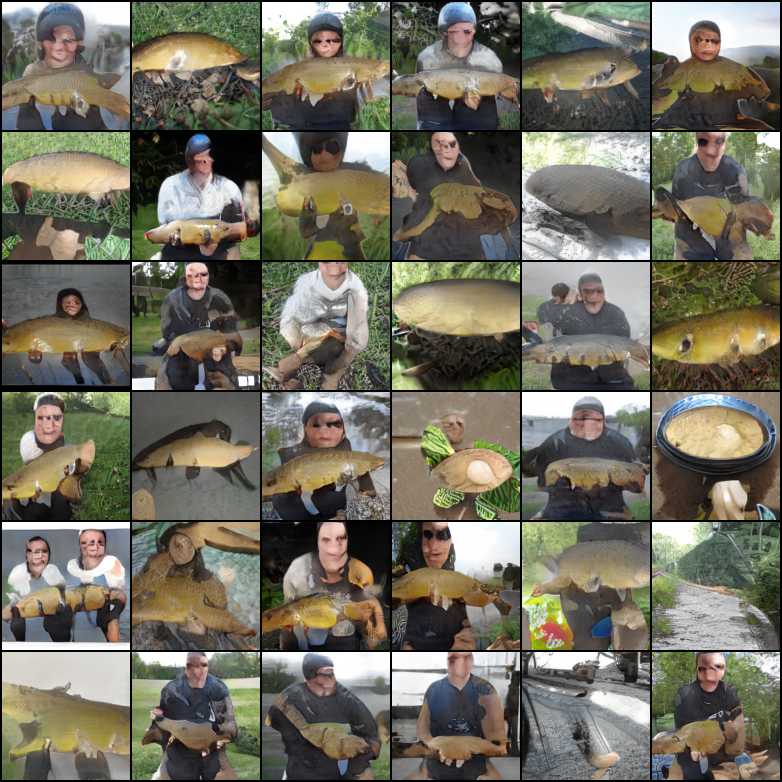}
 \caption{tench}
\end{subfigure}
\begin{subfigure}[t]{0.3\linewidth}
\includegraphics[width=\linewidth]{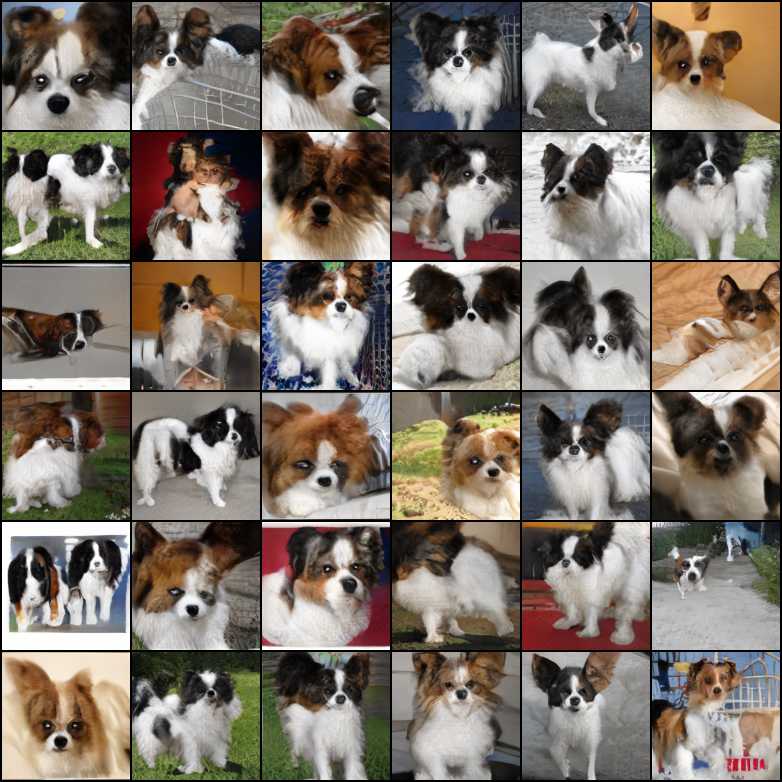}
 \caption{papillon}
\end{subfigure}
\begin{subfigure}[t]{0.3\linewidth}
\includegraphics[width=\linewidth]{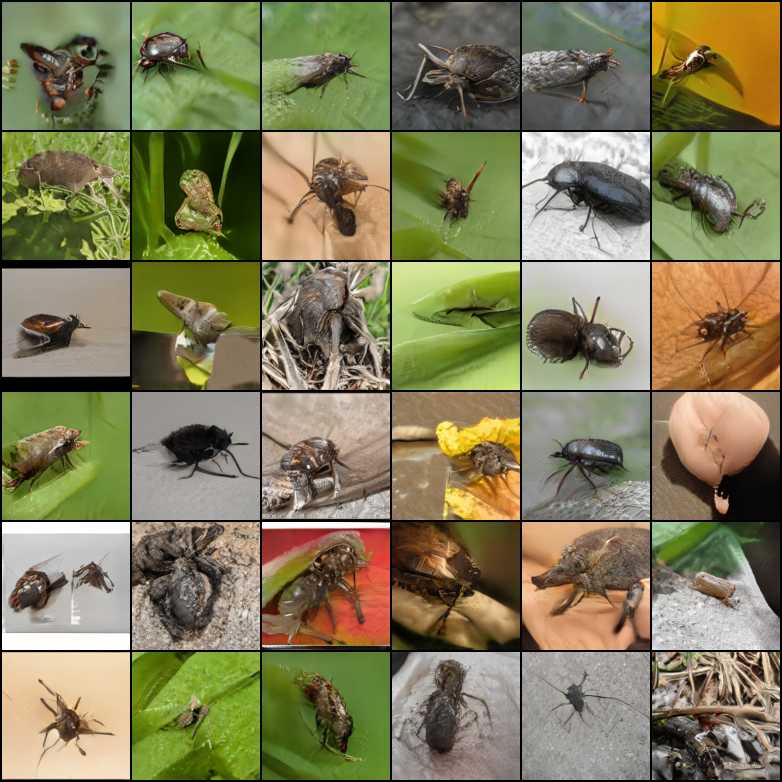}
\caption{weevil}
\end{subfigure}

\begin{subfigure}[t]{0.3\linewidth}
\includegraphics[width=\linewidth]{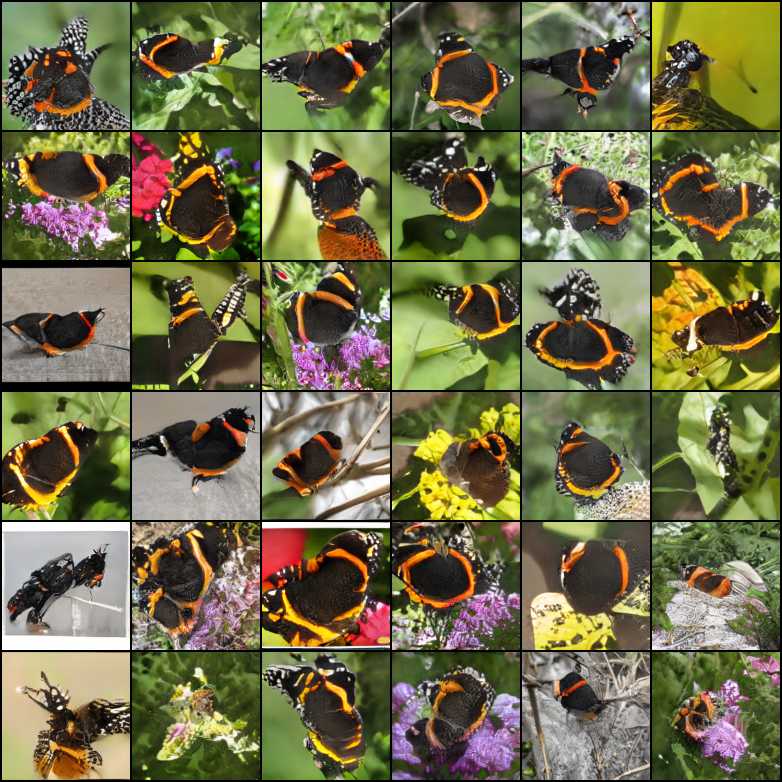}
 \caption{admiral}
\end{subfigure}
\begin{subfigure}[t]{0.3\linewidth}
\includegraphics[width=\linewidth]{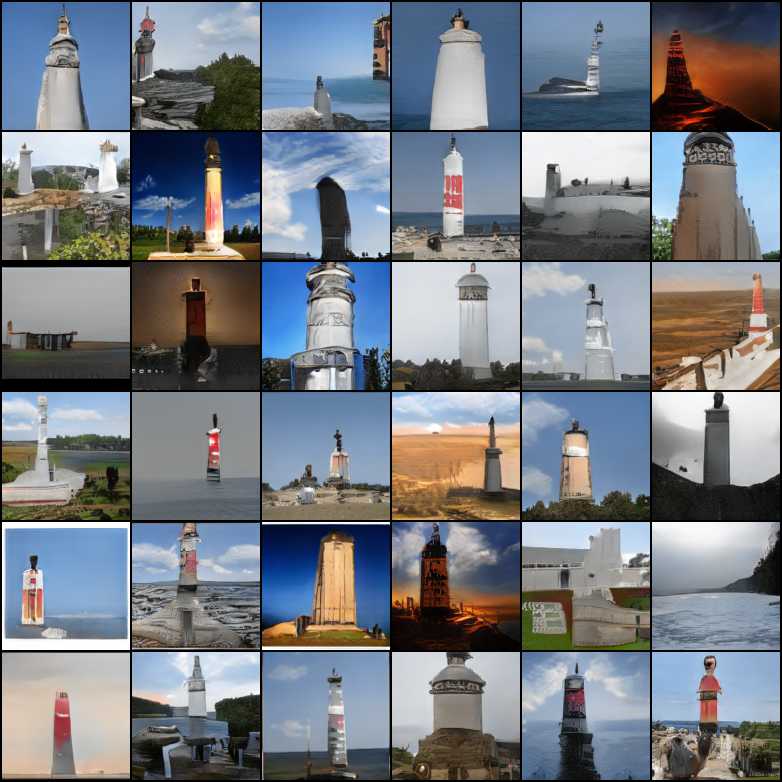}
 \caption{lighthouse}
\end{subfigure}
\begin{subfigure}[t]{0.3\linewidth}
\includegraphics[width=\linewidth]{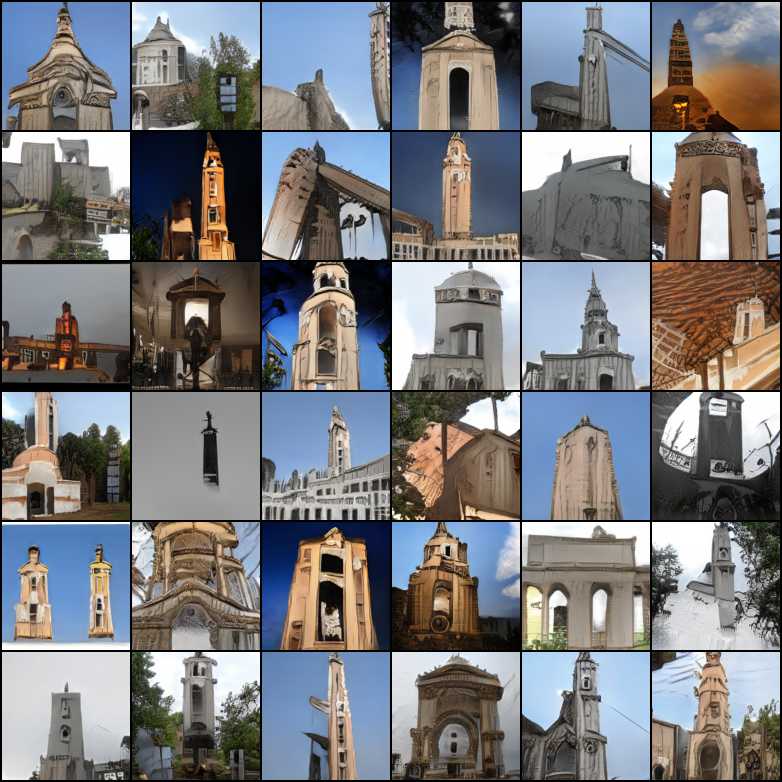}
 \caption{bell cote}
\end{subfigure}

\begin{subfigure}[t]{0.3\linewidth}
\includegraphics[width=\linewidth]{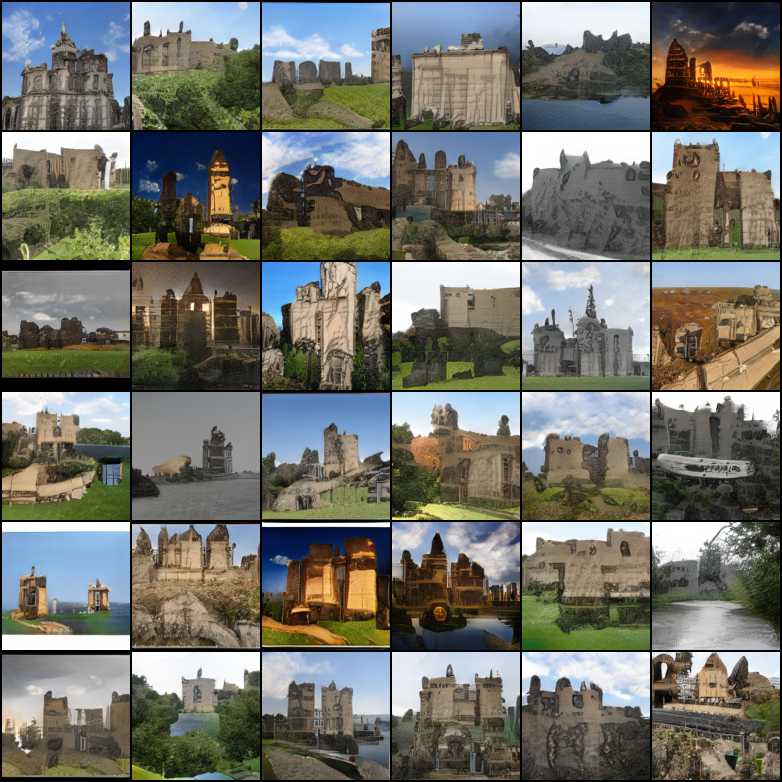}
 \caption{castle}
\end{subfigure}
\begin{subfigure}[t]{0.3\linewidth}
\includegraphics[width=\linewidth]{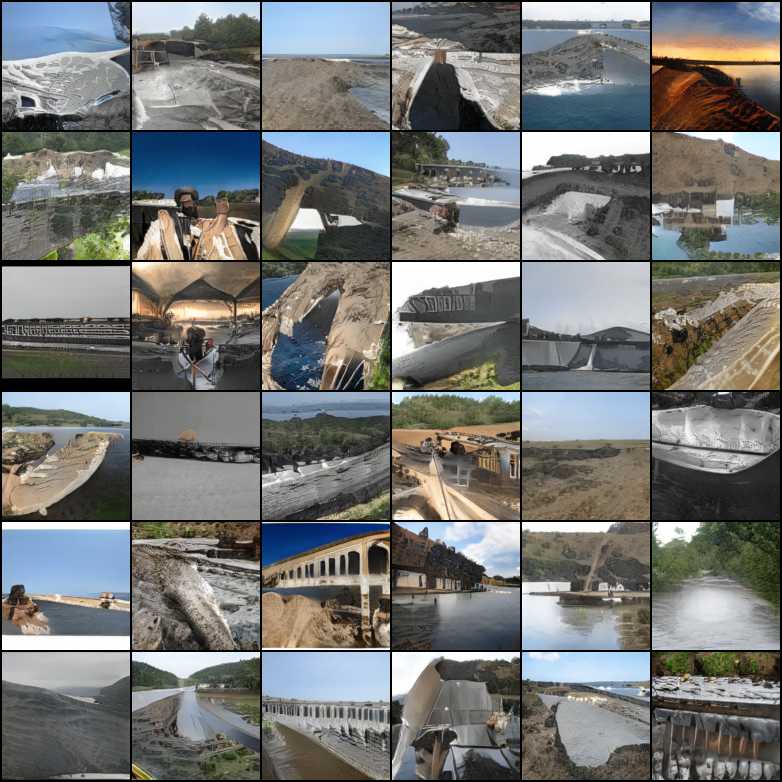}
 \caption{dam}
\end{subfigure}
\begin{subfigure}[t]{0.3\linewidth}
\includegraphics[width=\linewidth]{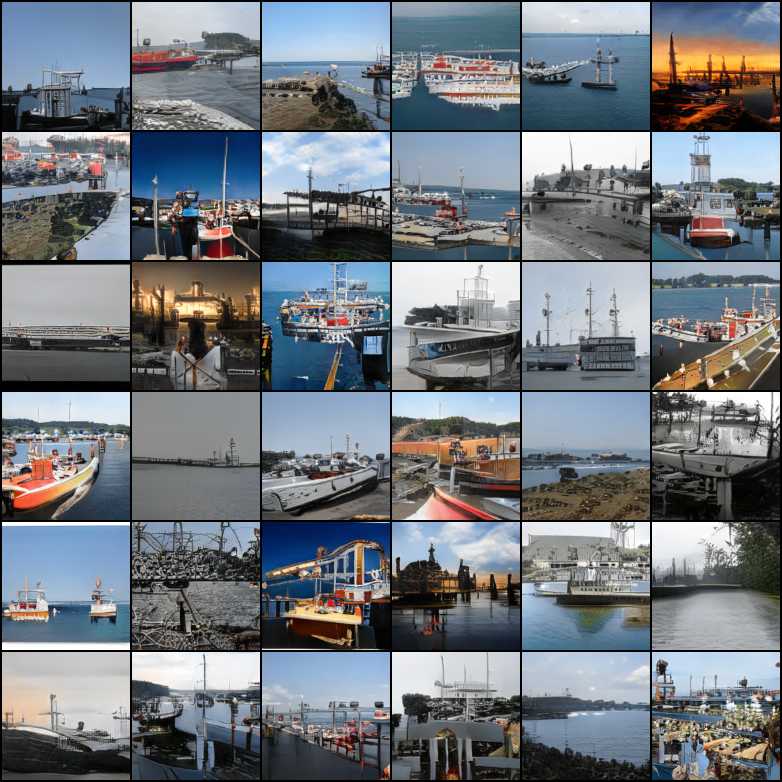}
 \caption{dock}
\end{subfigure}

\caption{Class conditional random samples for a GAN trained on the Imagenet dataset.}
\label{fig:result-imagenet-cond1}
\end{figure*}

\begin{figure*}[ht!]
\centering
\begin{subfigure}[t]{0.3\linewidth}
\includegraphics[width=\linewidth]{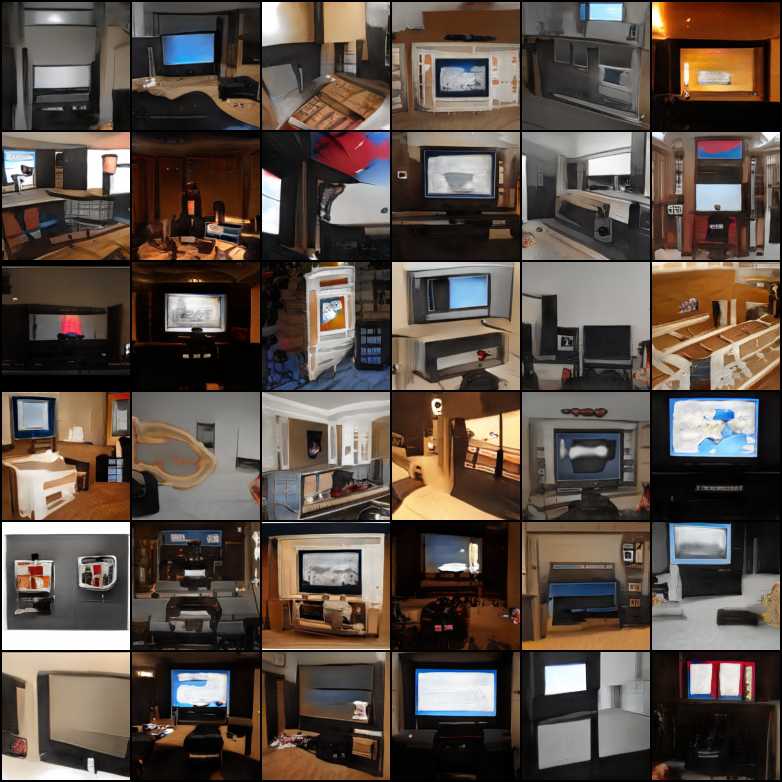}
 \caption{home theater}
\end{subfigure}
\begin{subfigure}[t]{0.3\linewidth}
\includegraphics[width=\linewidth]{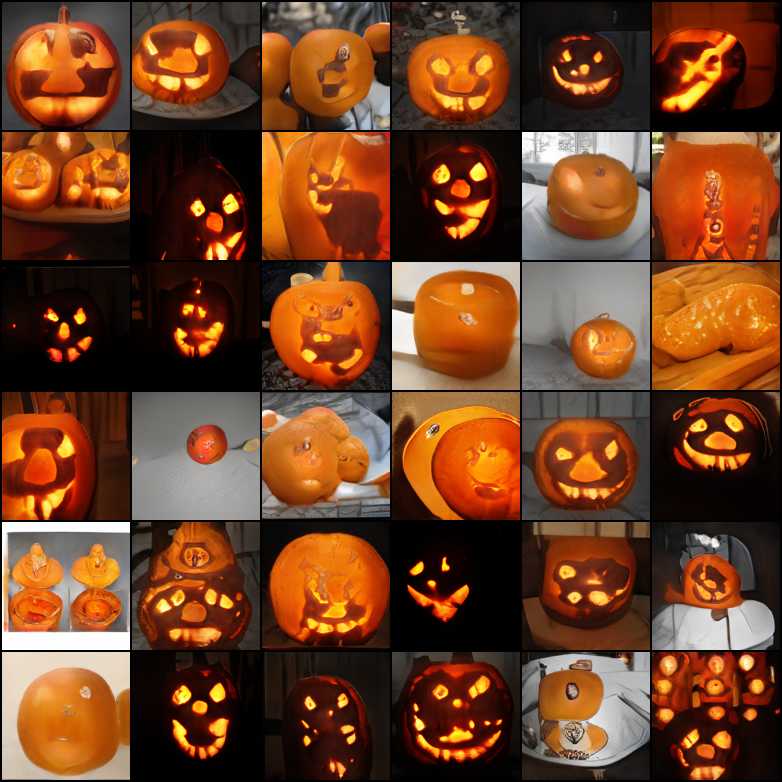}
 \caption{jack-o'-lantern}
\end{subfigure}
\begin{subfigure}[t]{0.3\linewidth}
\includegraphics[width=\linewidth]{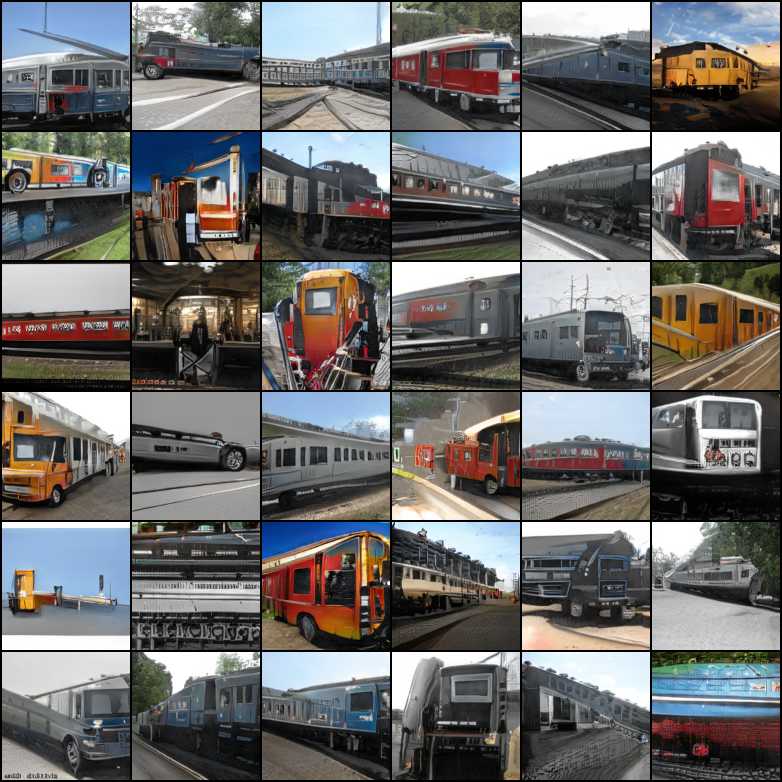}
\caption{passenger car}
\end{subfigure}

\begin{subfigure}[t]{0.3\linewidth}
\includegraphics[width=\linewidth]{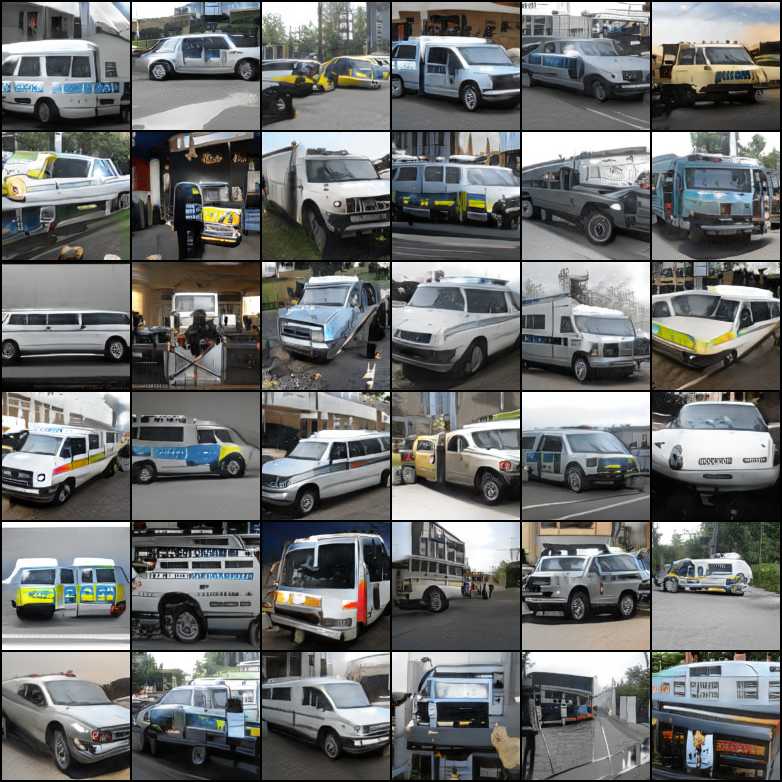}
 \caption{police van}
\end{subfigure}
\begin{subfigure}[t]{0.3\linewidth}
\includegraphics[width=\linewidth]{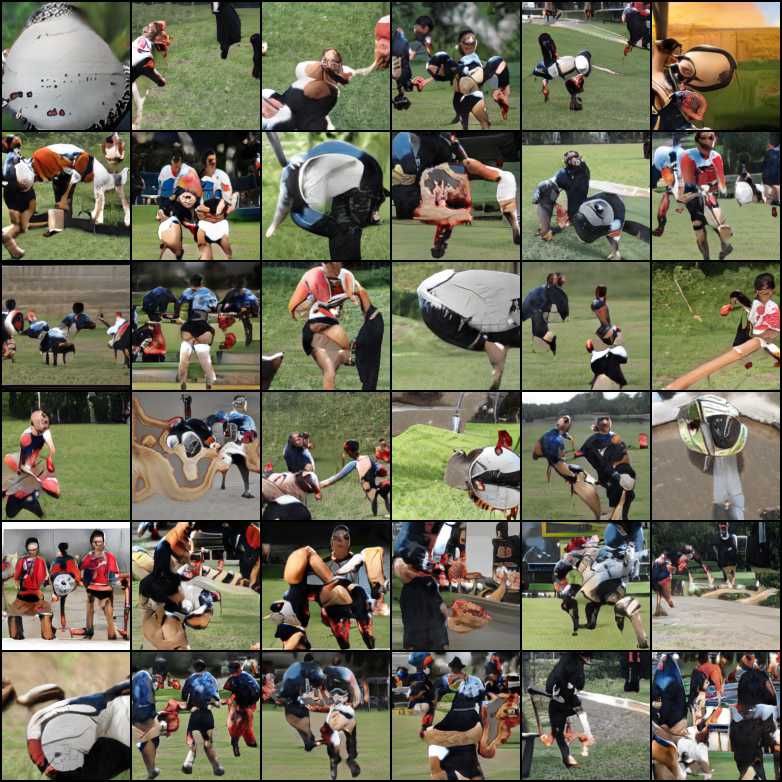}
 \caption{rugby ball}
\end{subfigure}
\begin{subfigure}[t]{0.3\linewidth}
\includegraphics[width=\linewidth]{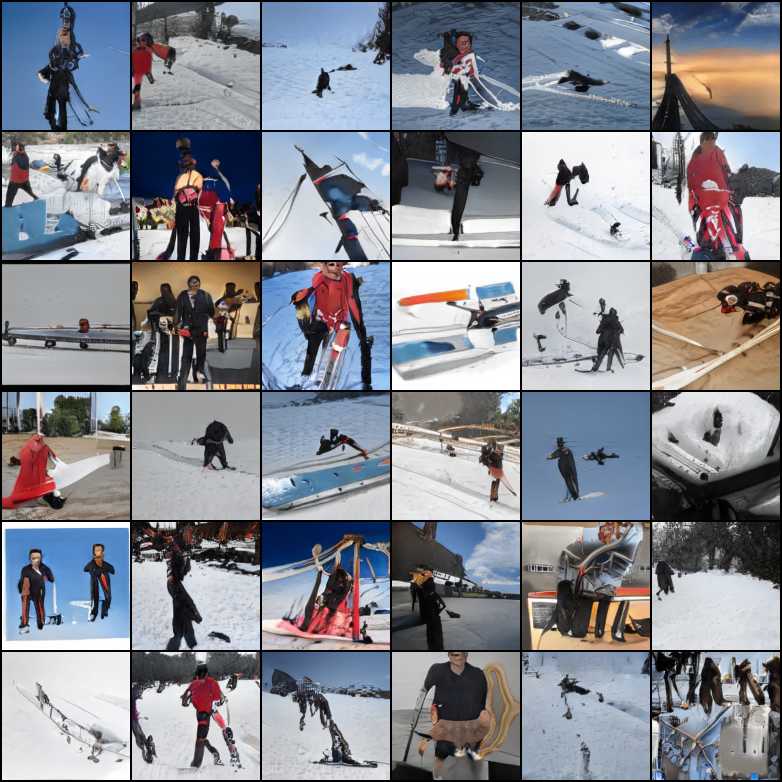}
 \caption{ski}
\end{subfigure}

\begin{subfigure}[t]{0.3\linewidth}
\includegraphics[width=\linewidth]{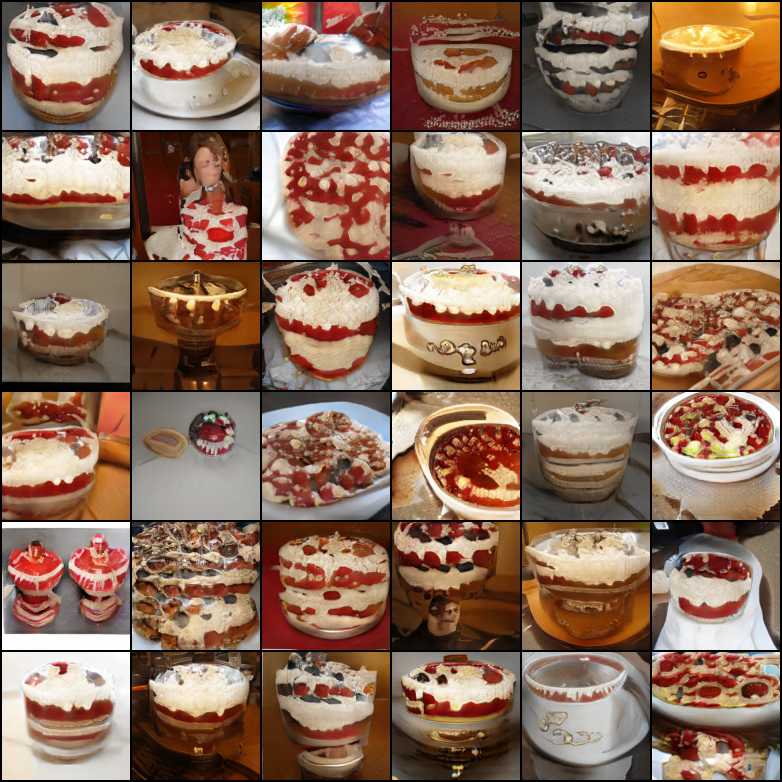}
 \caption{trifle}
\end{subfigure}
\begin{subfigure}[t]{0.3\linewidth}
\includegraphics[width=\linewidth]{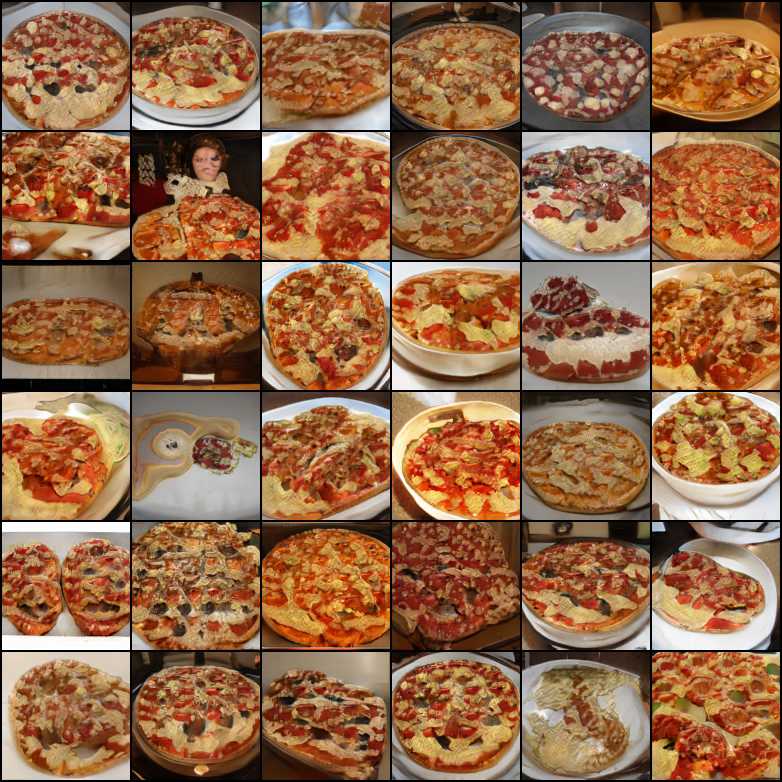}
 \caption{pizza}
\end{subfigure}
\begin{subfigure}[t]{0.3\linewidth}
\includegraphics[width=\linewidth]{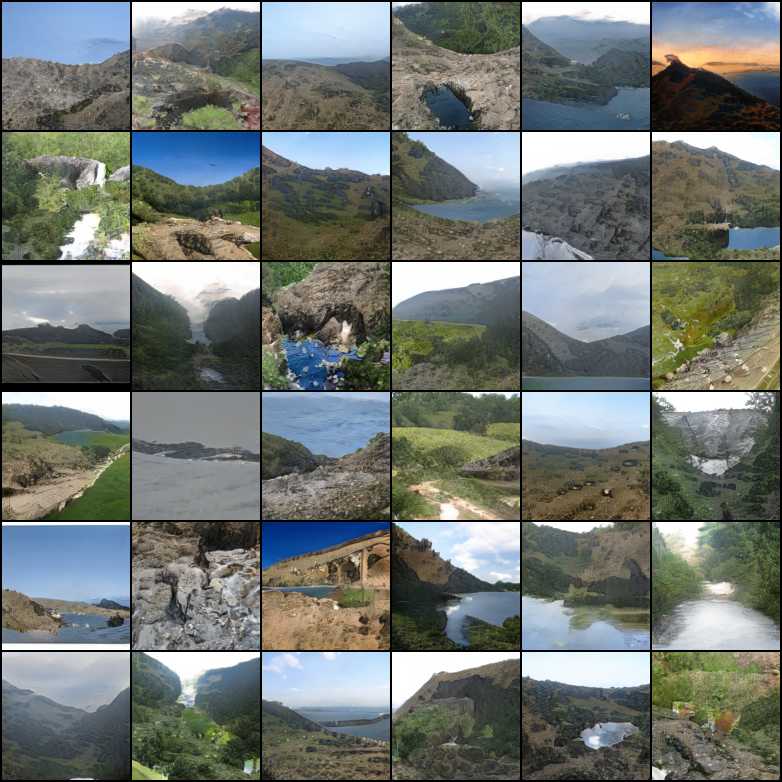}
 \caption{valley}
\end{subfigure}

\caption{Class conditional random samples for a GAN trained on the Imagenet dataset.}
\label{fig:result-imagenet-cond2}
\end{figure*}

Some random (unconditional) samples can be seen in Figure~\ref{fig:result-imagenet}.
Moreover, Figure~\ref{fig:result-imagenet-cond1} and Figure~\ref{fig:result-imagenet-cond2}
show conditional samples for some selected Imagenet classes. 
While not completely photorealistic, we find that our model can produce convincing samples from 
all $1000$ Imagenet classes.

\begin{figure}[t]
\centering
\includegraphics[width=1.\linewidth]{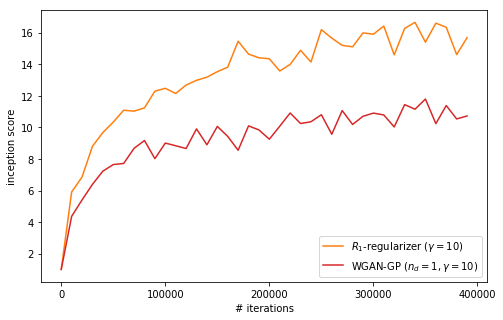}
\caption{
Inception score 
over the number of iterations for GAN training with $R_1$- and WGAN-GP-regularization
when training on Imagenet.
We find that $R_1$-regularization leads to higher inception scores for this
dataset and GAN-architecture.
}
\label{fig:imagenet-inception}
\end{figure}

We also compare the $R_1$-regularizer with WGAN-GP (with $1$ discriminator update per generator update). For this comparison, we did not use the exponential moving average over the weights 
and also did not normalize the embeddings of the labels to the unit sphere.
The resulting inception score\footnote{%
For measuring the inception score, we use the public implementation 
from \url{http://github.com/sbarratt/inception-score-pytorch}.}
over the number
of iterations is visualized in Figure~\ref{fig:imagenet-inception}.
We find that for this dataset and architecture we can achieve higher inception scores when using the $R_1$-regularizer in place of the WGAN-GP regularizer.

\paragraph{celebA and LSUN}
To see if the $R_1$-regularizers helps to train GANs for high-resolution image distributions, we
apply our method to the celebA dataset \cite{liu2015faceattributes} and 
to $4$ subsets of the
LSUN dataset \cite{yu15lsun} with resolution $256 \times 256$.
We use a similar training setup as for the Imagenet experiment, but we use a slightly different architecture
(Table~\ref{tab:architecture-celebA}).
As in the Imagenet-experiment, we use preactivation ResNet-blocks and Leaky RELU-nonlinearities everywhere
and we multiply the output of the ResNet-blocks with $0.1$.
We implemented the network in the PyTorch framework
and use the RMSProp optimizer
with $\alpha = 0.99$ and a learning rate of $10^{-4}$. 
We again found that results can be (slightly) improved by using an exponential moving average with decay $0.999$ over the weights to produce the final models.
As a regularization term, we use the $R_1$-regularizer with $\gamma = 10$. 
For the latent code $z$, we use a $256$ dimensional Gaussian distribution.
The batch size is $64$.
We trained each model for about $300.000$ iterations on 
$2$ GeForce GTX 1080 Ti GPUs.

\begin{figure*}[ht!]
\centering
\includegraphics[width=\linewidth]{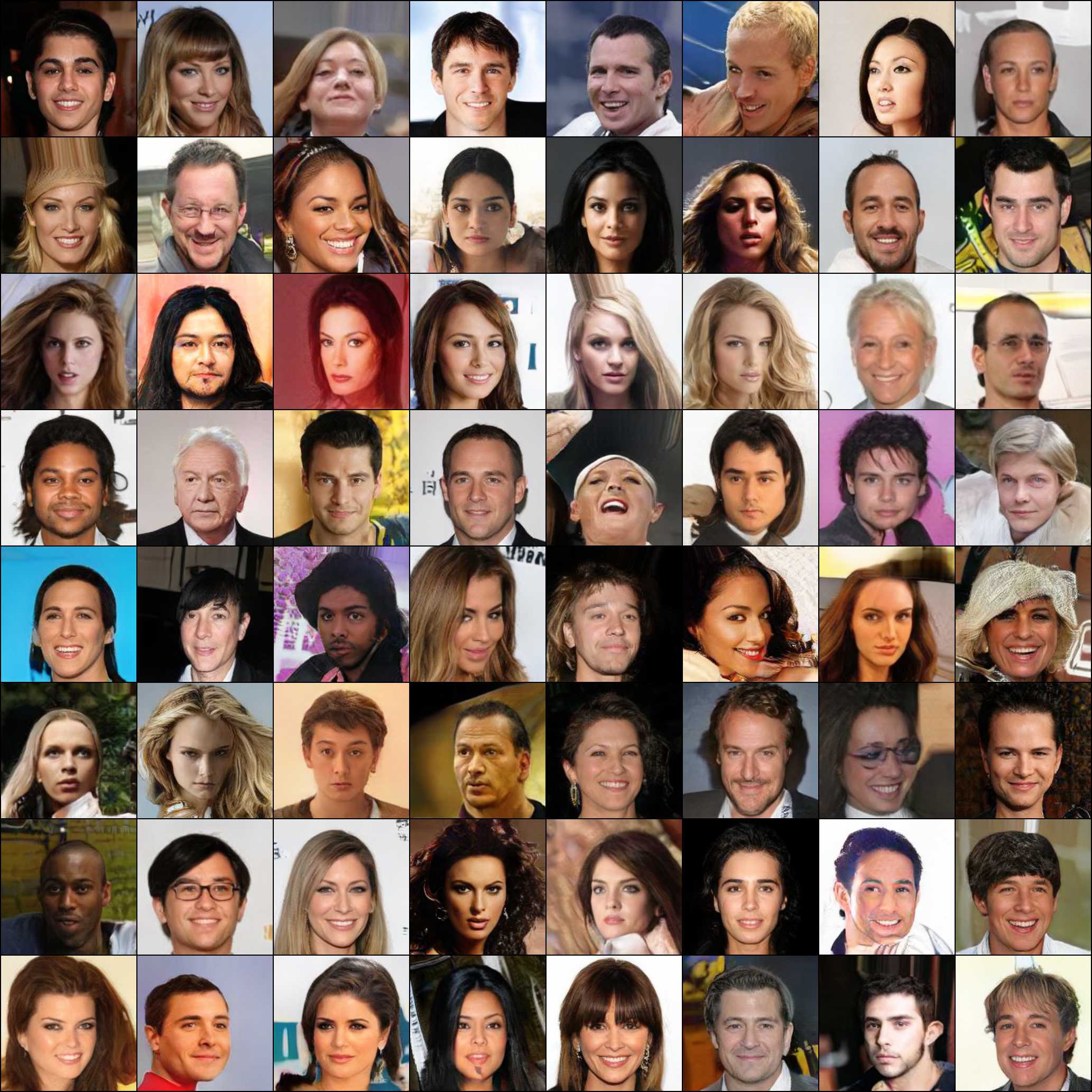}
\caption{Random samples for a GAN trained on the celebA dataset \cite{liu2015faceattributes} ($256 \times 256$) for a 
DC-GAN \cite{radford2015unsupervised} based architecture with additional residual connections \cite{he2016deep}.
For both the generator and the discriminator, we do not use
batch normalization.}
\label{fig:result-celebA}
\end{figure*}

\begin{figure*}[ht!]
\centering
\includegraphics[width=\linewidth]{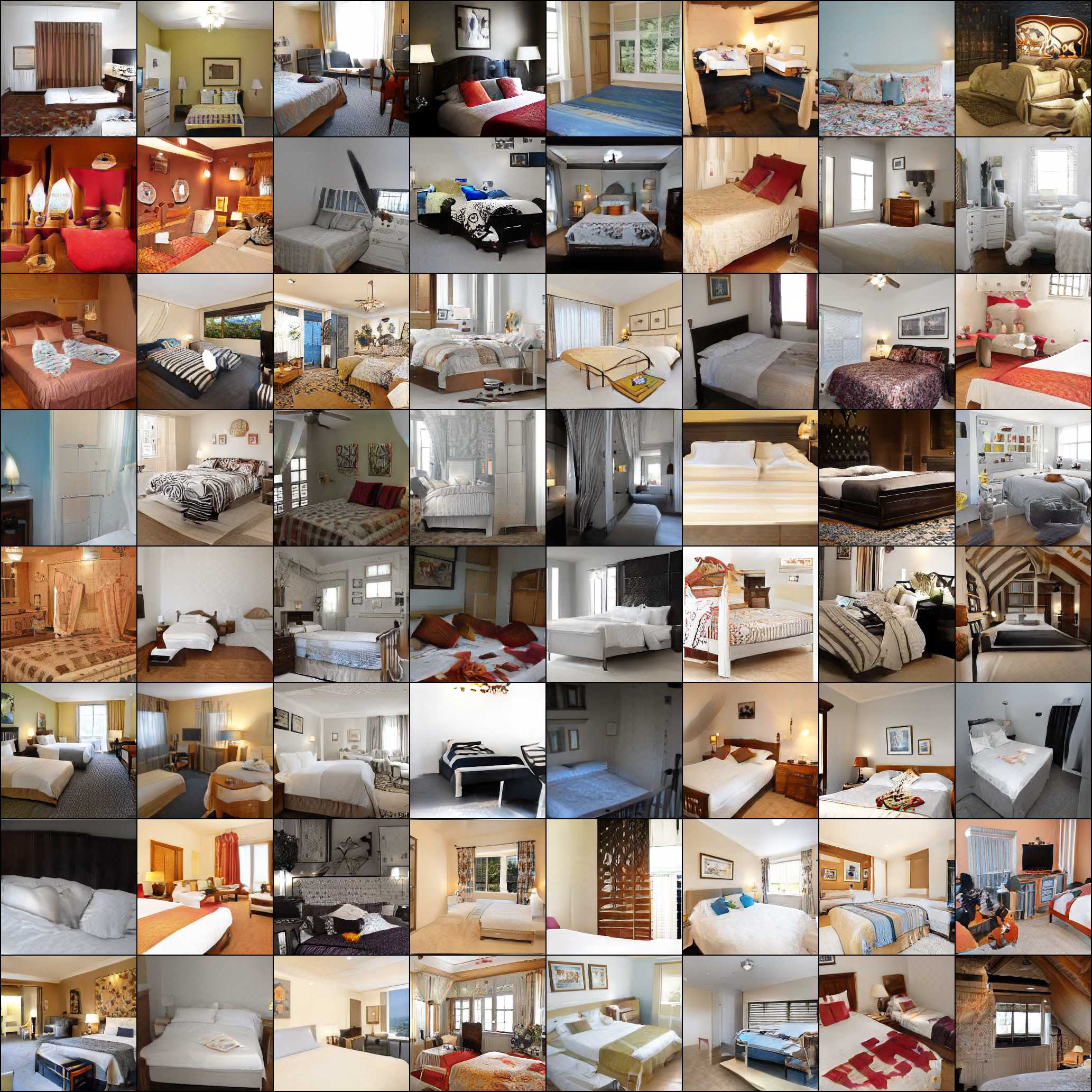}f
\caption{Random samples for a GAN trained  on the LSUN-bedroom dataset \cite{yu15lsun} ($256 \times 256$) for a 
DC-GAN \cite{radford2015unsupervised} based architecture with additional residual connections \cite{he2016deep}.
For both the generator and the discriminator, we do not use
batch normalization.}
\label{fig:result-lsun-bedroom}
\end{figure*}

\begin{figure*}[ht!]
\centering
\includegraphics[width=\linewidth]{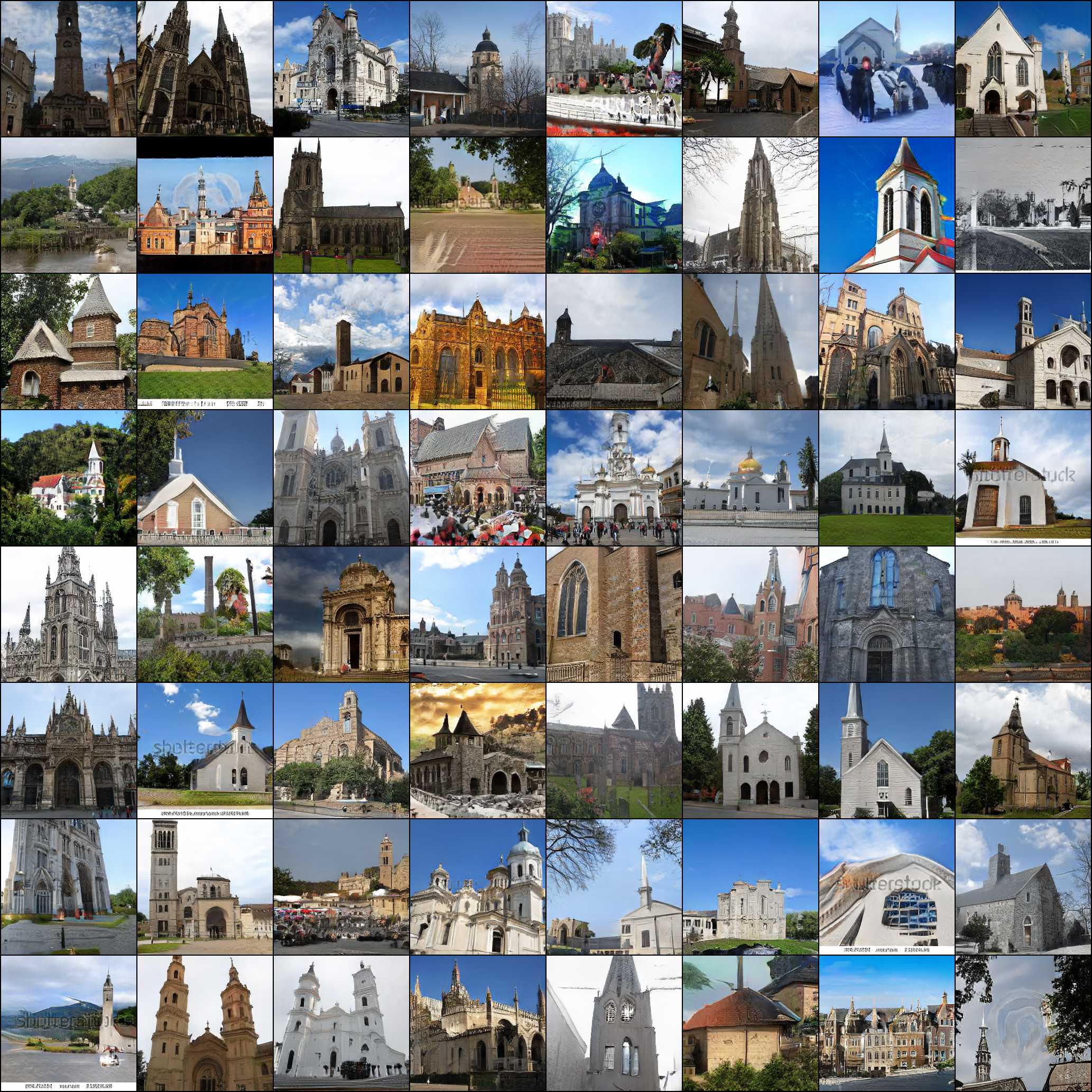}
\caption{Random samples for a GAN trained  on the LSUN-church dataset \cite{yu15lsun} ($256 \times 256$) for a 
DC-GAN \cite{radford2015unsupervised} based architecture with additional residual connections \cite{he2016deep}.
For both the generator and the discriminator, we do not use
batch normalization.}
\label{fig:result-lsun-church}
\end{figure*}

\begin{figure*}[ht!]
\centering
\includegraphics[width=\linewidth]{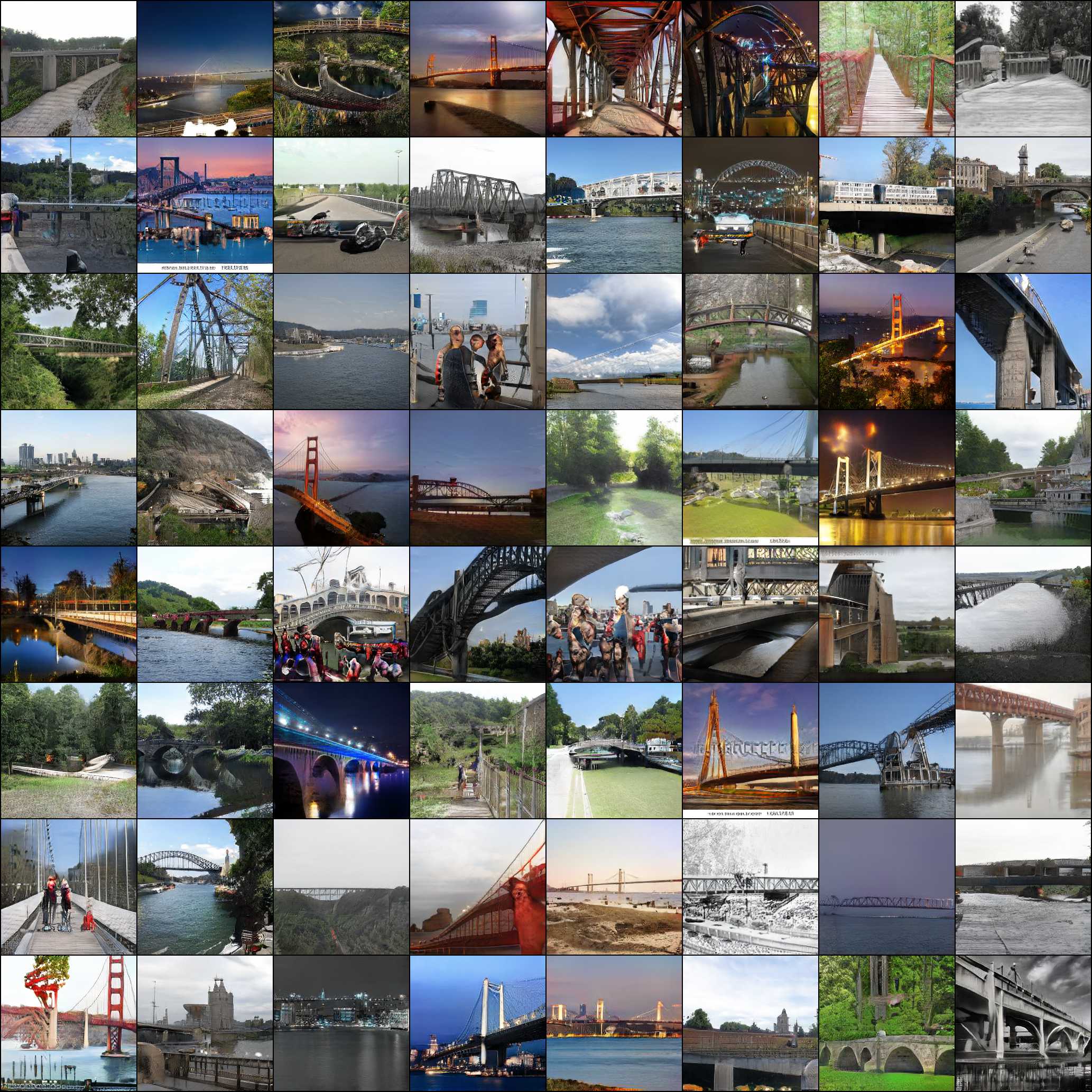}
\caption{Random samples for a GAN trained on the LSUN-bridge dataset \cite{yu15lsun} ($256 \times 256$) for a 
DC-GAN \cite{radford2015unsupervised} based architecture with additional residual connections \cite{he2016deep}.
For both the generator and the discriminator, we do not use
batch normalization.}
\label{fig:result-lsun-bridge}
\end{figure*}

\begin{figure*}[ht!]
\centering
\includegraphics[width=\linewidth]{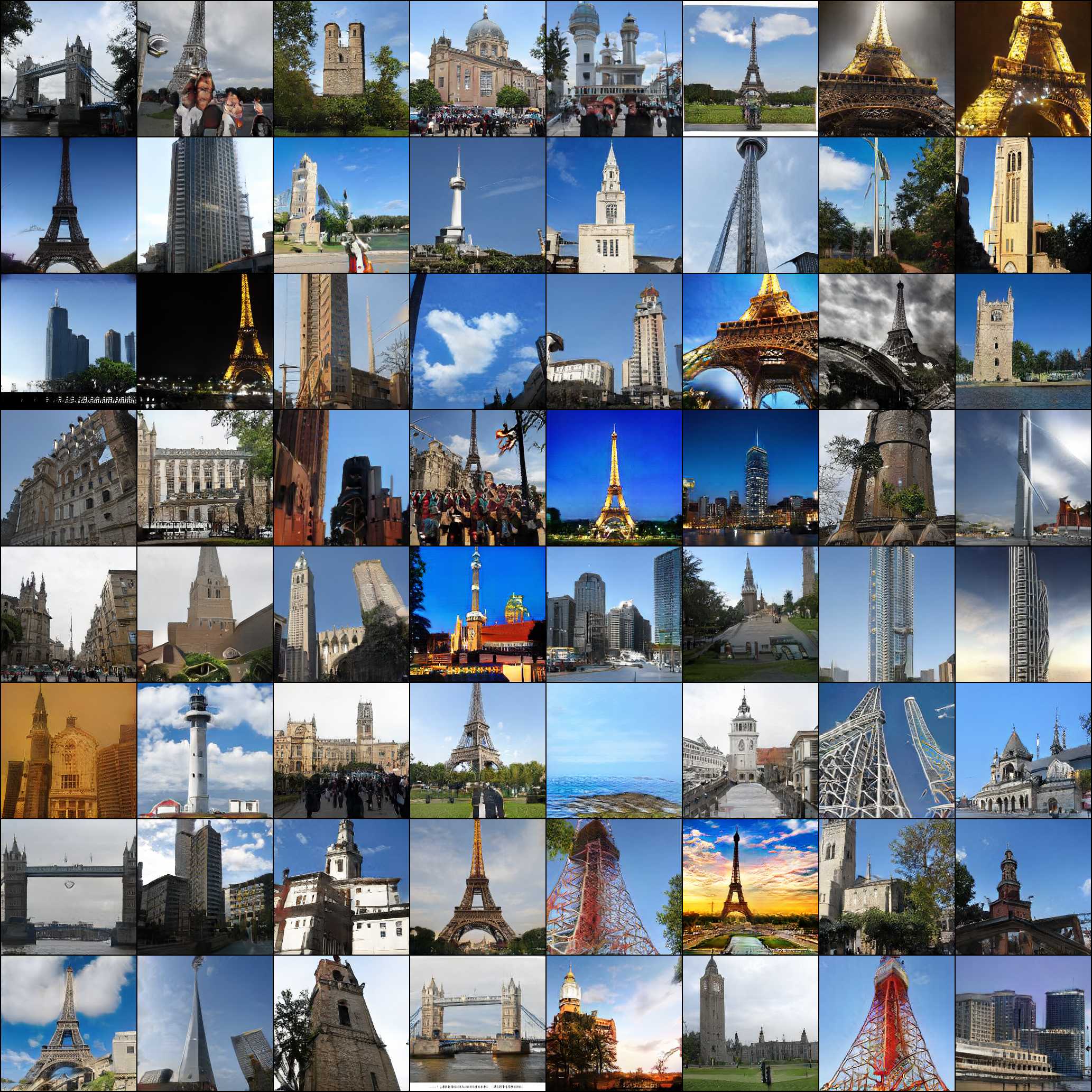}
\caption{Random samples for a GAN trained on the LSUN-tower dataset \cite{yu15lsun} ($256 \times 256$) for a 
DC-GAN \cite{radford2015unsupervised} based architecture with additional residual connections \cite{he2016deep}.
For both the generator and the discriminator, we do not use
batch normalization.}
\label{fig:result-lsun-tower}
\end{figure*}

We find that the $R_1$-regularizer successfully stabilizes training of this architecture.
Some random samples can be seen in Figures~\ref{fig:result-celebA}, \ref{fig:result-lsun-bedroom}, \ref{fig:result-lsun-church}, \ref{fig:result-lsun-bridge}
and \ref{fig:result-lsun-tower}.

\begin{figure*}[ht!]
\centering

\centering
\includegraphics[width=0.8\linewidth]{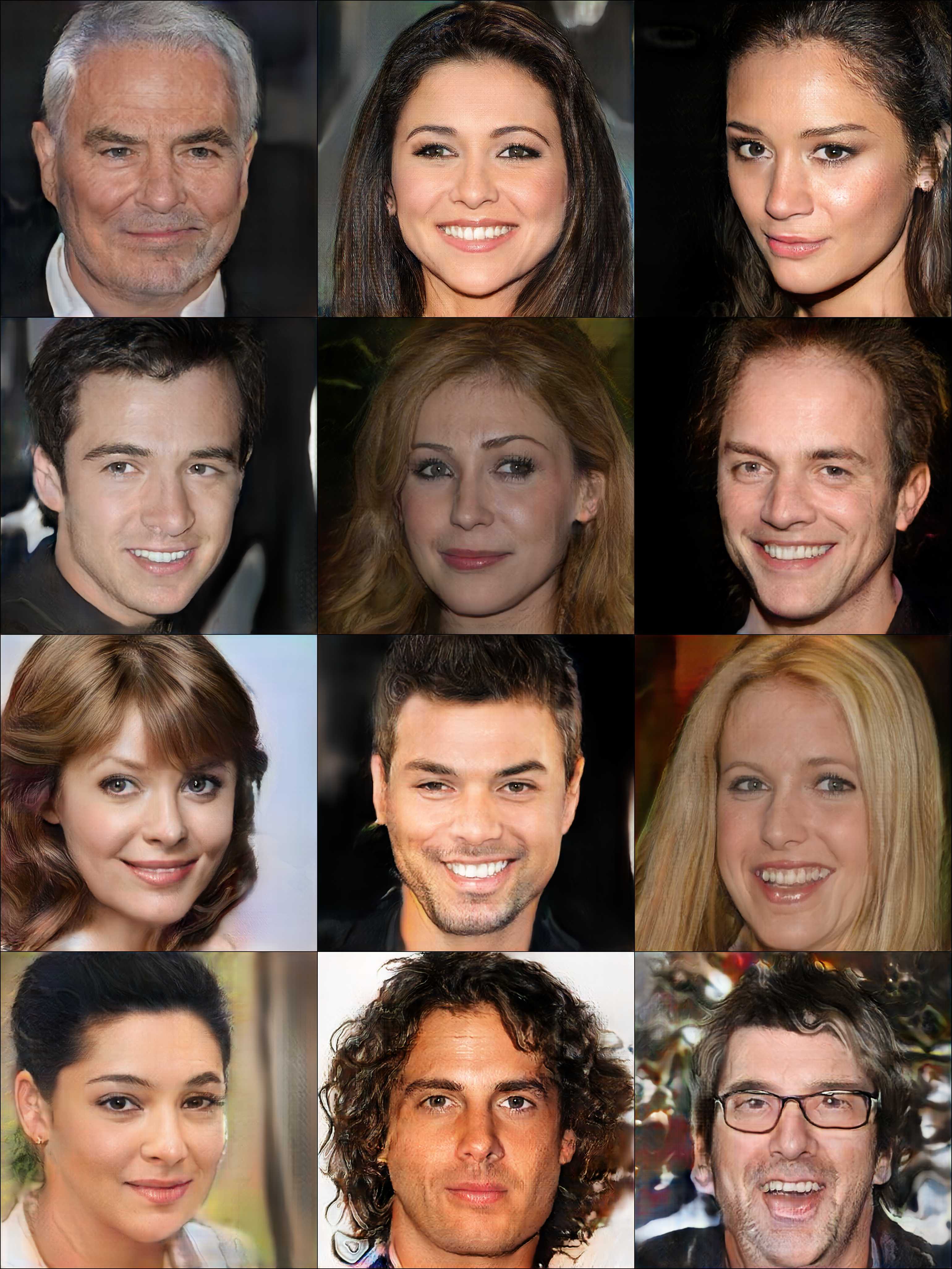}

\caption{Random samples for a GAN trained on the celebA-HQ dataset \cite{karras2017progressive} ($1024 \times 1024$) for a 
DC-GAN \cite{radford2015unsupervised} based architecture with additional residual connections \cite{he2016deep}.
During the whole course of training, we directly train the full-resolution generator and discriminator end-to-end,
i.e. we do not use any of the techniques described in \citet{karras2017progressive}
to stabilize the training. 
}
\label{fig:result-celebAHQ}
\end{figure*}

\paragraph{celebA-HQ}
In addition to the generative model for celebA with resolution $256 \times 256$,
we train a GAN on the celebA-HQ dataset \cite{karras2017progressive} with resolution $1024 \times 1024$.
We use almost the same architecture as for celebA (Table~\ref{tab:architecture-celebA}), but add two more
levels to increase the resolution from $256 \times 256$ to $1024 \times 1024$ and decrease the number of features
from $64$ to $16$.
Because of memory constraints, we also decrease the batch size to $24$.
As in the previous experiments, we use an exponential moving average with decay $0.999$ over the weights to produce the final model.
In contrast to \citet{karras2017progressive}, we train our model end-to-end during the whole course of training, i.e.
we do not use progressively growing GAN-architectures (nor any of the other techniques used by \citet{karras2017progressive} to stabilize the training). We trained the model for about $300.000$ iterations on 
$4$ GeForce GTX 1080 Ti GPUs.
We find that the simple $R_1$-regularizer stabilizes the training,
allowing our model to converge to a good solution without using a progressively growing GAN.
Some random samples are shown in Figure~\ref{fig:result-celebAHQ}.
\vfill

\FloatBarrier

\begin{table}[p]%
\begin{subfigure}[t]{\linewidth}
\centering
\scalebox{0.9}{
\begin{tabular}{lcc} \toprule
Layer & output size & filter\\ \midrule
Fully Connected & $256\cdot4\cdot4$ & $256 \rightarrow 256 \cdot 4 \cdot 4$\\ 
Reshape &   $256 \times 4 \times 4$ & -\\
TransposedConv2D  & $128 \times 8 \times 8$  & $256 \rightarrow 128$ \\ 
TransposedConv2D  & $64 \times 16 \times 16$  & $128 \rightarrow 64$ \\ 
TransposedConv2D  & $3 \times 32 \times 32$  & $64 \rightarrow 3$ \\ 
\bottomrule
\end{tabular}
}
\caption{Generator architecture}
\label{tab:architecture-generator-cifar10}
\end{subfigure}

\begin{subfigure}[t]{\linewidth}
\centering
\scalebox{0.9}{
\begin{tabular}{lcc} \toprule
Layer & output size & filter\\ \midrule
Conv2D  & $64 \times 16 \times 16$  & $3 \rightarrow 64$ \\ 
Conv2D  & $128 \times 8 \times 8$  & $64 \rightarrow 128$ \\
Conv2D  & $256 \times 4 \times 4$  & $128 \rightarrow 256$ \\
Reshape &   $256 \cdot 4 \cdot 4$ & -\\
Fully Connected & $256\cdot4\cdot4$ & $256\cdot4\cdot4 \rightarrow 1$\\ 
\bottomrule
\end{tabular}
}
\caption{Discriminator architecture}
\label{tab:architecture-discriminator-cifar10}
\end{subfigure}
\caption{Architectures for CIFAR-10-experiment.}
\label{tab:architecture-cifar10}
\end{table}

\begin{table}[p]
\begin{subfigure}[t]{\linewidth}
\centering
\scalebox{0.9}{
\begin{tabular}{lcc} \toprule
Layer & output size & filter\\ \midrule
Fully Connected & $1024\cdot4\cdot4$ & $512 \rightarrow 1024 \cdot 4 \cdot 4$ \\ 
Reshape &   $1024 \times 4 \times 4$ & -\\ \midrule
Resnet-Block &  $1024 \times 4 \times 4$ & $1024 \rightarrow 1024 \rightarrow 1024$ \\
Resnet-Block &  $1024 \times 4 \times 4$ & $1024 \rightarrow 1024 \rightarrow 1024$ \\
NN-Upsampling & $1024 \times 8 \times 8$ & - \\ \midrule
Resnet-Block &  $1024 \times 8 \times 8$ & $1024 \rightarrow 1024 \rightarrow 1024$ \\
Resnet-Block &  $1024 \times 8 \times 8$ & $1024 \rightarrow 1024 \rightarrow 1024$ \\
NN-Upsampling & $1024 \times 16 \times 16$ & - \\ \midrule
Resnet-Block &  $512 \times 16 \times 16$ & $1024 \rightarrow 512 \rightarrow 512$ \\
Resnet-Block &  $512 \times 16 \times 16$ & $512 \rightarrow 512 \rightarrow 512$ \\
NN-Upsampling & $512 \times 32 \times 32$ & - \\ \midrule
Resnet-Block &  $256 \times 32 \times 32$ & $512 \rightarrow 256 \rightarrow 256$ \\
Resnet-Block &  $256 \times 32 \times 32$ & $256 \rightarrow 256 \rightarrow 256$ \\
NN-Upsampling & $256 \times 64 \times 64$ & - \\ \midrule
Resnet-Block &  $128 \times 64 \times 64$ & $256 \rightarrow 128 \rightarrow 128$ \\
Resnet-Block &  $128 \times 64 \times 64$ & $128 \rightarrow 128 \rightarrow 128$ \\
NN-Upsampling & $128 \times 128 \times 128$ & - \\ \midrule
Resnet-Block &  $64 \times 128 \times 128$ & $128 \rightarrow 64 \rightarrow 64$ \\
Resnet-Block &  $64 \times 128 \times 128$ & $64 \rightarrow 64 \rightarrow 64$ \\
Conv2D  & $3 \times 128 \times 128$  & $64 \rightarrow 3$ \\

\bottomrule
\end{tabular}
}
\caption{Generator architecture}
\label{tab:architecture-generator-imagenet}
\end{subfigure}
\vspace{0.5cm}
\begin{subfigure}[t]{\linewidth}
\centering
\scalebox{0.9}{
\begin{tabular}{lcc} \toprule
Layer & output size & filter\\ \midrule
Conv2D  & $64 \times 128 \times 128$  & $3 \rightarrow 64$ \\ \midrule
Resnet-Block &  $64 \times 128 \times 128$ & $64 \rightarrow 64 \rightarrow 64$ \\
Resnet-Block &  $128 \times 128 \times 128$ & $64 \rightarrow 64 \rightarrow 128$ \\
Avg-Pool2D & $128 \times 64 \times 64$ & - \\ \midrule
Resnet-Block &  $128 \times 64 \times 64$ & $128 \rightarrow 128 \rightarrow 128$ \\
Resnet-Block &  $256 \times 64 \times 64$ & $128 \rightarrow 128 \rightarrow 256$ \\
Avg-Pool2D & $256 \times 32 \times 32$ & - \\ \midrule
Resnet-Block &  $256 \times 32 \times 32$ & $256 \rightarrow 256 \rightarrow 256$ \\
Resnet-Block &  $512 \times 32 \times 32$ & $256 \rightarrow 256 \rightarrow 512$ \\
Avg-Pool2D & $512 \times 16 \times 16$ & - \\ \midrule
Resnet-Block &  $512 \times 16 \times 16$ & $512 \rightarrow 512 \rightarrow 512$ \\
Resnet-Block &  $1024 \times 16 \times 16$ & $512 \rightarrow 512 \rightarrow 1024$ \\
Avg-Pool2D & $1024 \times 8 \times 8$ & - \\ \midrule
Resnet-Block &  $1024 \times 8 \times 8$ & $1024 \rightarrow 1024 \rightarrow 1024$ \\
Resnet-Block &  $1024 \times 8 \times 8$ & $1024 \rightarrow 1024 \rightarrow 1024$ \\
Avg-Pool2D & $1024 \times 4 \times 4$ & - \\ \midrule
Resnet-Block &  $1024 \times 4 \times 4$ & $1024 \rightarrow 1024 \rightarrow 1024$ \\
Resnet-Block &  $1024 \times 4 \times 4$ & $1024 \rightarrow 1024 \rightarrow 1024$ \\
Fully Connected & $1024\cdot4\cdot4$ & $1024\cdot4\cdot4 \rightarrow 1000$\\ 
\bottomrule
\end{tabular}
}
\caption{Discriminator architecture}
\label{tab:architecture-discriminator-imagenet}
\end{subfigure}
\caption{Architectures for Imagenet-experiment.}
\label{tab:architecture-imagenet}
\end{table}

\begin{table}[p]
\begin{subfigure}[t]{\linewidth}
\centering
\scalebox{0.9}{
\begin{tabular}{lcc} \toprule
Layer & output size & filter\\ \midrule
Fully Connected & $1024\cdot4\cdot4$ & $512 \rightarrow 1024 \cdot 4 \cdot 4$ \\ 
Reshape &   $1024 \times 4 \times 4$ & -\\ \midrule
Resnet-Block  &  $1024 \times 4 \times 4$ & $1024 \rightarrow 1024 \rightarrow 1024$ \\
NN-Upsampling & $1024 \times 8 \times 8$ & - \\ \midrule

Resnet-Block  &  $1024 \times 8 \times 8$ & $1024 \rightarrow 1024 \rightarrow 1024$ \\
NN-Upsampling & $1024 \times 16 \times 16$ & - \\ \midrule
Resnet-Block  &  $512 \times 16 \times 16$ & $1024 \rightarrow 512 \rightarrow 512$ \\
NN-Upsampling & $512 \times 32 \times 32$ & - \\ \midrule
Resnet-Block  &  $256 \times 32 \times 32$ & $512 \rightarrow 256 \rightarrow 256$ \\
NN-Upsampling & $256 \times 64 \times 64$ & - \\ \midrule
Resnet-Block  &  $128 \times 64 \times 64$ & $256 \rightarrow 128 \rightarrow 128$ \\
NN-Upsampling & $128 \times 128 \times 128$ & - \\ \midrule
Resnet-Block  &  $64 \times 128 \times 128$ & $128 \rightarrow 64 \rightarrow 64$ \\
NN-Upsampling & $64 \times 256 \times 256$ & - \\ \midrule
Resnet-Block  &  $64 \times 256 \times 256$ & $64 \rightarrow 64 \rightarrow 64$ \\
Conv2D  & $3 \times 256 \times 256$  & $64 \rightarrow 3$ \\

\bottomrule
\end{tabular}
}
\caption{Generator architecture}
\label{tab:architecture-generator-celebA}
\end{subfigure}
\vspace{0.5cm}
\begin{subfigure}[t]{\linewidth}
\centering
\scalebox{0.9}{
\begin{tabular}{lcc} \toprule
Layer & output size & filter\\ \midrule
Conv2D  & $64 \times 256 \times 256$  & $3 \rightarrow 64$ \\ \midrule
Resnet-Block  &  $64 \times 256 \times 256$ & $64 \rightarrow 64 \rightarrow 64$ \\
Avg-Pool2D & $64 \times 128 \times 128$ & - \\ \midrule
Resnet-Block  &  $128 \times 128 \times 128$ & $64 \rightarrow 64 \rightarrow 128$ \\
Avg-Pool2D & $128 \times 64 \times 64$ & - \\ \midrule
Resnet-Block  &  $256 \times 64 \times 64$ & $128 \rightarrow 128 \rightarrow 256$ \\
Avg-Pool2D & $256 \times 32 \times 32$ & - \\ \midrule
Resnet-Block  &  $512 \times 32 \times 32$ & $256 \rightarrow 256 \rightarrow 512$ \\
Avg-Pool2D & $512 \times 16 \times 16$ & - \\ \midrule
Resnet-Block  &  $1024 \times 16 \times 16$ & $512 \rightarrow 512 \rightarrow 1024$ \\
Avg-Pool2D & $1024 \times 8 \times 8$ & - \\ \midrule
Resnet-Block  &  $1024 \times 8 \times 8$ & $1024 \rightarrow 1024 \rightarrow 1024$ \\
Avg-Pool2D & $1024 \times 4 \times 4$ & - \\ \midrule
Fully Connected & $1024\cdot4\cdot4$ & $1024\cdot4\cdot4 \rightarrow 1$\\ 
\bottomrule
\end{tabular}
}
\caption{Discriminator architecture}
\label{tab:architecture-discriminator-celebA}
\end{subfigure}
\caption{Architectures for LSUN- and celebA-experiments.}
\label{tab:architecture-celebA}
\end{table}

\begin{figure}[ht!]

\begin{subfigure}[t]{0.49\linewidth}
\includegraphics[width=\linewidth]{img/simgd/gan}
 \caption{Standard GAN}
\end{subfigure}
\begin{subfigure}[t]{0.49\linewidth}
\includegraphics[width=\linewidth]{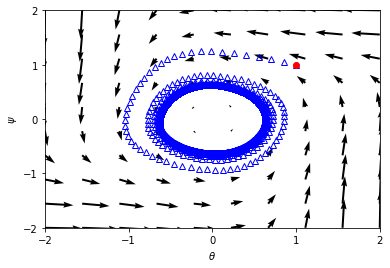}
 \caption{Non-saturating GAN}
\end{subfigure}

\begin{subfigure}[t]{0.49\linewidth}
\includegraphics[width=\linewidth]{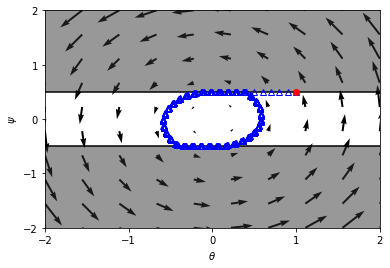}
\caption{WGAN}\label{fig:simgd-wgan}
\end{subfigure}
\begin{subfigure}[t]{0.49\linewidth}
\includegraphics[width=\linewidth]{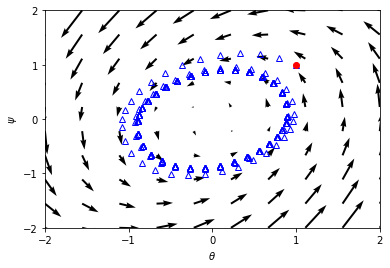}
 \caption{WGAN-GP}
\end{subfigure}

\begin{subfigure}[t]{0.49\linewidth}
\includegraphics[width=\linewidth]{img/simgd/gan_consensus}
 \caption{Consensus optimization}
\end{subfigure}
\begin{subfigure}[t]{0.49\linewidth}
\includegraphics[width=\linewidth]{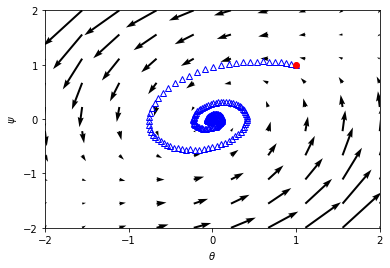}
 \caption{Instance noise}
\end{subfigure}

\begin{subfigure}[t]{0.49\linewidth}
\includegraphics[width=\linewidth]{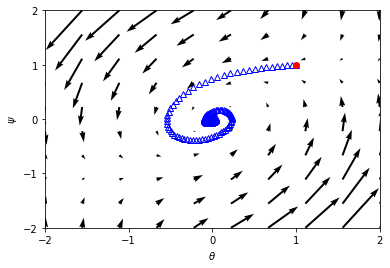}
 \caption{Gradient penalty}
\end{subfigure}
\begin{subfigure}[t]{0.49\linewidth}
\includegraphics[width=\linewidth]{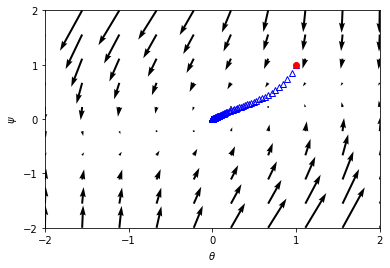}
 \caption{Gradient penalty (CR)}
\end{subfigure}

\caption{Convergence properties of different GAN training algorithms using simultaneous
gradient descent. The shaded area in Figure~\ref{fig:simgd-wgan} visualizes the set of forbidden values for 
the discriminator parameter $\psi$. The starting iterate is marked in red.}
\end{figure}

\begin{figure}[ht!]

\begin{subfigure}[t]{0.49\linewidth}
\includegraphics[width=\linewidth]{img/altgd1/gan}
\vspace{-15pt}
 \caption{Standard GAN}
\end{subfigure}
\begin{subfigure}[t]{0.49\linewidth}
\includegraphics[width=\linewidth]{img/altgd1/nsgan}
\vspace{-15pt}
 \caption{Non-saturating GAN}
\end{subfigure}

\begin{subfigure}[t]{0.49\linewidth}
\includegraphics[width=\linewidth]{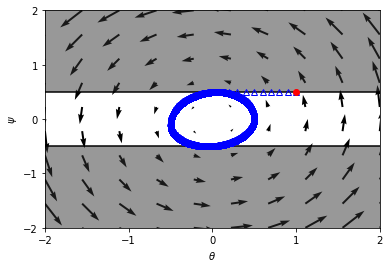}
\vspace{-15pt}
\caption{WGAN}\label{fig:altgd1-wgan}
\end{subfigure}
\begin{subfigure}[t]{0.49\linewidth}
\includegraphics[width=\linewidth]{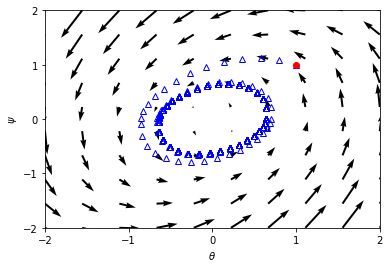}
\vspace{-15pt}
 \caption{WGAN-GP}
\end{subfigure}

\begin{subfigure}[t]{0.49\linewidth}
\includegraphics[width=\linewidth]{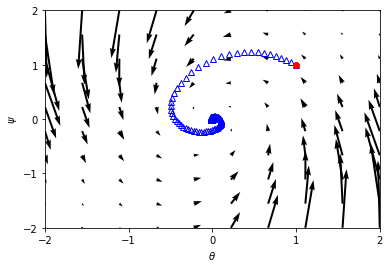}
\vspace{-15pt}
 \caption{Consensus optimization}
\end{subfigure}
\begin{subfigure}[t]{0.49\linewidth}
\includegraphics[width=\linewidth]{img/altgd1/gan_instnoise}
\vspace{-15pt}
 \caption{Instance noise}
\end{subfigure}

\begin{subfigure}[t]{0.49\linewidth}
\includegraphics[width=\linewidth]{img/altgd1/gan_gradpen}
\vspace{-15pt}
 \caption{Gradient penalty}
\end{subfigure}
\begin{subfigure}[t]{0.49\linewidth}
\includegraphics[width=\linewidth]{img/altgd1/gan_gradpen_critical}
\vspace{-15pt}
 \caption{Gradient penalty (CR)}
\end{subfigure}

\caption{Convergence properties of different GAN training algorithms using alternating
gradient descent with $1$ discriminator update per generator update
The shaded area in Figure~\ref{fig:altgd1-wgan} visualizes the set of forbidden values for 
the discriminator parameter $\psi$. The starting iterate is marked in red.}
\end{figure}

\begin{figure}[ht!]

\begin{subfigure}[t]{0.49\linewidth}
\includegraphics[width=\linewidth]{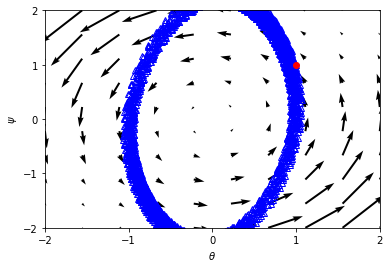}
\vspace{-15pt}
 \caption{Standard GAN}
\end{subfigure}
\begin{subfigure}[t]{0.49\linewidth}
\includegraphics[width=\linewidth]{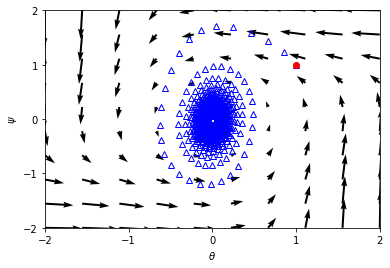}
\vspace{-15pt}
 \caption{Non-saturating GAN}
\end{subfigure}

\begin{subfigure}[t]{0.49\linewidth}
\includegraphics[width=\linewidth]{img/altgd5/wgan}
\vspace{-15pt}
\caption{WGAN}\label{fig:altgd5-wgan}
\end{subfigure}
\begin{subfigure}[t]{0.49\linewidth}
\includegraphics[width=\linewidth]{img/altgd5/wgan_gp}
\vspace{-15pt}
 \caption{WGAN-GP}
\end{subfigure}

\begin{subfigure}[t]{0.49\linewidth}
\includegraphics[width=\linewidth]{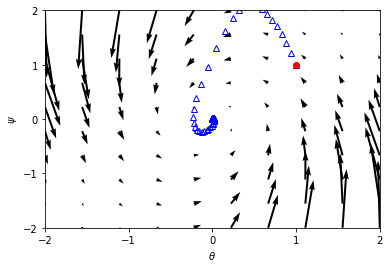}
\vspace{-15pt}
 \caption{Consensus optimization}
\end{subfigure}
\begin{subfigure}[t]{0.49\linewidth}
\includegraphics[width=\linewidth]{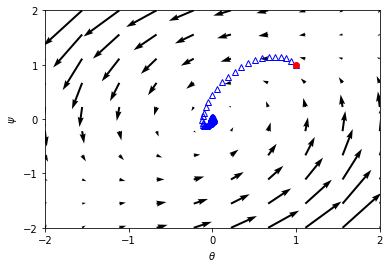}
\vspace{-15pt}
 \caption{Instance noise}
\end{subfigure}

\begin{subfigure}[t]{0.49\linewidth}
\includegraphics[width=\linewidth]{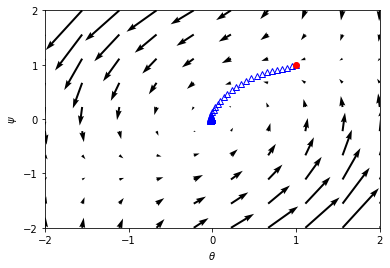}
\vspace{-15pt}
 \caption{Gradient penalty}
\end{subfigure}
\begin{subfigure}[t]{0.49\linewidth}
\includegraphics[width=\linewidth]{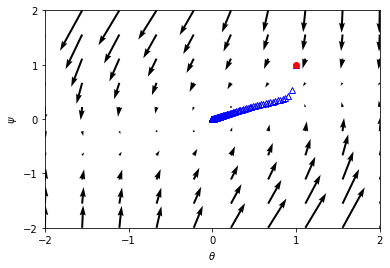}
\vspace{-15pt}
 \caption{Gradient penalty (CR)}
\end{subfigure}

\caption{Convergence properties of different GAN training algorithms using alternating
gradient descent with $5$ discriminator updates per generator update. 
The shaded area in Figure~\ref{fig:altgd5-wgan} visualizes the set of forbidden values for 
the discriminator parameter $\psi$. The starting iterate is marked in red.}
\end{figure}

\begin{figure*}[ht!]
\centering
\begin{subfigure}[t]{0.29\linewidth}
\includegraphics[width=\linewidth]{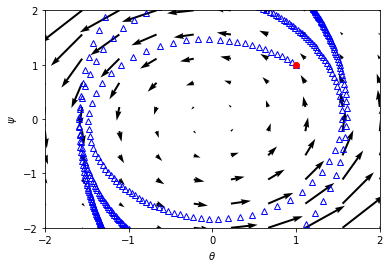}
 \caption{SimGD}
\end{subfigure}
\begin{subfigure}[t]{0.29\linewidth}
\includegraphics[width=\linewidth]{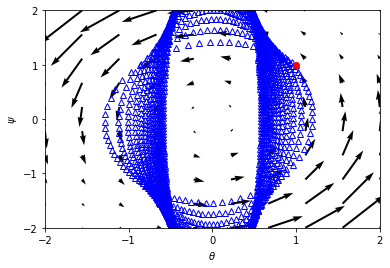}
 \caption{AltGD ($n_d=1$)}
\end{subfigure}
\begin{subfigure}[t]{0.29\linewidth}
\includegraphics[width=\linewidth]{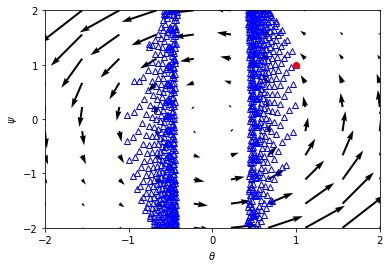}
\caption{AltGD ($n_d=5$)}
\end{subfigure}

\caption{Convergence properties of our GAN 
using two time-scale training as proposed by \citet{DBLP:conf/nips/HeuselRUNH17}.
For the Dirac-GAN we do not see any sign of convergence
when training with two time-scales. The starting iterate is marked in red.}
\label{fig:ttur}
\end{figure*}

\end{document}